%% file: usenix.tex
\documentclass[letterpaper,twocolumn,10pt]{article}
\usepackage{usenix}

\usepackage[square,numbers]{natbib}

\usepackage{tikz}
\usepackage{amsmath}

\usepackage{graphicx}
\usepackage{multirow}
\usepackage{makecell}
\usepackage{wrapfig}
\usepackage{amsfonts}
\usepackage{amsthm}
\usepackage{bm}
\usepackage{bbm}
\DeclareMathAlphabet\mathbfcal{OMS}{cmsy}{b}{n}

\usepackage{algorithm,algpseudocode}

\algnewcommand{\algorithmicforeach}{\textbf{for}}
\algdef{SE}[FOR]{ForEach}{EndForEach}[1]
  {\algorithmicforeach\ #1\ \algorithmicdo}
  {\algorithmicend\ \algorithmicforeach}

\usepackage{subcaption}

\newcommand{\red}[1]{\textcolor[RGB]{255, 49, 49}{#1}}
\newcommand{\green}[1]{\textcolor[RGB]{34,169,34}{#1}}
\newcommand{\blue}[1]{\textcolor[RGB]{0, 29, 246}{#1}}
\newcommand{\revision}[1]{#1}

\makeatletter
\renewcommand*{\p@section}{\S\,}
\renewcommand*{\p@subsection}{\S\,}
\makeatother

\usepackage{booktabs}

\usepackage{subcaption}
\usepackage[export]{adjustbox}

\usepackage{color, colortbl}
\definecolor{LightGray}{gray}{0.85}
\definecolor{White}{gray}{1.0}
\definecolor{Celadon}{RGB}{175, 225, 175}

\newcommand{\gc}{\cellcolor{LightGray}}
\newcommand{\pc}{\cellcolor{pink}}
\newcommand{\grc}{\cellcolor{Celadon}}

\input{math_command.tex}

\input{macro.tex}

\begin{document}

\date{}



\title{Towards A \underline{Proactive} ML Approach for Detecting Backdoor Poison Samples}



\author{
{\rm Xiangyu Qi}\\
Princeton University
\and
{\rm Tinghao Xie}\\
Princeton University
\and
{\rm Jiachen T. Wang}\\
Princeton University
 \and
 {\rm Tong Wu}\\
Princeton University
 \and
 {\rm Saeed Mahloujifar}\\
Princeton University
 \and
 {\rm Prateek Mittal}\\
Princeton University
}

\maketitle

\begin{abstract}

Adversaries can embed backdoors in deep learning models by introducing backdoor poison samples into training datasets. In this work, we investigate how to detect such poison samples to mitigate the threat of backdoor attacks. \textbf{First}, we uncover {\textit{a post-hoc workflow}} underlying most prior work, where defenders passively allow the attack to proceed and then leverage the characteristics of the post-attacked model to uncover poison samples. We reveal that this workflow does not fully exploit defenders' capabilities, and defense pipelines built on it are prone to failure or performance degradation in many scenarios. \textbf{Second}, we suggest a paradigm shift by promoting {\textit{a proactive mindset}} in which defenders engage proactively with the entire model training and poison detection pipeline, \textit{directly enforcing and magnifying distinctive characteristics of the post-attacked model to facilitate poison detection}. Based on this, we formulate a unified framework and provide practical insights on designing detection pipelines that are more robust and generalizable. 
\textbf{Third}, we introduce the technique of \textit{{Confusion Training}}~(CT) as a concrete instantiation of our framework. CT applies an additional poisoning attack to the already poisoned dataset, actively decoupling benign correlation while exposing backdoor patterns to detection. Empirical evaluations on 4 datasets and 14 types of attacks validate the superiority of CT over \revision{14} baseline defenses.\footnote{Our code repository is available at \url{https://github.com/Unispac/Fight-Poison-With-Poison}}


\end{abstract}

\input{sections/introduction}

\input{sections/preliminaries}

\input{sections/methodology}

\input{sections/confusion_training}

\input{sections/experiments}

\input{sections/discussions}

\input{sections/related}

\input{sections/conclusion}


\bibliographystyle{plainnat}
\bibliography{reference}

\newpage

\input{sections/appendix}


\end{document}

%% file: math_command.tex
\usepackage{dsfont}
\usepackage{bm}

\newtheorem{theorem}{Theorem}
\newtheorem{lemma}[theorem]{Lemma}
\newtheorem{definition}[theorem]{Definition}

\newtheorem{remark-star}{Remark}
\newtheorem{remark-star-1}{Remark}

\newtheorem{corollary}[theorem]{Corollary}

\newtheorem*{proof-sketch}{Proof Sketch}
\usepackage{booktabs}

\DeclareMathOperator*{\argmin}{\arg\!\min}
\DeclareMathOperator*{\argmax}{\arg\!\max}

\newcommand{\R}{\mathbb{R}}
\newcommand{\eps}{\varepsilon}

\newcommand{\norm}[1]{\left\lVert#1\right\rVert}
\newcommand{\N}{\mathcal{N}}
\newcommand{\I}{\mathbf{I}}

\newcommand{\A}{\mathcal{A}}
\newcommand{\erm}{\A}
\newcommand{\mH}{\mathcal{H}}
\newcommand{\Ind}{\mathds{1}}

\newcommand{\mX}{\mathcal{X}}

\newcommand{\clean}{\mathcal{P}}
\newcommand{\cleandist}{\mathcal{P}}
\newcommand{\trigger}{\mathcal{T}}
\newcommand{\labeltrigger}{\mathcal{L}}
\newcommand{\poiind}{\mathcal{J}}
\newcommand{\Dpoi}{D_{\mathrm{poison}}}
\newcommand{\Dclean}{D_{\mathrm{clean}}}
\newcommand{\Dpoiall}{\widetilde{D}}
\newcommand{\bdmodel}{\widetilde{\theta}}
\newcommand{\Dreserve}{D_{\mathrm{reserve}}}
\newcommand{\thres}{\tau}

\newcommand{\bmu}{\mathbf{\mu}}

\newcommand{\poi}{\mathrm{poi}}

\newcommand{\thetact}{\theta_{\mathrm{ct}}}
\newcommand{\decf}{\xi}
\newcommand{\charf}{\phi}

\newcommand{\Da}{\Dpoiall}
\newcommand{\Dconf}{D_{\mathrm{conf}}}
\newcommand{\Dzero}{D_{-1}}
\newcommand{\Done}{D_{+1}}

\newcommand{\Dzerobar}{\bar D_{-1}}
\newcommand{\Donebar}{\bar D_{+1}}

\newcommand{\kb}{k}

\newcommand{\loss}{\ell}

\newcommand{\xstar}{x^*}

\newcommand{\atclean}{a_{(x^*, y^*), \mathrm{cln}}^{(t)}}
\newcommand{\atbackdoor}{a_{(x^*, y^*), \mathrm{bkd}}^{(t)}}

\newcommand{\azeroclean}{a_{(x^*, y^*), \mathrm{cln-m}}}
\newcommand{\azerobackdoor}{a_{(x^*, y^*), \mathrm{bkd-m}}}

\newcommand{\epsone}{\eps_1}
\newcommand{\epstwo}{\eps_2}
\newcommand{\epsb}{\eps_b}

\newcommand{\zone}{\epsone \sum_{(x, y) \in \Dconf} \beta_{(x, y)}}
\newcommand{\ztwo}{\N \left(0, 2k\mu^2  \sum_{(x, y) \in \Dconf} \beta_{(x, y)} + \kb \nu^2 \sum_{(x, y) \in \Dpoi} \beta_{(x, y)} \right) }
\newcommand{\zthree}{\N\left(0, \kb \nu^2 \sum_{(x, y) \in \Dconf} \beta_{(x, y)} \right)}

\newcommand{\tilden}{\widetilde{n}}

%% file: macro.tex
\newif\iffinal

\iffinal
    \newcommand{\tianhao}[1]{}
    \newcommand{\tong}[1]{}
    \newcommand{\xiangyu}[1]{}
    \newcommand{\tinghao}[1]{}
\else
    \newcommand{\tianhao}[1]{{\bf \textcolor{purple}{[Tianhao: #1]}}}
    \newcommand{\xiangyu}[1]{{\bf \textcolor[RGB]{34,139,34}{Xiangyu: #1}}}
    \newcommand{\tong}[1]{\textcolor[RGB]{205,133,63}{Tong: #1}}
    \newcommand{\tinghao}[1]{{\textcolor{orange}{[Tinghao: #1]}}}
    
\fi

%% file: sections/introduction.tex
\section{Introduction}
\label{sec-1:intro}

Deep learning relies on large datasets~\cite{lin2014microsoft,russakovsky2015imagenet,thomee2016yfcc100m}. Yet, the creation of these datasets often involves automation and outsourcing, making it difficult to ensure strict supervision and leaving them vulnerable to \textit{backdoor poisoning attacks}~\cite{Chen2017TargetedBA,gu2017badnets,li2022backdoor}. In these attacks, a typical adversary will manipulate a few training samples by planting a \textit{backdoor trigger}~(e.g., a pixel patch) and (mis)labeling them as a \textit{target class}. These manipulated samples are referred to as \textit{backdoor poison samples}, and the compromised dataset is called a \textit{poisoned dataset}. This manipulation creates a backdoor correlation between the trigger and target class, causing models trained on the poisoned dataset to learn this backdoor while still maintaining normal behaviors under standard evaluation metrics. Backdoor poisoning attacks pose a significant risk as they allow adversaries to stealthily control models. To combat this, \textit{we investigate methods for detecting backdoor poison samples in potentially
poisoned datasets} as an additional safeguard before they are used by downstream applications.

\textbf{Limitations of Prior Works.}\footnote{\revision{The discussion is strictly within the context of \emph{detecting} backdoor poison samples; our analysis of limitations is not directly applicable to other tracks of defenses that focus on alternative defensive goals (refer Sec~\ref{subsec:goal_and_capability_of_our_defense}).}} Prior works on the detection of backdoor poison samples have primarily employed \textit{a post-hoc workflow}~\cite{chen2018activationclustering,gao2019strip,hayase21a,tran2018spectral,tang2021demon,chou2020sentinet}. In this workflow, defenders passively allow the attack to first proceed without intervention, by training a model on the given poisoned dataset using a routine procedure~(without any defense), resulting in the model being backdoored. 
By analyzing the model's behaviors, defenders then try to trace the attack back to the poison samples in the dataset. The rationale behind these post-hoc approaches is a passive assumption that post-attacked models~(i.e., backdoored models) will spontaneously exhibit distinctive behaviors~(e.g., latent separation characteristics~\cite{qi2023revisiting}) on poison and clean samples, which can be utilized to distinguish between the two populations~\cite{tran2018spectral,chen2018activationclustering,hayase21a,tang2021demon}.
However, in practice, these post-hoc approaches are prone to failure or performance degradation in many scenarios, because the characteristics they rely on could often be inherently weak~(e.g., when the poison rate is low~\cite{tran2018spectral,hayase21a}) or be deliberately suppressed by adaptive attacks~\cite{tang2021demon,qi2023revisiting}. \textit{In this work, we point out that these failure modes should be attributed to the underlying post-workflow which passively counts on post-hoc characteristics that are not within the control of defenders}. 

\textbf{Our Proposal: proactively enforce and magnify distinctive characteristics of the post-attacked model to facilitate poison detection.} To alleviate the limitations of the post-hoc workflow, this work advocates for a \underline{\textit{proactive mindset}} as an alternative. Instead of passively allowing the attack to proceed as per the adversary’s expectation and hoping for distinctive behaviors to emerge in the post-attacked model, we propose that defenders should engage proactively with the entire model training and detection pipeline. This involves directly enforcing and magnifying distinctive characteristics of the post-attacked model to facilitate poison detection. The key insight is that defenders have a \textit{"home field" advantage} against backdoor poisoning attacks~(that has not been fully exploited by post-hoc approaches), as they still maintain full control over the system space after the dataset has been compromised. This control allows defenders to strategically intervene in the attack process~(e.g., post-process data samples, manipulate training procedures) in a way that can proactively enforce post-attacked models to exhibit discriminative characteristics (as per the defenders' expectations), which can be more reliably used to separate clean and poison samples. We formulate this proposal within an abstract framework (Sec~\ref{subsec:proactive_mindset}) and provide practical insights on designing proactive poison detectors that are more robust and generalizable. 

\input{sections/assets/overview}

\textbf{Confusion Training~(CT): a concrete instantiation of our proposal.} To showcase how our proposal may empower the advancement of future research on backdoor poison samples detection, we introduce the technique of \textit{Confusion Training}~(CT) as a concrete instantiation of the proposal. As illustrated in Fig~\ref{fig:overview}, CT produces an \underline{inference model} via launching a training procedure jointly on \textit{regular batches} from the poisoned dataset and \textit{confusion batches} consisting of randomly labeled clean samples from a reserved clean set~(much smaller than the poisoned dataset). The random labeling decouples the benign correlations between normal semantic
features and semantic labels in the confusion batch. During training, a confusion batch is also assigned a \textit{large weight} while a regular batch is only assigned a small weight. In this way, the confusion batch serves as another set of poison samples that strongly obscure the benign correlations, making the clean data points hard to be fitted. 
On the other hand, since the confusion batch does not contain backdoor triggers, \textit{correlations between the trigger and target class still remain intact}. Therefore, the resulting inference model \revision{fails to fit most clean samples but still correctly fits most backdoor poison samples.} This allows defenders to distinguish between clean and poison samples based on the inference model's \underline{state of fitting}.

\textbf{Emperical and Theoretical Analysis.} We extensively evaluate a diverse set of 14 backdoor poisoning attacks~\cite{gu2017badnets,Chen2017TargetedBA,liu2017trojaning,turner2019label,barni2019new,nguyen2020input,li2021invisible,tang2021demon,qi2023revisiting,severi2021explanation,wu2022just,nguyen2021wanet} and 4 benchmark datasets, covering domains of both image~(CIFAR10~\cite{krizhevsky2009learning}, GTSRB~\cite{stallkamp2012man}, ImageNet~\cite{deng2009imagenet}) and malware~(Ember~\cite{anderson2018ember}) classifications. By comparing with \revision{14} baseline defenses~\cite{chou2020sentinet,gao2019strip,tran2018spectral,chen2018activationclustering,tang2021demon,hayase21a,liu2018fine,wang2019neural,li2021anti,li2021neural, zeng2021rethinking,huang2022backdoor,wang2022training,tao2022model} along with ablation studies, we show the effectiveness and superiority of the Confusion Training~(CT) defense, illustrating the potential of the proactive mindset that we promote in this work. \revision{We also provide a theoretical analysis in a simplified setting to formally illustrate the working principles of CT in Appendix~\ref{appendix:theory}.}

Finally, we summarize our contributions as follows:
\begin{itemize}
    \item We uncover a post-hoc workflow that is prevalent in previous research on detecting backdoor poison samples, and reveal the limitations it imposes on prior arts.
    \item We propose a proactive mindset as an alternative and formulate it within an abstract framework, providing grounded insights on designing proactive poison detection pipelines that are more robust and generalizable.
    \item We introduce Confusion Training~(CT) as a concrete illustration of our proposal, whose effectiveness is supported by both empirical and theoretical results. We position CT as evidence to showcase how our proposal of the proactive mindset may inspire future advancement in the detection of backdoor poison samples.
\end{itemize}

%% file: sections/assets/overview.tex
\begin{figure}[h!]
\begin{center}
\includegraphics[width=0.48\textwidth]{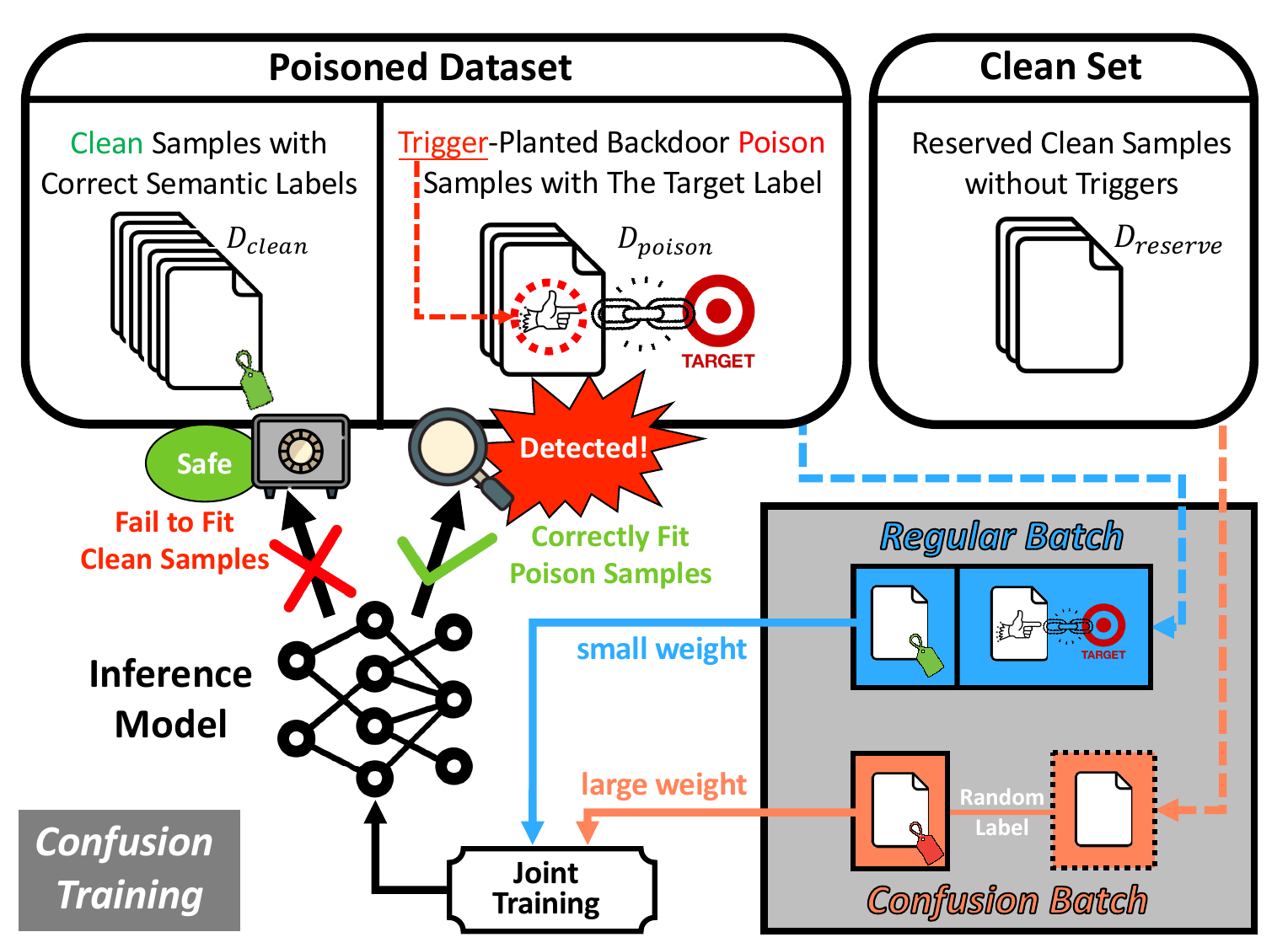}
\end{center}
\caption{
\textbf{Overview of our confusion training (CT) method for poison samples detection.} During training the inference model, we utilize randomly labeled clean samples (\textit{\textbf{confusion batch}}) that do not contain triggers to proactively decouple benign correlations. \revision{The resulting model thus fails to fit clean samples while still fitting backdoor poison samples correctly.}
}
\label{fig:overview}
\end{figure}

%% file: sections/preliminaries.tex
\section{Preliminaries}
\label{sec:background}


In this section, we define our setup (Sec \ref{subsec:notaitons}), formulate the threat model of backdoor poisoning attacks (Sec \ref{subsec:threat_model}), and clarify the goals and capabilities of our defense (Sec \ref{subsec:goal_and_capability_of_our_defense}).

\subsection{Model, Dataset and Training Procedure}
\label{subsec:notaitons}

We consider deep neural network~(DNN) classification model $f(\cdot;\theta): \mX \mapsto \Delta^{C}$, where $\mX := \R^d$ is the input space, $C$ is the number of classes, $\Delta^{C}$
is the space of probability simplex over $C$ classes, and $\theta$ is the model parameters. 
Note that $f(x; \theta)$ outputs a probability distribution over all possible classes. 
We also define the hard-label classifier $F(\cdot;\theta) := \argmax_c f_c(\cdot;\theta)$ that gives the label prediction. In general, a deep neural network $f(\cdot;\theta)$ can be decomposed as $l \circ h(\cdot;\theta)$, where $l$ is the last linear prediction layer, and $h$ is 
the feature extractor before the last linear layer. 
Given an input $x \in \mX$, we call $h(x;\theta) \in \mH$ the \emph{latent representation} of $x$ w.r.t. the model $f(\cdot;\theta)$ where $\mH$ denotes the space of latent representation. $l(\cdot;\theta)$ transforms the latent representation into a probability distribution over the classes $l \circ h(x;\theta) \in \Delta^{C}$. We use $\clean$ to denote the clean data distribution over $\mX \times \{1,\dots,C\}$, and $D = \{(x_i,y_i)\}_{i=1}^n$ is a clean dataset where each $(x_i,y_i) \sim \clean$ sampled independently. We use $\erm$ to denote the learning algorithm. Accordingly, $\theta := \erm(D)$ is the benign model trained on the clean dataset $D$. 
We call $\mathbb{P}_{(x,y) \sim \clean}\{F(x,\theta)=y\}$ the \emph{clean accuracy~(ACC)} of the classifier $F(\cdot;\theta)$. 

\subsection{Threat Model}
\label{subsec:threat_model}

In this paper, we follow the standard threat model of \textit{backdoor poisoning attacks} \citep{Chen2017TargetedBA}. 

\textbf{Adversary's Capabilities.} Similar to most existing backdoor poisoning attacks, 
the adversary: 
\textbf{1)} can manipulate a limited portion~(no more than a half) of the training dataset, 
\textbf{2)} has access to the victim's model architecture, training algorithm, and potential backdoor mitigation techniques.  
Besides, the adversary has no control over either the training process or the environment where models are deployed. 

\textbf{Notations.} We use $\trigger: \mX \mapsto \mX$ to denote the adversary's trigger planting strategy that plants the backdoor trigger to a data sample, $\labeltrigger: \mX \times [C] \mapsto [C]$ to denote the adversary's labeling function on poison samples, and $t$ to denote the target class. 
The adversary will manipulate $k$ samples in the clean dataset $D$, and we use $\poiind:=\{j_1,...,j_k\}$ to denote indices of the $k$ samples. Then, the resulting poisoned dataset is denoted as $\Dpoiall=\{(\Tilde{x}_i, \Tilde{y}_i)|i=1,\dots,n\}$, where 
\begin{align}
    & \Tilde{x}_i = 
    \begin{cases} \trigger(x_i) , & i \in \poiind\\
        x_i , & \text{o/w}
    \end{cases},
    & \Tilde{y}_i = 
    \begin{cases} \labeltrigger(x_i,y_i)  , & i \in \poiind\\
        y_i , & \text{o/w}
    \end{cases}.\label{eqn:poison_formulation}
\end{align}
For convenience, we also decompose $\Dpoiall := \Dclean  \cup  \Dpoi$, where $\Dpoi := \{(\Tilde{x}_i, \Tilde{y}_i) | i \in \poiind\}$ and $\Dclean := \Dpoiall \setminus \Dpoi$. 
That is, $\Dclean$ corresponds to the portion of clean samples, $\Dpoi$ is the portion of backdoor poison samples. 
The model $\bdmodel = \erm(\Dpoiall)$ trained on the poisoned dataset $\Dpoiall$ is the backdoored model. The quantity $\mathbb{P}_{(x,y)\sim \clean |_{y\ne t}}\{F(\trigger(x), \bdmodel)=t\}$ is the attack success rate~(ASR) on the backdoored model.

\textbf{Attack Goals.} 
The adversary aims to construct a poisoned dataset $\Dpoiall$ such that the trained model $\bdmodel := \erm(\Dpoiall)$ will be backdoored. In general, a desired backdoor poisoning attack satisfies the following conditions~(using notations from Sec~\ref{subsec:notaitons}):
\begin{align}
    & \mathbb{P}_{(x,y)\sim \clean}\Big\{F(x;\bdmodel) = y \Big\} \approx \mathbb{P}_{(x,y)\sim \clean}\Big\{F(x;\theta) = y \Big\},\label{eqn:high_clean_acc} \\
    & \mathbb{P}_{(x,y)\sim \clean|_{y \ne t}}\Big\{F(\trigger(x);\bdmodel) = t \Big\} \ge \thres,\label{eqn:high_asr}
\end{align}
Eqn~\ref{eqn:high_clean_acc} requires the backdoored model $F(\cdot;\bdmodel)$ still maintains a level of clean accuracy~(ACC) that is comparable to that of a benign model $F(\cdot;\theta)$, in order to ensure the stealthiness of the attack. Eqn~\ref{eqn:high_asr}, on the other hand, asserts that the backdoored model should have a high ASR ($>$ some threshold $\thres$).



\subsection{\revision{Detecting Backdoor Poison Samples}}
\label{subsec:goal_and_capability_of_our_defense}

We investigate methods for \textit{detecting backdoor poison samples as a means of defense} against backdoor poisoning attacks. 
We focus on offline detection that aims to identify potential backdoor poison samples within a given dataset. 

\textbf{Defender's Capabilities.} \textbf{1)} Autonomy over the poisoned dataset: After acquiring the poisoned dataset, the defender possesses complete autonomy and freedom to manipulate it. This includes the ability to access, scrutinize, and process the dataset, as well as the liberty to independently train models on it; \revision{\textbf{2)} Access to a small reserved clean set: similar to many prior backdoor defenses~\cite{tang2021demon,liu2017trojaning,liu2018fine,liu2021removing,truong2020systematic,zhao2020bridging,li2021neural}, our defender has access to a small clean set, e.g., as few as 250 to 2000 samples for CIFAR10~(Figure~\ref{fig:ablation_num_clean}). This small set, while insufficient for training a high-accuracy model, is intended to bootstrap defenses. In practice, this small clean dataset can be in-house data directly generated by the defenders themselves (e.g., a defender could gather photos using their own camera and label them) or data collected from trustworthy sources. Alternatively, \citet{zeng2022sift} show the feasibility of sifting out a small clean set directly from a poisoned dataset to bootstrap subsequent defenses.} 



\newcommand{\Dsusp}{D_{\mathrm{suspect}}}


\textbf{Defender's Goals.} The defense intends to isolate a subset $\Dsusp$ from the dataset $\Dpoiall$. We denote $\frac{|\Dsusp \cap \Dpoi|}{|\Dpoi|}$ as the True Positive Rate~(TPR) of the defense and $\frac{|\Dsusp \setminus \Dpoi|}{|\Dclean|}$ as the False Positive Rate~(FPR). The goals of the defense are three-fold: 
\textbf{1)} High TPR: eliminate as many as possible backdoor poison samples; \textbf{2)} Low FPR: the quantity of mistakenly eliminated clean samples should be controlled at a low level; 
\textbf{3)} Generalizability: the defense should be effective against different attacks and should keep robustness across different datasets and attack hyperparameters~(e.g., poison rate). \revision{Note that the defense we consider falls under poison-detection-based defenses~\cite{tran2018spectral,chen2018activationclustering,tang2021demon,hayase21a}, which differs from some other defenses with alternative goals (e.g., robust training, post-training backdoor removal)~\cite{li2021anti,huang2022backdoor,wang2022training,tao2022model, wang2019neural,li2021neural,liu2018fine}. }
%
%



\revision{\textbf{Poison Detection for Backdoor Defenses.} 
Detecting poison samples provides a \emph{flexible} foundation for countering backdoor attacks. \textbf{First}, if one can accurately identify and eliminate backdoor poison samples from a poisoned dataset, the threat of backdoor poisoning attacks can be mitigated at the outset.  {The cleansed dataset can be flexibly utilized to train any clean model, irrespective of its algorithm or architecture.} This flexibility presents advantages over many defenses~(e.g., robust training~\cite{li2021anti,huang2022backdoor,wang2022training,tao2022model}) that are tied to rigid training algorithms and optimized for specific model architectures. For instance, \textit{one has the flexibility to use a lightweight model (e.g., ResNet18) to cleanse a dataset and then use any model architecture (e.g., ViT~\cite{dosovitskiy2020image}) and training algorithms to train an advanced clean model} on the cleansed dataset. \textbf{Second}, poison detection can act as a foundational building block in other backdoor defenses. For example, some defenses~\cite{li2021anti,huang2022backdoor,wang2022training} also rely on isolating (only) a portion of poison samples and strategically employing them to mitigate backdoor learning during training. Thus, \textit{even if the poison detection is imperfect (For instance, TPR is not sufficiently high), it can still be integrated with other techniques to construct effective backdoor defenses}.}

%% file: sections/methodology.tex
\section{Methodological Analysis}
\label{sec:methodology}

\revision{\textbf{To tackle backdoor poison samples detection}\footnote{\revision{Our analysis strictly pertains to the context of detecting backdoor poison samples in poisoned datasets, which is formally indicated in Definitions~\ref{def:the-post-hoc-workflow} and \ref{def:the-proactive-workflow}. Thus, the analysis is not meant to be directly applied to other tracks of defenses ~\cite{li2021anti,huang2022backdoor,wang2022training,tao2022model,wang2019neural,li2021neural,liu2018fine} (that do not intend to accurately separate poison and clean samples) within the broad backdoor defenses literature~\cite{li2022backdoor}. }}, defenders rely on some distinctive characteristics of poison samples, as outlined in Sec~\ref{subsec:define_distinguishable_characteristics}. We analyze this issue methodologically and critique the post-hoc workflow of state-of-the-art approaches in Sec~\ref{subsec:post-hoc-workflow}, highlighting its limitations. In Sec~\ref{subsec:proactive_mindset}, we suggest a proactive mindset as an alternative, based on which we formulate a unified framework and offer practical insights on designing proactive poison detection pipelines.}

\subsection{\revision{Defining Poison Samples Detector}} 
\label{subsec:define_distinguishable_characteristics}

\begin{definition}[{Backdoor Characteristic Function}] A backdoor characteristic function is defined as $\charf(\cdot;\theta): \mX \times [C] \mapsto \mathcal{Z}$, which maps data point $(x,y) \in \mX \times [C]$ to characteristic vectors $\charf(x,y;\theta) \in \mathcal{Z}$, where $\mathcal{Z}$ is a characteristic space that facilitates discrimination between clean and backdoor poison samples, and $\theta$ is the parametrization (if applicable). 
\label{def:backdoor-characteristics-function}
\end{definition}

\begin{definition}[{Backdoor Poison Samples Detector}] A backdoor poison samples detector is the composition of a backdoor characteristic function $\charf(\cdot;\theta)$ and a decision function $\decf: \mathcal{Z} \times [C] \mapsto \{0,1\}$. Given a data entry $(x,y) \in \mX \times [C]$, the decision function $\decf$ takes ($\charf(x,y;\theta)$, y) as its input and predicts whether $(x,y)$ is a poison sample --- $\decf[\charf(x,y;\theta), y]=1$ indicates a positive prediction and $\decf[\charf(x,y;\theta), y]=0$ otherwise. 
\revision{We use $TPR(\charf, \theta, \decf)$ and $FPR(\charf, \theta, \decf)$ to denote the True Positive Rate and False Positive Rate of the poison detector.}
\label{def:detector}
\end{definition}





\vspace{-1.5em}
\subsection{The Post-hoc Workflow and Its Limitations}
\label{subsec:post-hoc-workflow}

\textbf{The Post-hoc Workflow.} We reveal that most prior studies~\cite{chen2018activationclustering,gao2019strip,hayase21a,tran2018spectral,tang2021demon,chou2020sentinet} \textit{on the detection of backdoor poison samples} have employed a post-hoc workflow, where the defender proceeds as follow: \textbf{1)} Allow the attack to first proceed as per the adversary’s expectation by routinely training a backdoored model $\bdmodel := \A(\Dpoiall)$ on the poisoned dataset. \textbf{2)} Apply some presumed characteristics $\charf(\cdot;\bdmodel)$ of the post-attacked model to project $\Dpoiall$ into the characteristic space for analysis. \textbf{3)} Optimize a decision function $\xi$ on $Z:=\{\charf(\Tilde{x},\Tilde{y};\bdmodel)| (\Tilde{x},\Tilde{y}) \in \Dpoiall\}$ with a goal of maximizing TPR while constraining FPR.
\textbf{4)} Apply the decision function $\xi[\charf(\Tilde{x},\Tilde{y};\bdmodel), \Tilde{y}]$ for every sample $(\Tilde{x}, \Tilde{y}) \in \Dpoiall$, and a sample is identified as a potential poison sample if and only if $\xi[\charf(\Tilde{x},\Tilde{y};\bdmodel), \Tilde{y}]=1$. The rationale behind this workflow is \textit{an assumption that post-attacked models will spontaneously exhibit distinctive characteristics} on poison and clean samples, which can be utilized to distinguish between the two populations. In Definition~\ref{def:the-post-hoc-workflow}, we formulate this workflow within an optimization framework.


\begin{definition}[{The Post-hoc Workflow}] \revision{A post-hoc poison detection defender solves the following problem for some (presumed) characteristics $\charf$ of a post-attacked model:
\begin{align*}
    & \widetilde{Q}:= \max_{\xi} \ TPR(\charf, \bdmodel, \decf),  \ \ \ \ \ \ s.t.\ \  FPR(\charf, \bdmodel, \decf)   \le \epsilon, \\
    & \bdmodel := \A(\Dpoiall),
\end{align*}
where the defender aims to optimize the TPR within an FPR budget $\epsilon$, by searching for a decision function $\decf$ w.r.t certain characteristic $\phi$ of the backdoored model $\bdmodel$.}
\label{def:the-post-hoc-workflow}
\end{definition}

\textbf{Failure Modes of Post-hoc Approaches: A Case Study.} The most successful examples of the post-hoc workflow are \textit{latent separation based poison detectors}~\cite{tran2018spectral,chen2018activationclustering,tang2021demon,hayase21a}. These methods operate on the premise that backdoored models will learn separate latent representations for clean and backdoor poison samples. Consequently, the latent representation, denoted as $h(\cdot;\bdmodel)$, is perceived as the backdoor characteristic function for poison detection. The next step involves conducting a clustering analysis in the latent representation space to derive a decision function $\decf$ that separates clean and backdoor poison samples. 
These approaches constitute a state-of-the-art frontier of backdoor defenses, with methods such as Spectral Signature~\cite{tran2018spectral} and Activation Clustering~\cite{chen2018activationclustering} being considered as canonical baselines within the literature, while SCAn~\cite{tang2021demon} and SPECTRE~\cite{hayase21a} have reported nearly perfect results against a diverse set of baseline attacks. These works heuristically optimize the decision function $\xi$ (as described in Definition~\ref{def:the-post-hoc-workflow}), primarily by refining the clustering analysis. However, they are prone to failure or performance degradation in many scenarios where the assumed characteristics are inherently weak or even deliberately suppressed. For instance, latent separation characteristics can be less discernible~\cite{tran2018spectral,hayase21a} when the poison rate is low (Fig~\ref{fig:vis_high_low_poison_rate}). Recently, \citet{tang2021demon} and \citet{qi2023revisiting} also propose adaptive backdoor poisoning attacks that can even intentionally suppress the latent separation (Fig~\ref{fig:vis_adaptive_non_adaptive}). Furthermore, these characteristics can also vary across different datasets (Fig~\ref{fig:vis_cifar10_gtsrb}).



Besides the latent separation based approaches, \citet{gao2019strip} assume that backdoor models' predictions on poison samples have less entropy under intentional perturbation, and \citet{chou2020sentinet} count on abnormal regions in the backdoored model's saliency map. However, they also suffer from similar failure modes. For example, it is well understood that characteristics assumed by the two works do not hold true for many backdoor attacks with non-local triggers~(e.g., \cite{Chen2017TargetedBA,li2021invisible}).

\textbf{Methodological Limitations.} \textit{We posit that this post-hoc workflow is fundamentally limited in that it does not fully exploit defenders' capabilities}. Particularly, this workflow only passively builds detection pipelines based on some post-hoc characteristics that are not within the control of defenders. However, these (presumed) characteristics could often be inherently weak or even deliberately suppressed by attackers.

\subsection{Towards A  Proactive Mindset}
\label{subsec:proactive_mindset}

\textbf{The "Home Field" Advantage.} In this study, we highlight the "home field" advantage held by defenders in the face of backdoor poisoning attacks. This advantage stems from the fact that once attackers have poisoned a training dataset and then subsequently handed it over to defenders, the latter will have \textit{full control and autonomy over both the poisoned dataset and their own operational environment}. Conversely, the attackers will not be able to interfere with any subsequent defense measures taken by the defenders.

We point out that the post-hoc workflow~(Definition~\ref{def:the-post-hoc-workflow}) does not fully exploit this "home field" advantage. It passively allows the attack to first proceed without exerting any intervention, even though it has the capability to do so. As a result, the post-hoc characteristics of the attacked model are completely out of the control of defenders. As we have reviewed in Sec~\ref{subsec:post-hoc-workflow}, these post-hoc characteristics could inherently fail to emerge~\textit{(so we need to proactively enforce them)} and might also be deliberately suppressed by attackers when they design the attack~\textit{(so we should not allow the attack to proceed as per the adversary's expectation)}.

\textbf{A Proactive Mindset: proactively enforce and magnify distinctive characteristics of the post-attacked model to facilitate poison detection.} To alleviate the limitations of the post-hoc workflow, we suggest a paradigm
shift by promoting a proactive mindset. We encourage defenders to fully exploit their "home field" advantage by engaging proactively with the entire model training and detection pipeline. Specifically, we highlight that defenders have the potential to strategically intervene in the attack process (e.g., post-process data samples, manipulate training procedures) in a way that can proactively enforce post-attacked models to exhibit discriminative characteristics (as per the defenders’ expectations), which can be more reliably used to separate clean and poison samples. Formally, this paradigm shift can be concisely expressed by a key adaptation on the prior framework in Defintion~\ref{def:the-post-hoc-workflow}, and we present this new formulation in Definition~\ref{def:the-proactive-workflow} as follows.

\begin{definition}[{The Proactive Mindset}] \revision{A proactive poison detection defender selects a backdoor characteristic function $\charf$, and proactively enforces and magnifies this characteristic in a post-attacked model by solving the following problem:
\begin{align*}
    & Q^*:= \max_{\xi, \mathbfcal{A^*} } \ TPR(\charf, \mathbfcal{\theta^*}, \decf),  \ \ \ \ \ \ s.t.\ \  FPR(\charf, \mathbfcal{\theta^*}, \decf)   \le \epsilon, \\
    & \mathbfcal{\theta^{*}} := \mathbfcal{A^*}(\Dpoiall),
\end{align*}
where $\mathbfcal{A^*}$ is proactively designed by the defender to magnify the intended characteristic to facilitate poison detection.}
\label{def:the-proactive-workflow}
\end{definition}


\begin{corollary}[Nondegradation] For fixed $\phi, \epsilon, \Dpoiall$,
\label{def:nondegradation}
\begin{align*}
    & Q^* \ge \widetilde{Q},
\end{align*}
where $\widetilde{Q}$ is the ideally optimal TPR in Definition~\ref{def:the-post-hoc-workflow}, while $Q^*$ is the counterpart in Definition~\ref{def:the-proactive-workflow}.
\end{corollary}

\textbf{Practical Insights.} For a simple methodological illustration, in Definition~\ref{def:the-proactive-workflow}, we intentionally adopt a very inclusive variable $\A^*$ to express the proactiveness of our proposal. \textit{In practice, $\A^*$ can encapsulate a number of design choices.} A major design choice that we focus on in this work it the training algorithm that generates the post-attacked model. The key insight is that the goal of $\A^*(\Dpoiall)$ is never to find a model that has high accuracy, but a model that exhibits distinguishable characteristics on poison and clean samples. Thus, an empirical risk minimization procedure $\A$ that aims to best fit the poisoned dataset is not necessarily an optimal take. Instead, we suggest defenders proactively search for specialized training algorithms that can generate better $\theta^*$ to facilitate poison detection. The design of confusion training~(see Sec~\ref{sec:confusion_training} for technical details) in this work is an illustration.

Additionally, Definition~\ref{def:the-proactive-workflow} also sheds light on other dimensions, including data post-processing. The utilization of conventional data augmentation techniques, while common, can be deemed as a simplistic take on this dimension, and it has been shown to amplify latent separation between poison and clean samples~\cite{qi2023revisiting}. By explicitly positioning the data post-processing as a design choice, we posit that there might exist more delicate post-processing procedures, which can be utilized to better facilitate poison detection. We encourage future research to investigate what could be an optimal take on this dimension. Besides, the choice of model architecture can also make a difference. For example, against Adap-Blend Attack~\cite{qi2023revisiting} on CIFAR10, the standard version of ResNet18 exhibits much stronger latent separation than that of a tailored version of ResNet20 (see Fig~\ref{fig:vis_resnet18_resnet20}). Later, in Sec~\ref{subsec:ablation_exps}, we also show that model architecture choice can play an important role in the implementation of confusion training.

\input{sections/figs/case_demo}

%% file: sections/figs/case_demo.tex
\begin{figure}
\centering
\begin{subfigure}{0.21\textwidth}
    \centering
    \begin{tabular}{|@{}r@{}|@{}l@{}|} \hline
    \includegraphics[width=0.49\textwidth, valign=b]{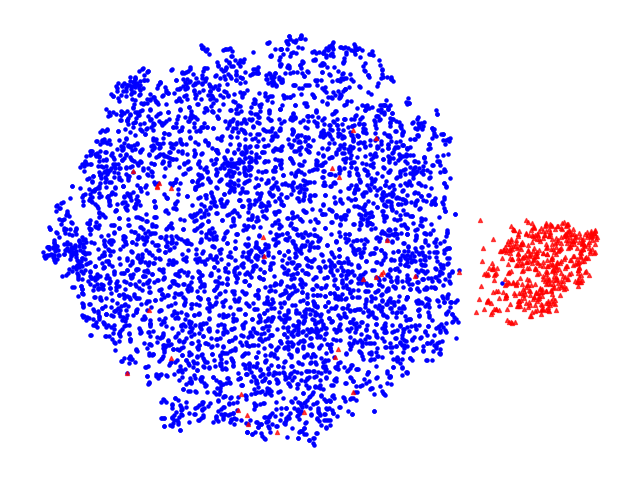} &
    \includegraphics[width=0.49\textwidth, valign=b]{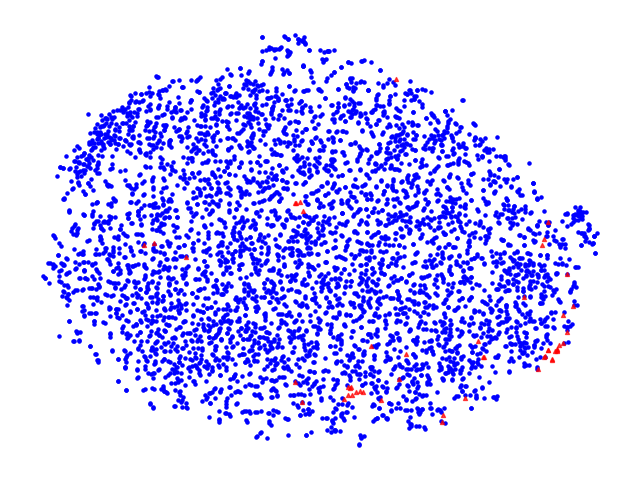} \\ \hline
    \end{tabular}
    \caption{High v.s. low poison rate.}
    \label{fig:vis_high_low_poison_rate}
\end{subfigure}
\hfill
\begin{subfigure}{0.21\textwidth}
    \centering
    \begin{tabular}{|@{}r@{}|@{}l@{}|} \hline
    \includegraphics[width=0.49\textwidth, valign=b]{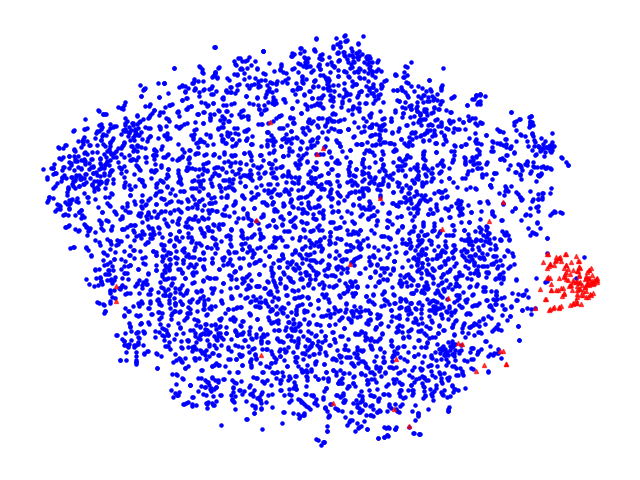} &
    \includegraphics[width=0.49\textwidth, valign=b]{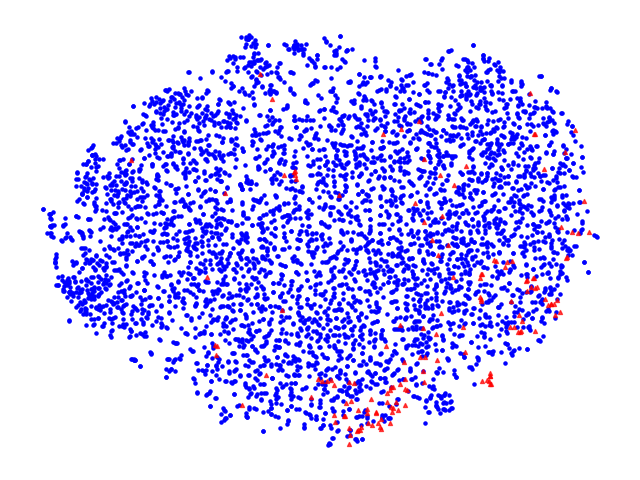} \\ \hline
    \end{tabular}
    \caption{Blend v.s. Adap-Blend.}
    \label{fig:vis_adaptive_non_adaptive}
\end{subfigure}
\\
\begin{subfigure}{0.21\textwidth}
    \centering
    \begin{tabular}{|@{}r@{}|@{}l@{}|} \hline
    \includegraphics[width=0.49\textwidth, valign=b]{sections/imgs/case_demo/tsne_cifar10_blend_0.003_alpha=0.200_trigger=hellokitty_32.png_poison_seed=0_full_base_aug_seed=666.pt_class=0.png} &
    \includegraphics[width=0.49\textwidth, valign=b]{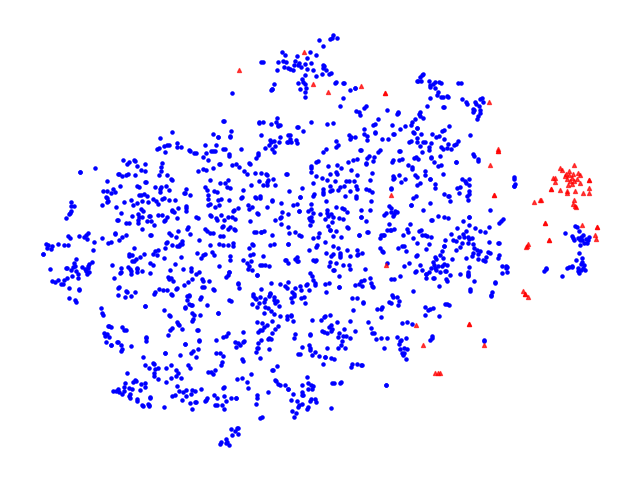} \\ \hline
    \end{tabular}
    \caption{CIFAR10 v.s. GTSRB.}
    \label{fig:vis_cifar10_gtsrb}
\end{subfigure}
\hfill
\begin{subfigure}{0.21\textwidth}
    \centering
    \begin{tabular}{|@{}r@{}|@{}l@{}|} \hline
\includegraphics[width=0.49\textwidth, valign=b]{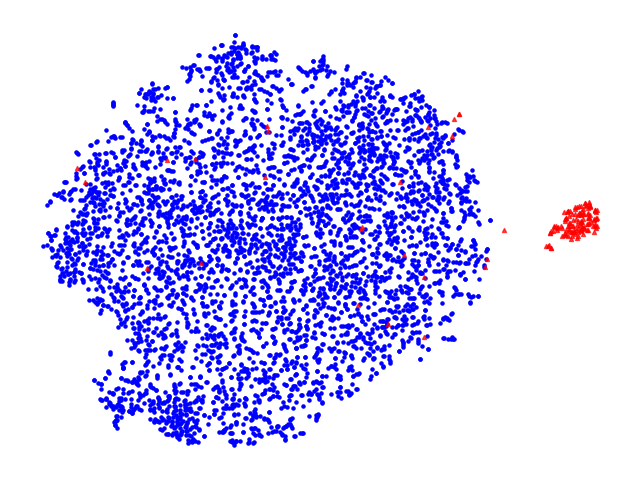} &
    \includegraphics[width=0.49\textwidth, valign=b]{sections/imgs/case_demo/tsne_cifar10_adaptive_blend_0.003_alpha=0.150_cover=0.003_trigger=hellokitty_32.png_poison_seed=0_full_base_aug_seed=999.pt_class=0.png} \\ \hline
    \end{tabular}
    \caption{ResNet18 v.s. ResNet20\footnotemark}
    \label{fig:vis_resnet18_resnet20}
\end{subfigure}
\caption[]{\textbf{Case illustrations.} We visualize the latent representations of the target class training samples. \red{Red} and \blue{blue} points represent poison and clean samples, respectively. Fig~\ref{fig:vis_high_low_poison_rate} compares Blend~\cite{Chen2017TargetedBA} attacks with $1\%$ and $0.1\%$ poison rates. Fig~\ref{fig:vis_adaptive_non_adaptive} shows Blend attack and its adaptive version~\cite{qi2023revisiting} ($0.3\%$ poison rate). Fig~\ref{fig:vis_cifar10_gtsrb} demonstrates the Blend attack on CIFAR10 and GTSRB ($0.3\%$ poison rate). Fig~\ref{fig:vis_resnet18_resnet20} visualizes Adap-Blend attack~\cite{qi2023revisiting} on the standard ResNet18 and the tailored version of ResNet20~\cite{he2016deep} ($0.3\%$ poison rate).}
\label{fig:case_demo}
\end{figure}

\footnotetext{Compared to the standard ResNet18 (11.2M parameters), the tailored version of ResNet20 has much smaller capacity (270K parameters).}

%% file: sections/confusion_training.tex
\section{Confusion Training}
\label{sec:confusion_training}

To showcase how the proactive mindset that we promote in Sec~\ref{subsec:proactive_mindset} may empower the advancement of future research on backdoor poison sample detection, we introduce the technique of Confusion Training (CT) as a concrete instantiation of the proposal. This section presents the design of CT. We start from a high-level overview in Sec~\ref{subsec:ct_overview}, followed by technical details in Sec~\ref{subsec:ct_details}. \revision{To formally illustrate the working principles of  CT, we provide a theoretical analysis of CT in Appendix~\ref{appendix:theory}.}

\subsection{Overview}
\label{subsec:ct_overview}

We sketch the confusion training pipeline in Algorithm~\ref{alg:simplified_sketch_of_confusion_training}, and present a high-level overview of the approach as follows.

\input{sections/assets/confusion_training_algorithm_sketch}

\textbf{The Reserved Clean Set.} The defender initially has a \textit{reserved clean set}~($\Dreserve$) at hand, which consists of a small number of clean samples~(without backdoor triggers) that serve to~(roughly) approximate the clean distribution of $\Dclean$. In practice, $\Dreserve$ is much smaller~(Fig~\ref{fig:ablation_num_clean}) than the training dataset $\Dpoiall$ to an extent that training solely on $\Dreserve$ is not sufficient to generate an accurate model, as stated in Sec~\ref{subsec:goal_and_capability_of_our_defense}. For clarity, in this subsection, we also assume $D_{reserve}$ is correctly labeled, which is consistent with prior arts~\cite{tang2021demon,li2021neural}. In Sec~\ref{subsec:ct_details}, we show this assumption could be removed, and an unlabeled reserved set $\Dreserve$ is sufficient in practice. 

\textbf{Initialization.} We initialize the model with $\A(\Dpoiall)$ in Line~\ref{line:pretrain}, which is the backdoored model routinely trained on the poisoned dataset. This initialization provides a prior of the backdoor and empirically makes confusion training more stable.

\textbf{Confusion Training.} Confusion training proceeds by a joint training on the poisoned dataset $\Dpoiall = \Dclean \cup \Dpoi$ and a set of randomly mislabeled clean samples~(referred to as \underline{\textit{confusion batch}}), as presented in Line~\ref{line:confusion-loss}. 
Specifically, in each iteration of the gradient descent update, a random batch of clean data $(\mathbf{X}_i',\mathbf{Y}_i')$ is sampled from the reserved clean set $\Dreserve$~(Line~\ref{line:clean-batch}). The confusion batch $(\mathbf{X}_i',\mathbf{Y}^*)$ is constructed by reassigning labels (Line~\ref{line:random-mislabel}) to this clean batch with a randomly selected $\mathbf{Y}^*$ different from the ground-truth $\mathbf{Y}_i'$. The model is then trained jointly on the confusion batch $(\mathbf{X}_i',\mathbf{Y}^*)$ and a batch $({\mathbf X}'_i, {\mathbf Y}'_i)$ from the poisoned dataset $\Dpoiall$~(Line~\ref{line:poison-batch}).

Conceptually, the random mislabeling step in Line~\ref{line:random-mislabel} \textit{decouples the benign correlations} between normal semantic features and labels in the confusion batch. In practice, we also assign the confusion batch a large weight $\lambda-1 \textgreater 1$. As a result, \textit{the confusion batch serves as another set of poison samples that strongly obscure the benign correlations, making the clean data points hard to be fitted.} 
On the other hand, since the confusion batch does not contain the backdoor trigger, the backdoor correlation still remains intact. Therefore, at the end of confusion training, \textit{the resulting model $\thetact$ \revision{fails to fit} most clean samples in $\Dclean$ but still correctly fits the set of backdoor poison samples $\Dpoi$}. This allows defenders to distinguish between clean and backdoor poison samples based on the model’s \textit{\underline{state of fitting}} on each sample of $\Dpoiall$.

\textbf{Poison Detection.} In our design, we employ the \textit{"consistency of prediction"} of the post-attacked model $\thetact$ as the indicator to separate poison and clean populations. As formulated in Line~\ref{line:decision-rule}, a training sample $(\Tilde{x},\Tilde{y}) \in \Dpoiall$ is suspected as a backdoor poison sample if and only if the model's prediction $F(\Tilde{x};\thetact)$ on $\Tilde{x}$ is consistent with its label $\Tilde{y}$, i.e., $(\Tilde{x},\Tilde{y})$ is fitted. We note that this is similar to the process of a label-only membership inference~\cite{choquette2021label}, and thus we also call the post-attacked model $\thetact$ as an \textit{{inference model}} in our context. 

\textbf{The Proactive Mindset.} We position confusion training as an illustration of how a proactive defender can directly enforce and magnify distinctive characteristics in post-attacked models to facilitate poison detection. The procedure from Line~\ref{line:pretrain} to Line~\ref{line:end_of_confusion_training} can be deemed as an instantiation of $\A^*$ that we formulate in Definition~\ref{def:the-proactive-workflow}. In Line~\ref{line:start-of-detection}-\ref{line:end-of-detection}, we take the label prediction of the inference model $\phi(\cdot;\thetact):= F(\cdot;\thetact)$ as the backdoor characteristic function and employ $\xi(F(\Tilde{x};\thetact),\Tilde{y}):=\mathbbm{1}(F(\Tilde{x};\thetact)=\Tilde{y})$ as the decision function.


\subsection{Technical Details of The Design}
\label{subsec:ct_details}

Now, we delve into the engineering techniques that are employed to convert Algorithm~\ref{alg:simplified_sketch_of_confusion_training} into a practical implementation.

\textbf{Creating The Confusion Batch.} In practice, we have three technical considerations in the creation of the confusion batch.

\textit{1) Dynamic Labels.} One challenge for implementing confusion training is the overfitting effect of deep learning models. If we use fixed labels for the confusion batch, the inference model $\thetact$ will overfit all samples in both $\Dpoiall$ and the confusion batch by naive memorization. Then, the decision rule in Line~\ref{line:decision-rule} will trivially identify the entire $\Dpoiall$ as positive. To combat overfitting, the random labels $Y^*$ of the confusion batch are regenerated in every iteration~(Line~\ref{line:random-mislabel}). This dynamic labeling endows confusion training with a dynamic nature --- since labels are constantly changing, the model can not be optimized by simply memorizing.

\textit{2) Random {Mis}labeling.} For samples in the confusion batch, we rule out their {\textit{ground truth labels}} during random labeling. This can make the model's fitting on clean samples even worse than a random classifier, and thus helps to reduce the false positive rate~(FPR) of poison detection below $1/C$. 
    
\textit{3) Using unlabeled $\Dreserve$.} Since the ground truth labels of $\Dreserve$ are required for implementing the random mislabeling, we initially assume $\Dreserve$ is labeled in Sec~\ref{subsec:ct_overview}. In practice, we remove this assumption by utilizing the fact that the backdoored model $\bdmodel:=\A(\Dpoiall)$ routinely trained on the poisoned dataset already has a sound clean accuracy in principle. Thus, we directly use the pseudo labels generated by $\bdmodel$ in our implementations. The rationale behind this design is that we do not need $\Dreserve$ to be perfectly labeled in our use case. 
We deem this as another advantage of our method over some prior arts~\cite{tang2021demon,li2021neural} that require accurately labeled clean samples for backdoor defense.

\textbf{Iterative Poison Distillation.} In our practical implementation, we iteratively run confusion training~(Line~\ref{line:pretrain}-\ref{line:end_of_confusion_training}) for multiple rounds. 
In principle, after one round of confusion training, most clean samples in $\Dclean$ will have high loss values due to the \revision{failure of fitting} while backdoor poison samples in $\Dpoi$ that are correctly fitted by $\thetact$ will still concentrate in a region of lower loss values. Thus, after each round of confusion training, we filter out a ratio of samples with the highest loss values and only use the left part for the next round of confusion training. This is analogous to a physical distillation process. By iteratively removing impurities~(clean samples) separated by the previous round of distillation, the density of the desired substances~(backdoor poison samples) will be increasingly higher. Such a process will thus keep amplifying the backdoor correlations~(higher poison rate) meanwhile weakening the benign correlations ~(less clean samples) in the distilled poisoned training set. Consequently, the quality of the resulting inference model will also keep improving.

\textbf{Identify The Target Class.} Many state-of-the-art poison detection approaches~\cite{chen2018activationclustering,tang2021demon,hayase21a} apply clustering analysis to identify potential target classes and only perform poison detection on training samples that are labeled to these classes. This technique helps to reduce the FPR of the detection because it avoids false positives on obviously innocent classes. In our implementation, we also follow this practice by using a Gaussian Mixture Model similar to \citet{tang2021demon}.

We refer readers to Algorithm~\ref{alg:find_target_classes}-\ref{alg:full_version} in Appendix~\ref{appendix_subsec:ct_full_algorithms} for a formal description of these techniques as well as our implementation details.

%% file: sections/assets/confusion_training_algorithm_sketch.tex
\begin{algorithm}[h]
\caption{\mbox{Poison Detection with Confusion Training}}

\textbf{Input:} Dataset $\Dpoiall = \{(\Tilde{x}_i,\Tilde{y}_i)\}_{i=1}^n$ to be cleansed, reserved clean set $\Dreserve$, loss function $\loss$, confusion factor $\lambda$, number of confusion iterations $m$\\
\textbf{Output:} The cleansed dataset $D^*$
\begin{algorithmic}[1]
  \State $\thetact \leftarrow \erm(\Dpoiall)$
  \Comment{Pretrain} \label{line:pretrain}
\For{$i=1,...,m$}\Comment{Confusion Training}
      \State $(\mathbf{\widetilde{X}}_i, \mathbf{\widetilde{Y}}_i) \gets$ a random batch from $\Dpoiall$  \label{line:poison-batch}                 
      \State $({\mathbf X}'_i, {\mathbf Y}'_i) \gets$ a random batch from $\Dreserve$ \label{line:clean-batch}         
      \State ${\mathbf Y}^* \gets$ random mislabels other than ${\mathbf Y}'_i$  \label{line:random-mislabel}  
      \State $\ell_{ct} \gets  \frac{\loss(f(\mathbf{\widetilde{X}}_i;\theta_{ct}), \mathbf{\widetilde{Y}}_i) + (\lambda-1) \cdot  \loss(f({\mathbf X}'_i;\theta_{ct}), \mathbf{Y}^*)}{\lambda}$ \label{line:confusion-loss}  
      \State $\thetact \gets$ one step gradient descent on $\ell_{\text{ct}}$
\EndFor\label{line:end_of_confusion_training}

\State $D^* \gets \Dpoiall$

\For{$i=1,...,n$} \label{line:start-of-detection} \Comment{Poison Detection}
\If{$\Tilde{y}_i == F(\Tilde{x}_i;\theta_{ct})$} \label{line:decision-rule}
\State $D^* \gets$ remove $(\Tilde{x}_i, \Tilde{y}_i)$ from $D^*$
\EndIf
\EndFor \label{line:end-of-detection}
\State \Return $D^*$
\end{algorithmic}
\label{alg:simplified_sketch_of_confusion_training}
\end{algorithm}

%% file: sections/experiments.tex
\input{sections/tables/table_detection.tex}

\section{Empirical Evaluation of Confusion Training}
\label{sec:experiments}

In this section, we present the empirical evaluation of the confusion training~(CT) defense. We discuss the evaluation setup in Sec~\ref{subsec:evaluation_protocols}, followed by Sec~\ref{subsec:consistent_effectiveness_across_settings}, presenting our major results and ablation studies on CIFAR10 and GTSRB. Sec~\ref{subsec:ct_scalability} demonstrates the scalability and generalizability of CT on two large datasets for image classification and malware detection, namely ImageNet and Ember. We discuss additional adaptive attacks against CT in Sec~\ref{subsec:adaptive-attacks}.

\subsection{Evaluation Setup}
\label{subsec:evaluation_protocols}
\label{subsubsec:exp_setup}

\textbf{Datasets.} Our primary results focus on two commonly studied image classification datasets, CIFAR10~\cite{krizhevsky2009learning} and GTSRB~\cite{stallkamp2012man}. 
These datasets are the primary focus of prior works on backdoor attacks and defenses, and their use allows us to perform a thorough comparison with state-of-the-art approaches. We further consider the 1000 classes ImageNet dataset and the Ember dataset for malware classification in Sec~\ref{subsec:ct_scalability}.  

\textbf{Models.} For a consistent evaluation, we use ResNet18~\cite{he2016deep} as the default model architecture for our primary experiments. Later in Sec~\ref{subsec:ablation_exps}, we present additional ablation results on other architectures~(VGG16~\cite{vgg}, MobileNetV2~\cite{sandler2018mobilenetv2}, DenseNet121~\cite{huang2017densely}). 

\textbf{Metrics.} 
We report our evaluation results with two sets of metrics: \textbf{1)} for poison detection defenses, we report the \textit{{true positive rate ({TPR})}} and \textit{{false positive rate ({FPR})}} of the detection~(Table~\ref{tab:main_detection_final}); 
\textbf{2)} for \revision{end-to-end} backdoor defenses, we report the {\textit{clean accuracy ({ACC})}} and {\textit{attack success rate ({ASR})}} of the final model protected by the defense. 
\revision{Moreover, to compare all the different types of defenses, we also evaluate the ACC and ASR for poison detection defenses. As noted earlier in Sec~\ref{subsec:goal_and_capability_of_our_defense}, the end-to-end performance indicators~(ACC and ASR) of poison detection defenses depend on how subsequent models are trained~(e.g., algorithms/architectures) on the cleansed dataset. For a fair and straightforward comparison, these two metrics are measured by directly retraining the same model architecture~(ResNet18 by default) from scratch on the dataset cleansed by the corresponding detection approach.} For rigor of the evaluation, all major experiments are independently repeated three times, and we report the results in the format of "mean (standard deviation)".

\textbf{Attacks.} To validate the generalizability of confusion training, we evaluate it against 11 different types of backdoor poisoning attacks in the literature, including \textbf{1)} classical \textit{dirty label} attacks~(BadNet~\cite{gu2017badnets}, Blend~\cite{Chen2017TargetedBA}, Trojan~\cite{liu2017trojaning}), \textbf{2)} \textit{clean label} attacks~(CL~\cite{turner2019label}, SIG~\cite{barni2019new}), \textbf{3)} attacks with \textit{sample-specific triggers}~(Dynamic~\cite{nguyen2020input}, ISSBA~\cite{li2021invisible}, WaNet~\cite{nguyen2021wanet}), as well as \textbf{4)} \textit{latent space adaptive attacks}~(TaCT~\cite{tang2021demon}, Adap-Blend~\cite{qi2023revisiting}, Adap-Patch~\cite{qi2023revisiting}). During the implementation of these attacks, we strictly followed the protocols and suggested configurations of their original papers and open-source implementations.
On CIFAR10, we implement and evaluate all these attacks. 
On GTSRB, we omit CL and ISSBA since their original papers only release poison sets or models for CIFAR10. In our main evaluation, \textit{we try our best to pick a minimum \textbf{poison rate} (defined as $|\Dpoi| / |\Dpoiall|$) for each attack, maximizing its stealthiness while still maintaining a high ASR}. The rationale behind this choice is that a low poison rate is more realistic and always preferable by attackers in practical scenarios, and attacks with lower poison rates are also harder to  defend --- we choose the setting that is most realistic and challenging. In Sec~\ref{subsec:ablation_exps}, we present ablation results on different poison rates for comprehensiveness. In Appendix~\ref{appendix:detailed_configurations_of_baselines}, we describe the detailed configurations of all the considered attacks.


\textbf{Baseline Defenses.} We compare confusion training with 14 backdoor defenses in the literature: \textbf{1)} To illustrate the superiority of our proactive mindset, we include the \textit{6 post-hoc poison detection approaches}~(SentiNet~\cite{chou2020sentinet}, STRIP~\cite{gao2019strip}, SS~\cite{tran2018spectral}, AC~\cite{chen2018activationclustering}, SCAn~\cite{tang2021demon} and SPECTRE~\cite{hayase21a}) that we analyze in Sec~\ref{subsec:post-hoc-workflow}. \textbf{2)} For comprehensiveness, we also include the frequency based detection method~(Frequency~\cite{zeng2021rethinking}) that directly scans input samples without utilizing post-attacked models' characteristics, presenting it as an outlier that does not fit into the paradigm we discuss. \textbf{3)} We also include \revision{7} other representative backdoor defenses that are not built on poison detection. Specifically, FP~\cite{liu2018fine}, NC~\cite{wang2019neural}, ABL~\cite{li2021anti} and NAD~\cite{li2021neural} are covered in Table~\ref{tab:main_defense}, \revision{while DBD~\cite{huang2022backdoor}, MOTH~\cite{tao2022model} and NONE~\cite{wang2022training} are deferred to Appendix~\ref{appendix:runtime-defenses}}. Hyperparameters of these baseline defenses are optimized following their original papers and open-source implementations. 




\subsection{Effectivness Across Settings: A Benchmark Evaluation on CIFAR10 and GTSRB}
\label{subsec:consistent_effectiveness_across_settings}

\input{sections/tables/table_defense.tex}

We report our major results on CIFAR10 and GTSRB, two standard benchmark datasets in backdoor research. Specifically, we present an overview of the results in \ref{subsubsec:robustness_across_settings}, followed by a thorough analysis in \ref{subsubsec:comparing_with_prior_arts} and ablation studies in \ref{subsec:ablation_exps}.

\subsubsection{Consistent Effectiveness Across Settings}
\label{subsubsec:robustness_across_settings}

We present our main results in Table~\ref{tab:main_detection_final}~(for poison detection based defenses) and Table~\ref{tab:main_defense}~(for all defenses).  We consider a poison detector is \green{\textbf{successful}} against an attack if the TPR is more than $80\%$; otherwise, we say the detector is \red{\textbf{unsuccessful}}. We consider a defense is \green{\textbf{successful}} against an attack only if the ASR is reduced below $20\%$, otherwise \red{\textbf{unsuccessful}}. 

\textbf{TPR and FPR.} According to Table~\ref{tab:main_detection_final}, as a backdoor poison samples detector, {CT is \green{\textbf{successful}} in detecting poison samples across all attacks and datasets we evaluate}. In all settings, CT consistently detects over $90\%$ poison samples~(oftentimes detects $100\%$). In terms of false positives, CT also achieves a low FPR comparable to the best of other detectors. 

\textbf{ASR and ACC.}
According to Table~\ref{tab:main_defense}, as a general backdoor defense, {CT also \green{\textbf{succeeds}} in reducing the ASR in all settings}. As shown, CT consistently reduces the ASR to less than $5\%$. On the other hand, the ACC drop induced by CT defense is moderate -- the ACC is always higher than 92.4\% on CIFAR10 and 96.0\% on GTSRB, in parallel with the best of other defenses. Note that, for a fair comparison, we use the same ResNet18 architecture to retrain models on the cleansed dataset to report ASR and ACC. \textit{In practice, users can freely train models with more advanced architecture and larger capacity on the cleansed dataset to achieve higher accuracy}.


\subsubsection{Strengths of Confusion Training over Prior Arts}
\label{subsubsec:comparing_with_prior_arts}


\textbf{Prior arts are less effective against latent space adaptive attacks.} The three latent space adaptive attacks~(TaCT~\cite{tang2021demon}, Adap-Patch and Adap-Blend~\cite{qi2023revisiting}) constitute the most challenging cases in our evaluation. In these adaptive attacks, not all trigger-planted samples are labeled to the target class in the poisoned dataset. Thus, the backdoor correlations between the trigger and target class are intentionally complicated, the trigger signal becomes less dominant in the prediction of post-attacked models, and the latent separation between poison and clean samples is also suppressed~\cite{qi2023revisiting}.
As shown in Table~\ref{tab:main_defense}, \textit{all of the 11 baseline defenses fail to defend against at least one of the adaptive attacks on at least one dataset}. \textbf{First}, the suppression of latent separation characteristic makes the four post-hoc poison detection methods (SS~\cite{tran2018spectral}, AC~\cite{chen2018activationclustering}, SCAn~\cite{tang2021demon}, SPECTRE~\cite{hayase21a}) built on this characteristic less effective, as seen in Table~\ref{tab:main_detection_final}. \textbf{Second}, complicated backdoor correlations make it harder to fit the backdoor poison samples, reducing the effectiveness of ABL~\cite{li2021anti}, which assumes poison samples will be fitted faster than clean samples. \textbf{Third}, as the backdoor trigger becomes less dominant, approaches like SentiNet~\cite{chou2020sentinet}, Strip~\cite{gao2019strip}, and NC~\cite{wang2019neural} that rely on the dominance of backdoor triggers are also less effective. \textbf{Additionally}, Frequency~\cite{zeng2021rethinking} suffers from degradation against Adap-Blend and Adap-Patch because these attacks use more implicit triggers. NAD~\cite{li2021neural} also performs worse against Adap-Blend on GTSRB, which could be due to the weakened distinguishability between clean and poison populations in latent space, on which NAD performs knowledge distillation. FP~\cite{liu2018fine} in general performs poorly due to its ineffectiveness in our evaluation settings, where poison rates of all attacks are low. 

\textbf{In contrast, CT demonstrates stronger robustness against such adaptive attacks.} As shown in Table~\ref{tab:main_detection_final}, CT still consistently achieves over $90\%$ TPR against all of the three latent space adaptive attacks on both datasets. This suffices to defend against these attacks, as ASRs on all the models trained on the cleansed datasets are reduced to less than $4\%$. 

\textbf{This feat can be attributed to the proactive mindset underlying the design of CT}, which steers clear of the reliance on those post-hoc characteristics used by prior arts. Specifically, CT makes no assumption about the type of triggers or labeling of backdoor poison samples. It neither passively waits for backdoored models to "magically" exhibit some characteristics that will expose backdoor poison samples. Instead, \textit{CT starts from an arguably more fundamental premise that the successful execution of a backdoor poisoning attack must be accompanied by the existence of backdoor correlations and corresponding backdoor poison samples~(deviating from the clean distribution) that can be fitted}. Overall, the CT design aims to 
fit backdoor poison samples~(that we don't have knowledge of) while avoiding fitting samples from the clean distribution~(that we can approximate) by proactively making them.

\textbf{This design also makes CT intrinsically more robust on other dimensions.} Based on our evaluation results in Table~\ref{tab:main_detection_final},\ref{tab:main_defense}, we summarize them as follows. 

\textit{1) CT is effective against clean-label attacks~(CL~\cite{turner2019label}, SIG~\cite{barni2019new}).} Clean label attacks are designed to evade human inspection by avoiding mislabeling any poison samples. They do not impose any challenges on CT as this strategy can not stop CT from fitting the backdoor correlations.

\textit{2) CT is effective against sample-specific triggers~(Dynamic~\cite{nguyen2020input}, WaNet~\cite{nguyen2021wanet}, ISSBA~\cite{li2021invisible}).} Sample-specific attacks use different triggers for different backdoor samples. These backdoor attacks with such diversed triggers lead to the reduced effectiveness of many defenses, including SentiNet, STRIP, Frequency, SPECTRE, NC, ABL, and NAD. Meanwhile, CT still defends against these attacks effectively.

\textit{3) CT is effective against implicit triggers~(Blend~\cite{Chen2017TargetedBA}, SIG~\cite{barni2019new}, WaNet~\cite{nguyen2021wanet}, ISSBA~\cite{li2021invisible}).} There are some attacks that use more implicit triggers for stealthiness. Unlike other attacks that patch a significant trigger over images, these defenses manage to reduce the strength of the trigger signals, making them less noticeable. 
According to our results, no other defenses in our evaluation can steadily succeed against all these attacks on both datasets. However, CT is consistently successful in inhibiting the ASR to $<1.6\%$.

\textit{4) CT is effective across datasets.} Another significant issue with the baseline defenses is that they do not generalize equally across datasets. A notable example is SPECTRE, a post-hoc approach based on latent separation, which performs well on CIFAR10 against all attacks but fails in multiple cases on GTSRB. This is a direct result of the post-hoc workflow reviewed in Sec~\ref{subsec:post-hoc-workflow}, as we previously demonstrated in Fig~\ref{fig:vis_cifar10_gtsrb} that the latent separation characteristics of the post-attacked model could vary across datasets if they are not proactively controlled. As a result, post-hoc approaches are prone to failure. In contrast, the proactive nature of CT makes itself more stable across datasets. The effectiveness of CT on additional datasets will be discussed in Sec~\ref{subsec:ct_scalability}.

\subsubsection{Ablation Studies}
\label{subsec:ablation_exps}

\input{sections/figs/ablation_poison_rate_succint}

Next, we present additional ablation studies on three factors: \textbf{1)} poison rates of attacks, \textbf{2)} model architectures used for CT, and \textbf{3)} size of the reserved clean set for CT. We investigate the impacts of these factors on the effectiveness of CT. For the ablation study on each factor, we only vary this single factor while keeping all other factors consistent with the setup of the main evaluation in Sec~\ref{subsubsec:exp_setup}. All ablation studies are performed on CIFAR10.

\textbf{Poison Rates.} Fig~\ref{fig:ablation_poison_rate_succint} presents our ablation study on the poison rate of different attacks (we only demonstrate the three most challenging attacks here; refer to Appendix~\ref{appendix:ablation_poison_rate} for more results). Specifically, we consider the set of poison rates $\{0.1\%, 0.3\%, 0.5\%, 1\%, 5\%, 10\%\}$ which covers both the high and low poison rate regions. For each setting, we compare CT with the two strongest baseline detectors SPECTRE and SCAn. \textit{CT is consistently effective (TPR$\approx$100\%) and outperforms the baselines in both low and high poison rates regions against most attacks (see Fig~\ref{fig:ablation_poison_rate} in Appendix).} Even for three attacks in Fig~\ref{fig:ablation_poison_rate_succint} on which most other defenses fail, CT still achieves TPR$\approx$100\% at most time, beyond the reach of the two baselines. Though the TPR of CT slightly drops in the ultra-low poison rate of $0.1\%$ against Adap-Patch and Adap-Blend, it is still noticeably better than the baselines.

\input{sections/figs/ablation_arch.tex}

\textbf{Model Architectures.} In addition to ResNet18~\cite{he2016deep}, which is used in the main evaluation, we also consider three other model architectures, VGG16~\cite{vgg}, MobileNetV2~\cite{sandler2018mobilenetv2}, and DenseNet121~\cite{huang2017densely}.
For a fair comparison, the hyperparameters are optimized for each architecture. The ablation results on the model architectures are shown in Fig~\ref{fig:ablation_arch}. \textit{The key takeaway is that some architectures are more suitable for implementing CT than others.} As shown, CT with ResNet18 and MobileNetV2 are stable across different attacks, achieving high TPR (close to $100\%$ in most cases) while maintaining an FPR below $5\%$. In contrast, VGG16 and DenseNet121 are less stable, with large variations in performance and frequent failures. According to our empirical experiences, this is because the dynamics of confusion training require easy-to-train architectures. VGG16 does not use skip links, while DenseNet121 is too deep, and thus they are more difficult to train and less stable for CT. This observation is consistent with our practical insights in Sec~\ref{subsec:proactive_mindset} and sheds light on a future research direction where specialized model architectures can be designed to facilitate poison detection.

\input{sections/figs/ablation_clean_num.tex}

\textbf{Size of The Reserved Clean Set.} CT requires a small reserved clean set to launch. Our default implementation in the main evaluation uses a clean set of 2000 samples. We now investigate how the effectiveness of CT varies when the size of the reserved clean set becomes smaller. Specifically, we experiment with fewer clean samples of $\{1000, 500, 250\}$ respectively and present the results~(TPR and FPR) of the position detection in Fig~\ref{fig:ablation_num_clean}. Overall, the TPR of our CT detection pipeline maintains a relatively robust performance, though the TPR would become worse against the two latent space adaptive attacks~(Adap-Patch and Adap-Blend). However, a noticeable trade-off in the FPR is evident. Remarkably, when the quantity of reserved clean samples is reduced to 250, the FPR can, at times, exceed $20\%$, leading to a substantial loss of training data. Intuitively, this is due to the fact that a smaller clean set approximates the clean distribution worse. Thus the confusion batch is also doing worse in decoupling the benign correlations --- many clean samples could still be fitted, leading to higher FPR. The impact of the high FPR on overall defense performance (ACC and ASR) is also evaluated, focusing on the extreme case where only 250 clean samples are used to bootstrap CT for CIFAR10. Even with over $20\%$ of the CIFAR10 training set discarded, CT still achieved an impressive accuracy exceeding $91\%$ in all cases, thanks to the maintained high TPR, which ensured a low Attack Success Rate (ASR). This suggests that for larger datasets, a high FPR might be tolerable during dataset cleansing, as the remaining data can still be used to train a high-quality model. A closely relevant research topic to this observation is Dataset Pruning~\cite{paul2021deep}, which shows that many training data can be removed from the dataset while models trained on the remaining dataset can still be accurate.




\input{sections/tables/table_few_clean}

\subsection{Scalability and Generalizability}
\label{subsec:ct_scalability}



\textbf{Generalization to Domain Other Than Computer Vision.} CT only assumes backdoor poison samples could be fitted after the benign correlations have been decoupled, without constraining the modality it works on. This suggests the potential generalizability of CT to different domains other than computer vision. To illustrate this, we evaluate CT on the Ember~\cite{anderson2018ember} training dataset, a malware classification dataset that consists of 600K feature vectors extracted from Portable Executable (PE) files. Specifically, we consider the attacks proposed by \citet{severi2021explanation}. We use the same EmberNN architecture used in \citet{severi2021explanation} to build models and also follow the same configurations. We consider both the constrained and unconstrained attacks of \citet{severi2021explanation} with $1\%$ poison arte. The constrained attack confines backdoor triggers to those features that are editable in PE files, while the unconstrained attack removes this constraint. Table~\ref{tab:ember} presents our results, validating the effectiveness of CT.

\textbf{Scale to Realistic Settings.} To study the scalability to more realistic settings we will encounter in the real world, we also evaluate CT on the ImageNet-1k training set, consisting of 1.3M images~(26x larger than CIFAR10) with high resolutions. We adapt BadNet~\cite{gu2017badnets}, Trojan~\cite{liu2017trojaning}, and Blend~\cite{Chen2017TargetedBA} with a $1\%$ poison rate to implement attacks on ImageNet. We use ResNet18~\cite{he2016deep} for both confusion training on the poisoned dataset and subsequent retraining on the cleansed dataset. Table~\ref{tab:imagenet} presents the results of CT defense against the attacks. As shown, CT still remains effective on ImageNet --- with nearly perfect $TPR>99\%$ and negligible $\mathbf{FPR\approx 0.1}$, models trained on cleansed sets reduce $ASR$ to 0 without suffering any drop of ACC. \revision{An intriguing observation is that the FPR here is even lower than we get for CIFAR10 and GTSRB. We hypothesize that this is because when the size of the dataset is larger and the clean task is more complicated, it also becomes harder for the inference model to fit (memorize) clean samples under the disruption of confusion training, making it even easier to separate clean and poison samples under CT.}

\revision{Additionally, as previously noted in Sec~\ref{subsec:goal_and_capability_of_our_defense}, after we use ResNet18 to cleanse the dataset, we have the flexibility to use any model architecture and algorithm for subsequent training on the cleansed dataset. We confirm this \textbf{by replacing ResNet18 (11 M parameters) with ViT-B/16~\cite{dosovitskiy2020image} (150 M parameters) and applying the state-of-the-art training algorithm GSAM~\cite{zhuang2022surrogate}} during the retraining phase for cases we evaluate in Table~\ref{tab:imagenet}. {As shown in Table~\ref{tab:retrain-vit}, we consistently get clean models with 0 ASR, but the ACC increases to $80\%$ rather than $67\%$ in Table~\ref{tab:imagenet}.} Symmetrically, this also indicates that, in realistic scenarios involving training large models, it is possible to initially deploy CT with a lightweight inference model to sanitize the dataset. The cost would be acceptable, as the computation demanded to train a lightweight model for dataset cleansing can be negligible compared with subsequent training of large models on the cleansed dataset.}



\input{sections/tables/table_ember.tex}
\input{sections/tables/table_imagenet.tex}

\input{sections/tables/retrain_ViT}






\subsection{Additional Analysis on Adaptive Attacks}
\label{subsec:adaptive-attacks}

\textbf{Enforce dependency between benign and backdoor correlations.} CT makes very few assumptions about the configurations of backdoor poisoning attacks, but \textit{it does assume that the backdoor correlations can still be fitted after the benign correlations have been decoupled}. Then, a principled class of adaptive attacks against CT can be those attacks where the backdoor correlations depend on benign correlations, making it more difficult to fit backdoor poison samples after the benign correlations have been decoupled. Some attacks that we evaluate in Table~\ref{tab:main_detection_final},\ref{tab:main_defense} already fall in this class, including \textbf{1)} sample-specific attacks where triggers depend on clean semantics of the samples and \textbf{2)} latent space adaptive attacks where trigger planted samples are conditionally labeled to the target class based on the semantic class~(e.g. TaCT~\cite{tang2021demon}). Results in Table~\ref{tab:main_detection_final},\ref{tab:main_defense}, and Fig~\ref{fig:ablation_poison_rate} illustrate that CT still exhibits good robustness. Additionally, we add an evaluation on an all-to-all attack~\cite{gu2017badnets} with $1\%$ poison rate on CIFAR10, where trigger-planted samples from class $i$ should be misclassified to class $i+1$. This clearly imposes a dependency between the backdoor and clean tasks. Against this attack, CT still achieves $86.6\%$ TPR and $2.2\%$ FPR, \revision{and retraining a ResNet18 on the cleansed dataset results in $5.7\%$ ASR and $93.3\%$ ACC.} We hypothesize that this is because backdoor poison samples are always out of the distribution of clean data and thus can still be fitted though the fitting on clean distribution is disrupted.

\textbf{Launch attack without artifacts.} Most existing attacks unavoidably introduce artifacts into poison samples as they inject trigger patterns that significantly deviate from the clean distribution. These artifacts make poison samples easy to be fitted by CT. This explains why CT is consistently effective against different attacks. Thus, another angle to adaptively attack CT is to launch attacks without significant artifacts. One candidate is the rotation attack~\cite{wu2022just}, which uses image rotation as the backdoor trigger without introducing additional artifacts. We implement this attack following the original paper's configuration with $1\%$ poison rate. Still, CT can achieve $83.6\%$ TPR and $2.8\%$ FPR, \revision{and retraining a ResNet18 on the cleansed dataset will have $7.2\%$ ASR and $92.9\%$ ACC.}

%% file: sections/tables/table_detection.tex
\begin{table*}[ht]
\caption{
\textbf{Comparison of CT with existing backdoor poison samples detectors.} We report the true positive rate (TPR) and false positive rate (FPR) of the poison detection in the format of "mean (standard deviation)" over three independent experiments.} 
\aboverulesep=0.1ex
\belowrulesep=0ex 
\resizebox{1.0\linewidth}{!}{ 
\begin{tabular}{l|ccccccccccccccccc}
\toprule\rule{0pt}{1.0EM}
($\%$)  &  mean~(std) &
\multicolumn{2}{c}{SentiNet}  &
\multicolumn{2}{c}{STRIP}  &
\multicolumn{2}{c}{SS} & \multicolumn{2}{c}{AC} &
\multicolumn{2}{c}{Frequency} &
\multicolumn{2}{c}{SCAn} &\multicolumn{2}{c}{SPECTRE}  & \multicolumn{2}{c}{CT~(ours)}  \\ 

\cmidrule(lr){3-4}  \cmidrule(lr){5-6} \cmidrule(lr){7-8} \cmidrule(lr){9-10} \cmidrule(lr){11-12}  \cmidrule(lr){13-14} \cmidrule(lr){15-16} \cmidrule(lr){17-18} \rule{0pt}{1.0EM}
& & TPR & FPR & TPR & FPR & TPR & FPR & TPR & FPR & TPR & FPR & TPR & FPR & TPR & FPR & TPR & FPR \\
\midrule\rule{0pt}{1.0EM}

\multirow{10}*{\shortstack{C\\I\\F\\A\\R\\10}} 
& No Poison & \gc - & 5.0 (0) & \gc - & 10.3 (0.2) &  \gc - & 15.0 (0) & \gc - & 3.5 (1.0) & \gc - & 0.8 (0) & \gc - & 2.4 (0.6)  & \gc - & 7.5 (0) & \gc - & 2.2 (0.6) \\
\cline{2-18}\rule{0pt}{1.0EM}
& BadNet & \grc 100 (0) & 5.0 (0) & \grc 100 (0) & 10.1 (0.7) & \grc 100 (0) & 4.2 (0) & \grc 100 (0) & 2.5 (0.1) & \grc 100 (0) & 0.8 (0) & \grc 100 (0) & 1.5 (1.1)  & \grc 100 (0) & 2.0 (0) & \grc 100 (0) & 1.0 (0.8)\\
\cline{2-18}\rule{0pt}{1.0EM}
& Blend & \pc 2.2 (0.8) & 5.0 (0) & \pc 14.0 (6.3) & 10.0 (0.6) & \grc 91.8 (7.4) & 4.2 (0) & \pc 65.6 (46.4) & 3.6 (1.6) & \grc 90.7 (0) & 0.8 (0) & \grc 93.1 (0.6) & 0 (0) & \grc 100 (0) & 2.0 (0) & \grc 100 (0) & 1.6 (0.8)  \\
\cline{2-18}\rule{0pt}{1.0EM}
& Trojan & \grc 100 (0) & 5.0 (0) & \grc 100 (0) & 10.6 (0.5) & \pc 18.0 (10.7) & 4.5 (0.1) & \pc 0 (0) & 3.5 (3.9) & \grc 100 (0) & 0.8 (0) & \grc 100 (0) & 1.9 (2.0) & \grc 100 (0) & 2.0 (0) & \grc 100 (0) & 1.8 (0.5) \\
\cline{2-18}\rule{0pt}{1.0EM}
& CL & \pc 2.0 (2.9) & 5.0 (0) & \grc 100 (0) & 10.3 (0.3) & \pc 58.7 (33.8) & 4.3 (0.1) & \pc 0 (0) & 5.6 (2.7) & \grc 100 (0) & 0.8 (0) & \grc 100 (0) & 1.9 (1.6)  & \grc 100 (0) & 2.0 (0) &  \grc 100 (0) & 0.7 (0.4)  \\
\cline{2-18}\rule{0pt}{1.0EM}
& SIG & \pc 34.0 (57.1) & 5.0 (0) & \grc 97.5 (1.5) & 9.5 (0.2) &  \grc 99.0 (0.8) & 28.6 (0) &  \grc 99.5 (0.6) & 2.6 (2.0) & \pc 41.0 (0) & 0.7 (0) & \grc 98.0 (0.9) & 0 (0)  & \grc 99.0 (0.7) & 13.3 (0) & \grc 100 (0) & 0.3 (0.2)  \\
\cline{2-18}\rule{0pt}{1.0EM}
& Dynamic & \grc 100 (0) & 5.0 (0) & \grc 99.8 (0.3) & 10.0 (0.3) & \pc 68.0 (21.5) & 4.3 (0.1) & \pc 0 (0) & 4.8 (1.3) & \grc 99.3 (0) & 0.8 (0) & \grc 96.0 (1.1) & 0 (0)  & \grc 100 (0) & 2.0 (0) & \grc 99.8 (0.3) & 1.4 (0.6)  \\
\cline{2-18}\rule{0pt}{1.0EM}
& ISSBA & \pc 9.7 (6.7) & 5.0 (0) & \pc 67.4 (45.8) & 10.0 (0.1) & \grc 94.7 (7.4) & 28.7 (0.1) & \grc 93.1 (9.7) & 3.1 (2.6) & \grc 80.3 (0) & 0.8 (0) & \grc 92.4 (10.5) & 0 (0)  & \grc 92.8 (10.1) & 9.3 (5.6) & \grc 100 (0) & 0.7 (0.5)\\
\cline{2-18}\rule{0pt}{1.0EM}
& WaNet & \pc 2.3 (0.5) & 5.0 (0) & \pc 4.3 (0.2) & 9.7 (0.8) & \grc 87.2 (0.7) & 48.0 (0.1) & \grc 94.8 (0.3) & 3.8 (1.5) & \pc 6.1 (0) & 10.8 (0) & \grc 87.7 (2.8) & 0 (0) & \grc 88.7 (3.4) & 22.7 (0.2) & \grc 96.5 (0.6) & 0.3 (0.2) \\
\cline{2-18}\rule{0pt}{1.0EM}
& TaCT & \grc 100 (0) & 5.0 (0) & \pc 50.7 (6.8) & 9.8 (0.2) & \pc 25.3 (28.3) & 4.4 (0.1) & \pc 0 (0) & 4.5 (2.8) & \grc 100 (0) & 1.1 (0) & \pc 0 (0) & 1.6 (1.2)  & \grc 100 (0) & 2.0 (0) & \grc 100 (0) & 1.5 (1.1) \\
\cline{2-18}\rule{0pt}{1.0EM}
& Adap-Patch & \grc 94.4 (2.8) & 5.0 (0) & \pc 0.5 (0.6) & 10.4 (0.2) & \pc 1.1 (0.6) & 4.5 (0) & \pc 0 (0) & 2.8 (0.2) & \pc 65.3 (0) & 1.2 (0) & \pc 0 (0) & 2.0 (0.4)  & \pc 77.8 (0.8) & 2.0 (0) &  \grc 96.0 (2.4) & 2.8 (0.4) \\
\cline{2-18}\rule{0pt}{1.0EM}
& Adap-Blend & \pc 1.4 (1.2) & 5.0 (0) & \pc 0.9 (1.3) & 9.9 (0.7) & \pc 51.6 (30.4) & 4.4 (0.1) & \pc 0 (0) & 3.6 (1.4) & \pc 78.0 (0) & 1.0 (0) & \pc 0 (0) & 1.6 (1.1)  & \grc 90.2 (2.3) & 2.0 (0) & \grc 96.2 (4.0) & 3.5 (0.7) \\

\bottomrule
\toprule
\rule{0pt}{1.0EM}

\multirow{8}*{\shortstack{G\\T\\S\\R\\B}}
& No Poison & \gc - & 5.0 (0) & \gc - &  11.0 (0.4) & \gc - & 39.1 (0) & \gc - & 7.1 (1.1) & \gc - & 34.4 (0) & \gc - & 2.1 (1.3)  & \gc - & 3.1 (0.1) & \gc - & 3.3 (1.1)  \\
\cline{2-18}\rule{0pt}{1.0EM}
&  BadNet & \grc 100 (0) & 5.0 (0) & \grc 99.7 (0.5) & 10.8 (0.6) & \grc 98.0 (2.3) & 38.3 (0) & \grc 97.0 (1.9) & 7.1 (1.4) & \grc 100 (0) & 34.3 (0) & \grc 99.0 (1.0) & 0.9 (1.2)  & \grc 98.4 (2.3) & 7.1 (0.5) & \grc 100 (0) & 0 (0) \\
\cline{2-18}\rule{0pt}{1.0EM}
&  Blend & \pc 42.2 (20.2) & 5.0 (0) & \pc 58.4 (39.5) & 11.2 (0.8) & \grc 93.7 (3.5) & 38.4 (0) & \grc 95.6 (1.4) & 8.9 (0.7) & \grc 93.6 (0) & 34.3 (0) & \grc 95.5 (2.0) & 2.3 (0.5)  & \grc 98.7 (1.0) & 8.1 (1.2) & \grc 100 (0) & 0.4 (0.2) \\
\cline{2-18}\rule{0pt}{1.0EM}
& Trojan & \grc 100 (0) & 5.0 (0) & \grc 99.9 (0.2) & 10.1 (0.7) & \grc 97.9 (2.1) & 38.3 (0.1) & \grc 98.7 (1.5) & 5.8 (0.6) & \grc 100 (0) & 34.3 (0) & \grc 100 (0) & 0.2 (0.4) & \grc 100 (0) & 6.5 (0.3) & \grc 100 (0) & 0.5 (0.2) \\
\cline{2-18}\rule{0pt}{1.0EM}
&  SIG & \pc 0.1 (0.1) & 5.0 (0) & \pc 51.6 (13.0) & 11.3 (0.3) & \pc 61.2 (1.8) & 49.8 (0) & \grc 80.2 (7.9) & 8.1 (2.0) & \pc 63.5 (0) & 34.1 (0) & \pc 64.3 (11.4) & 1.0 (1.4)  & \pc 39.1 (27.6) & 8.5 (1.2) & \grc 100 (0) & 0.5 (0.1) \\
\cline{2-18}\rule{0pt}{1.0EM}
&  Dynamic & \pc 79.7 (9.1) & 5.0 (0) & \grc 96.9 (3.9) & 11.3 (0.3) & \grc 91.6 (4.3) & 17.8 (0) & \grc 86.6 (4.2) & 8.1 (2.9) & \grc 84.8 (0) & 34.3 (0) & \pc 78.5 (6.3) & 1.0 (1.4)  & \grc 98.3 (2.4) & 3.0 (0) & \grc 99.2 (1.2) & 1.8 (1.5)  \\
\cline{2-18}\rule{0pt}{1.0EM}
& WaNet & \pc 2.3 (0.9) & 5.0 (0) & \pc 9.5 (2.3) & 10.1 (1.1) & \pc 54.0 (1.7) & 49.7 (0.1) & \pc 0 (0) & 3.3 (1.2) & \pc 33.9 (0) & 40.7 (0) & \grc 93.1 (6.9) & 1.6 (1.4) & \pc 0 (0) & 10.2 (1.2) & \grc 99.6 (0.3) & 0.6 (0.4) \\
\cline{2-18}\rule{0pt}{1.0EM}
& TaCT & \pc 22.1 (38.2) & 5.0 (0) & \pc 35.9 (5.8) & 11.6 (0.6) & \grc 98.7 (0.3) & 25.3 (0) & \grc 99.5 (0.4) & 9.6 (1.9) & \grc 100 (0) & 34.7 (0) & \grc 100 (0) & 2.6 (1.4)  & \pc 0 (0) & 5.5 (0) & \grc 98.5 (1.6) & 2.4 (0.9) \\
\cline{2-18}\rule{0pt}{1.0EM}
& Adap-Patch & \grc 100 (0) & 5.0 (0) & \pc 3.0 (1.2) & 11.6 (0.2) & \pc 64.7 (3.2) & 25.4 (0) & \pc 0 (0) & 4.1 (1.4) & \pc 69.2 (0) & 34.6 (0) & \pc 0 (0) & 0.8 (1.1)  & \pc 0 (0) & 3.7 (0.3) & \grc 94.2 (2.2) & 2.6 (1.2) \\
\cline{2-18}\rule{0pt}{1.0EM}
& Adap-Blend & \pc 1.7 (0.7) & 5.0 (0) & \pc 2.6 (1.8) & 11.1 (0.8) & \pc 43.4 (6.3) & 17.9 (0) & \pc 0 (0) & 6.9 (1.3) & \pc 79.7 (0) & 34.4 (0) & \pc 0 (0) & 0.8 (1.1) & \pc 0 (0) & 3.5 (0.1) & \grc 98.3 (1.6) & 1.6 (0.5) \\
\bottomrule
\end{tabular}
}
\label{tab:main_detection_final}
\end{table*}

%% file: sections/tables/table_defense.tex
\begin{table*}[h]
\caption{
\textbf{Evaluation of CT's effectiveness as a backdoor defense.} We report the defended models' clean accuracy (ACC) and attack success rate (ASR) in the format of "mean (standard deviation)" over three independent experiments. } 
\aboverulesep=0.1ex
\belowrulesep=0ex 
\resizebox{\linewidth}{!}{
\begin{tabular}{l|cccccccccccccccc}
\toprule\rule{0pt}{1.0EM}
  ($\%$)  &  mean~(std) &
\multicolumn{1}{c}{} &
\multicolumn{1}{c}{No Defense} & 
\multicolumn{1}{c}{SentiNet}  &
\multicolumn{1}{c}{STRIP}  &
\multicolumn{1}{c}{SS} & \multicolumn{1}{c}{AC} &
\multicolumn{1}{c}{Frequency} &
\multicolumn{1}{c}{SCAn} &
\multicolumn{1}{c}{SPECTRE} &
\multicolumn{1}{c}{FP} &
\multicolumn{1}{c}{NC} &
\multicolumn{1}{c}{ABL} &
\multicolumn{1}{c}{NAD} &
\multicolumn{1}{c}{CT~(ours)}  \\ 
\midrule\rule{0pt}{1.0EM}

\multirow{22}*{\shortstack{C\\I\\F\\A\\R\\10}} 
& \multirow{2}*{\shortstack{No Poison}} &  ACC &  94.1 (0.1) & 93.4 (0.6) &  92.8 (0.1) &  91.3 (0.1) &  91.6 (0.4) & 93.7 (0.2) &  92.8 (0.1) &  92.0 (0.1) &  82.9 (1.1) & 93.5 (0.7)  &  91.6 (2.5) &  87.4 (0.6) &  92.2 (0.3)   \\
& &  ASR  & \gc - & \gc - & \gc - & \gc - & \gc - & \gc - & \gc - & \gc - & \gc - & \gc -  & \gc - & \gc - & \gc -   \\
\cline{2-16}\rule{0pt}{1.0EM}

& \multirow{2}*{\shortstack{BadNet}} &  ACC &  93.8 (0.1) & 93.2 (0.2) &  92.7 (0.2) &  92.8 (0.1) &  91.7 (0.3) & 93.3 (0.4) &  92.7 (0.2) &  93.2 (0.2) &  83.5 (0.4) &  93.7 (0.2)  & 90.9 (0.5) & 86.8 (1.3) &  93.2 (0.1)   \\
& & ASR & \pc 100 (0) & \grc 0.7 (0.1) & \grc 0.8 (0.1) & \grc 0.6 (0) & \grc 0.8 (0.2) & \grc 0.7 (0.1) & \grc 0.8 (0.1) & \grc 0.8 (0.1) & \pc 100 (0) & \grc 1.0 (0.7) & \grc 7.7 (5.8) & \grc 2.7 (1.6) & \grc 0.8 (0.2)  \\
\cline{2-16}\rule{0pt}{1.0EM}

& \multirow{2}*{\shortstack{Blend}} &  ACC & 93.6 (0.1) & 93.1 (0.5) &  92.8 (0.3) &  92.4 (0.3) &  90.7 (2.0) & 93.2 (0.3) &  93.1 (0.2) &  93.1 (0.2) & 83.6 (0.1) & 93.1 (0.6) & 90.5 (0.9) & 82.3 (6.9) &  93.1 (0.2)   \\
& & ASR & \pc 93.5 (0.2) & \pc 90.7 (1.3) & \pc 88.5 (0.3) & \grc 7.3 (3.1) & \pc 31.2 (41.3) & \pc 33.7 (5.4) & \grc 7.1 (1.5) & \grc 2.4 (0.1) & \pc 85.2 (8.6) & \pc 63.2 (44.1)  & \grc 6.9 (5.2)  & \grc 11.7 (2.7) & \grc 1.6 (0.3)   \\
\cline{2-16}\rule{0pt}{1.0EM}

& \multirow{2}*{\shortstack{Trojan}} & ACC & 93.8 (0.1) & 93.2 (0.1) & 93.0 (0.6) & 92.7 (0.1) & 91.7 (1.9) & 93.3 (0.4) & 93.1 (0.7) & 93.1 (0.4) & 81.1 (2.4) & 93.3 (0.2)  & 92.4 (1.7) & 86.4 (0.5) & 92.9 (0.1) \\
& & ASR & \pc 99.9 (0.1) & \grc 1.6 (0.1) & \grc 2.4 (1.1) & \pc 99.8 (0.1) & \pc 1.7 (0.2) & \grc 2.3 (0.4) & \grc 1.7 (0.5) & \grc 1.7 (0.2) & \pc 71.6 (41.2) & \grc 1.0 (0.5)  & \grc 3.3 (3.2) & \grc 13.0 (2.7) & \grc 1.7 (0.4) \\
\cline{2-16}\rule{0pt}{1.0EM}

& \multirow{2}*{\shortstack{CL}} &  ACC &  93.6 (0) & 92.8 (0.4) &  93.0 (0) &  92.7 (0.2) &  90.4 (2.7) & 93.4 (0.1) &  92.8 (0.1) &  93.0 (0.1) & 82.1 (1.8) & 93.5 (0.2)  & 81.7 (3.0) & 86.4 (1.5) &  93.5 (0.2)   \\
& & ASR & \pc 99.8 (0.1) & \pc 99.6 (0.6) & \grc 1.4 (0.2) & \pc 54.2 (39.8) & \pc 97.9 (1.7) & \grc 1.5 (0.5) & \grc 1.2 (0.4) & \grc 1.6 (0.2) & \pc 94.3 (3.5) &  \grc 1.0 (0.4)  & \grc 13.8 (8.7) & \grc 12.0 (10.5) & \grc 0.9 (0)   \\
\cline{2-16}\rule{0pt}{1.0EM}

& \multirow{2}*{\shortstack{SIG}} &  ACC &  93.8 (0) & 92.6 (0.6) &  92.6 (0.1) &  90.1 (0.8) &  90.3 (0) & 93.3 (0.2) &  92.9 (0.2) &  89.7 (0.2) &  82.8 (1.1) & 93.7 (0.4) & 87.5 (1.8) & 86.3 (0.9) &  93.2 (0)   \\
& & ASR & \pc 80.2 (0.6) & \pc 81.2 (0.7) & \grc 10.1 (16.7) & \grc 0.8 (0.2) & \grc 0.1 (0) & \pc 67.3 (2.9) & \grc 1.1 (0.8) & \grc 0.8 (0.2) & \pc 56.9 (28.5) & \pc 82.7 (0.8) & \pc 29.0 (31.9) & \grc 2.3 (1.1) & \grc 0.1 (0)   \\
\cline{2-16}\rule{0pt}{1.0EM}

& \multirow{2}*{\shortstack{Dynamic}} &  ACC &  93.8 (0) & 92.8 (0.4) &  93.0 (0) & 92.6 (0.1) &  92.9 (0.9) & 93.3 (0.2) &  93.4 (0.2) &  93.2 (0.1) &  82.2 (1.2) &  93.3 (0.2)  &  89.5 (5.0) &  84.3 (3.2) & 93.2 (0)   \\
& & ASR & \pc 99.3 (0.4) & \grc 4.6 (1.0) & \grc 4.9 (0.6) & \pc 60.3 (38.2)  & \pc 98.4 (0.4) & \grc 6.8 (1.1) & \grc 9.6 (2.5) & \grc 4.8 (1.0) & \pc 95.0 (7.9) & \grc 5.1 (2.3)  & \pc 95.8 (5.8) & \pc 35.1 (23.5) & \grc 3.7 (1.6)   \\
\cline{2-16}\rule{0pt}{1.0EM}

& \multirow{2}*{\shortstack{ISSBA}} &  ACC & 93.7 (0) & 92.3 (0.7) &  93.0 (0.2) &  90.5 (0.2) &  92.5 (0.5) & 92.8 (0.4) &  93.2 (0.2) &  89.8 (0) & 83.4 (0.3) &  93.5 (0.3)  & 89.2 (4.9) & 86.7 (1.1) &  93.2 (0)  \\
& & ASR &  \pc 100 (0) & \pc 34.7 (55.7) & \pc 33.8 (46.7) & \grc 1.2 (0.1) & \grc 1.6 (0.2) & \grc 0.8 (0.1) & \grc 0.8 (0.1) & \grc 1.5 (0.2) & \grc 0.1 (0.2) & \grc 0.5 (0.1)  & \pc 99.6 (0.6)  & \grc 2.8 (1.2) & \grc 0.6 (0)   \\
\cline{2-16}\rule{0pt}{1.0EM}

& \multirow{2}*{\shortstack{WaNet}} &  ACC & 93.0 (0.5) & 92.7 (0.2) & 92.4 (0.6) & 87.4 (0.2) & 90.9 (0.6) & 92.3 (0.1) & 92.7 (0.3) & 87.4 (0.1) & 82.6 (0.4) & 92.5 (0.6) & 88.1 (5.9) & 86.2 (1.2) & 93.0 (0.2)  \\
& & ASR & \pc 93.9 (0.8) & \pc 86.4 (2.1) & \pc 81.8 (4.5) & \grc 3.6 (0.4) & \grc 1.1 (0.2) & \pc 86.3 (2.6) & \grc 2.0 (0.3) & \grc 2.7 (0.9) & \grc 1.9 (1.6) & \pc 90.8 (5.0) & \pc 22.1 (52.6) & \grc 1.4 (0.5) & \grc 1.0 (0.1) \\
\cline{2-16}\rule{0pt}{1.0EM}

& \multirow{2}*{\shortstack{TaCT}} &  ACC & 93.6 (0) & 93.3 (0.5) &  93.0 (0.5) &  92.6 (0.2) &  91.1 (2.1) & 93.2 (0.1) &  92.7 (0.2) &  93.2 (0) &  83.6 (0.3) & 92.9 (0.2)  & 89.3 (4.2) & 84.8 (0.6) &  92.5 (0.5)  \\
& & ASR & \pc 99.6 (0.2) & \grc 1.8 (0.5) & \pc 97.3 (0.4) & \pc 98.7 (0.1) & \pc 98.4 (0.9) & \grc 0.9 (0.5) & \pc 98.0 (0.9) & \grc 1.0 (0) & \pc 96.4 (2.0) & \grc 4.0 (3.4) & \pc 99.0 (0.9) & \grc 7.5 (3.6) & \grc 1.2 (0.1)  \\
\cline{2-16}\rule{0pt}{1.0EM}

& \multirow{2}*{\shortstack{Adap-Patch}} &  ACC & 93.5 (0.2) & 92.7 (0.4) &  93.0 (0.1) & 92.5 (0) &  91.6 (1.7) & 93.5 (0.5) &  92.5 (0.1) &  92.9 (0) & 85.3 (3.1) &  93.4 (0.3)  &  86.9 (5.9) & 86.5 (1.1) &  92.4 (0.1)  \\
& & ASR & \pc 100 (0) & \grc 1.4 (1.0) & \pc 98.3 (0.1) & \pc 99.8 (0.1) & \pc 99.8 (0.1) & \pc 98.5 (2.0) & \pc 99.6 (0.1) & \grc 1.1 (0.5) & \pc 99.9 (0.1) & \pc 33.8 (56.9)  & \pc 99.9 (0.1) & \grc 11.1 (7.0) & \grc 1.4 (0.1) \\
\cline{2-16}\rule{0pt}{1.0EM}

& \multirow{2}*{\shortstack{Adap-Blend}} &  ACC & 94.0 (0) & 93.1 (0.5) &  93.2 (0.1) &  92.8 (0) &  91.8 (0.6) & 92.9 (0.3) &  93.0 (0.2) &  92.7 (0.1) &  83.9 (0.6) &  93.1 (1.2)  &  85.7 (4.6) & 86.3 (1.2) &  92.8 (0.2)  \\
& & ASR & \pc 84.8 (1.1) & \pc 77.5 (4.0) & \pc 80.3 (0.6) & \pc 41.5 (28.3) & \pc 78.7 (2.4) & \pc 20.5 (5.3) & \pc 81.6 (5.0) & \grc 3.9 (0.9) & \pc 61.9 (38.6) & \pc 82.9 (5.2) & \pc 95.1 (3.9) & \grc 7.5 (2.4) & \grc 2.2 (0.1)  \\

\bottomrule
\toprule 
\rule{0pt}{1.0EM}

\multirow{18}*{\shortstack{G\\T\\S\\R\\B}} & \multirow{2}*{\shortstack{No Poison}} &  ACC &  97.0 (0.4) & 96.5 (0.2) &  96.2 (0.1) &  94.8 (0.1) &  95.3 (0.1) & 95.7 (0.6) &  96.5 (0.2) &  95.9 (0.2) &  86.1 (1.8) &  96.2 (1.7)  &  84.0 (6.6) &  96.2 (0.2) &  96.3 (0.1)   \\
& & ASR & \gc - & \gc -  & \gc - & \gc - & \gc - & \gc - & \gc - & \gc - & \gc - & \gc  - & \gc - & \gc - & \gc -   \\
\cline{2-16}\rule{0pt}{1.0EM}

& \multirow{2}*{\shortstack{BadNet}} &  ACC &  96.7 (0.1) & 96.6 (0.5) &  96.3 (0.1) &  94.7 (0.4) &  95.7 (0.5) & 95.5 (0.4) &  96.2 (0.2) &  95.4 (0.3) &  87.2 (0.6) &  96.9 (0.5)  &  86.0 (8.5) &  96.5 (0.3) &  96.4 (0.1)   \\
& & ASR & \pc 100 (0) & \grc 0.1 (0.1) & \grc 0.2 (0) & \grc 0.2 (0) & \grc 0.2 (0) & \grc 0.2 (0.1) & \grc 0.2 (0) & \grc 0.2 (0) & \pc 37.0 (45.0) & \grc 0.9 (0.3) & \grc 12.6 (17.0) & \grc 0.8 (0.5) & \grc 0.2 (0)   \\
\cline{2-16}\rule{0pt}{1.0EM}

& \multirow{2}*{\shortstack{Blend}} &  ACC &  96.5 (0) & 96.3 (0.3) &  95.9 (0.3) &  94.2 (0.4) &  94.9 (0.5) & 95.3 (0.2) &  96.2 (0.2) & 95.3 (0.1) &  86.7 (0.7) & 96.4 (0.4)  &  90.6 (1.8) & 96.2 (0.5) &  96.2 (0)   \\
& & ASR & \pc 98.2 (0.1) & \pc 97.6 (0.3) & \pc 89.9 (2.4) & \pc 72.9 (13.1) & \pc 71.8 (3.9) & \pc 75.4 (2.7) & \pc 37.0 (18.0) & \grc 14.6 (11.7) & \pc 97.1 (0.9) & \pc 27.9 (3.8) & \grc 14.6 (3.9) & \pc 41.5 (13.1) & \grc 1.4 (0.8)   \\
\cline{2-16}\rule{0pt}{1.0EM}

& \multirow{2}*{\shortstack{Trojan}} & ACC & 97.1 (0.2) & 96.8 (0.1) & 96.6 (0.2) & 95.1 (0.1) & 95.7 (0.2) & 95.5 (0.4) & 96.7 (0.5) & 96.0 (0.3) & 87.1 (0.2) & 96.8 (0.3)  & 92.9 (0.8) & 96.2 (0.2) & 96.8 (0.1) \\
& & ASR & \pc 100 (0) & \grc 0.2 (0) & \grc 0.1 (0.1) & \grc 0.6 (0.2) & \grc 0.3 (0.2) & \grc 0.2 (0.1) & \grc 0.2 (0) & \grc 0.1 (0) & \pc 98.3 (1.1) & \grc 1.5 (0.9)  & \grc 6.5 (4.4) & \grc 0.7 (0.1) & \grc 0.2 (0.1)  \\
\cline{2-16}\rule{0pt}{1.0EM}

& \multirow{2}*{\shortstack{SIG}} &  ACC &  96.8 (0.3) & 96.5 (0.3) &  96.0 (0) &  93.2 (0.4) &  95.0 (0.2) & 95.5 (0.3) &  96.3 (0.2) &  95.2 (0) &  85.9 (1.7) & 97.2 (0.1)  &  76.2 (19.8) & 96.3 (0.3) &  96.4 (0)   \\
& & ASR & \pc 52.9 (2.1) & \pc 49.6 (4.5) & \pc 49.7 (0.1) & \pc 49.6 (4.9) & \grc 8.5 (6.8) & \pc 39.2 (1.6) & \pc 26.5 (5.9) & \pc 35.2 (12.3) & \pc 58.6 (5.9) & \pc 50.5 (4.7)  & \pc 57.5 (49.2) & \grc 12.4 (3.4) & \grc 0.4 (0.2)   \\
\cline{2-16}\rule{0pt}{1.0EM}

& \multirow{2}*{\shortstack{Dynamic}} &  ACC &  97.0 (0.2) & 96.7 (0.3) &  96.1 (0) &  95.9 (0) &  95.3 (0.1) & 95.6 (0.3) &  96.5 (0.3) &  95.9 (0.2) & 87.0 (0.2) &  96.3 (1.0)  &  87.0 (5.6) & 96.7 (0.5) &  96.4 (0.2)   \\
& & ASR & \pc 100 (0) & \pc 30.7 (47.3) & \grc 0.6 (0) & \grc 2.5 (2.0) & \grc 8.6 (1.0) & \grc 14.1 (5.7) & \grc 10.7 (3.2) & \grc 1.1 (0.2) & \pc 100 (0) & \pc 30.3 (16.9)  & \pc 68.5 (54.6) & \pc 55.0 (48.7) & \grc 0.3 (0.1)  \\
\cline{2-16}\rule{0pt}{1.0EM}

& \multirow{2}*{\shortstack{WaNet}} & ACC & 92.9 (5.7) & 95.6 (0.2) & 95.3 (0.4) & 93.5 (0.3) & 95.3 (0.1) & 94.0 (0.4) & 96.4 (0.8) & 93.4 (0.8) & 85.5 (0.8) & 96.7 (0.4) & 78.1 (7.6) & 96.9 (0.2) & 96.6 (0.1)  \\
& & ASR & \pc 70.0 (3.2) & \pc 74.0 (2.6) & \pc 70.6 (6.1) & \grc 7.1 (0.6) & \pc 76.5 (1.9) & \pc 52.0 (3.3) & \grc 0.7 (0.8) & \pc 75.9 (3.0) & \grc 6.6 (5.2) & \pc 28.8 (37.7) & \pc 89.4 (13.8) & \grc 0.3 (0.1) & \grc 0.2 (0.1) \\
\cline{2-16}\rule{0pt}{1.0EM}

& \multirow{2}*{\shortstack{TaCT}} &  ACC &  96.9 (0.1) & 96.3 (0.3) &  95.9 (0.1) &  95.7 (0.2) &  94.8 (0.2) & 95.6 (0.2) &  96.3 (0.1) &  95.8 (0.1) &  86.5 (0.8) &  96.7 (0.6)  &  92.3 (5.6) &  96.3 (0.3) &  96.6 (0.4)   \\
& & ASR & \pc 99.8 (0.1) & \pc 99.7 (0.5) & \pc 99.8 (0) & \grc 0.2 (0) & \grc 0.2 (0) & \grc 0.5 (0.3) & \grc 0.1 (0.1) & \pc 100 (0) & \pc 99.9 (0.2) & \pc 88.8 (19.4)  & \pc 68.5 (54.6) & \grc (5.0) & \grc 1.9 (0.8) \\
\cline{2-16}\rule{0pt}{1.0EM}

& \multirow{2}*{\shortstack{Adap-Patch}} &  ACC &  96.6 (0.1) & 96.6 (0.5) &  96.2 (0.2) &  95.3 (0.3) &  95.8 (0) & 95.7 (0.3) &  96.5 (0.2) &  95.8 (0.4) &  86.4 (0.4) &  96.7 (0.6)  &  67.0 (40.1) & 96.4 (0.1) &  96.3 (0.1)   \\
& & ASR & \pc 53.4 (2.3) & \grc 0.2 (0.1) & \pc 48.1 (1.4) & \grc 10.2 (5.1) & \pc 42.8 (5.9) & \grc 12.2 (2.6) & \pc 52.5 (4.3) & \pc 52.8 (2.2) & \pc 24.0 (20.0) & \pc 27.8 (8.5)  & \pc 74.7 (24.8) & \grc 2.1 (0.8) & \grc 0.2 (0)\\
\cline{2-16}\rule{0pt}{1.0EM}

& \multirow{2}*{\shortstack{Adap-Blend}} &  ACC &  96.7 (0.2) & 96.6 (0.5) &  96.5 (0.1) &  95.5 (0.1) &  94.9 (0.1) & 95.7 (0.1) &  96.3 (0.2) &  96.1 (0.4) & 86.7 (0.4) & 96.0 (0.2) & 77.5 (9.7) & 96.3 (0.3) &  96.0 (0.1)  \\
& & ASR & \pc 93.9 (0.8) & \pc 91.2 (2.9) & \pc 90.1 (0.5) & \pc 74.3 (5.0) & \pc 91.4 (0.6) & \pc 55.2 (3.6) & \pc 92.7 (2.1) & \pc 93.4 (2.4) & \pc 84.7 (3.4) & \pc 42.7 (14.0) & \pc 83.9 (17.9) & \pc 26.1 (12.2) & \grc 0.3 (0.1) \\

\bottomrule
\end{tabular}
}
\label{tab:main_defense}
\end{table*}

%% file: sections/figs/ablation_poison_rate_succint.tex
\begin{figure}
\begin{subfigure}{0.155\textwidth}
    \includegraphics[width=\textwidth]{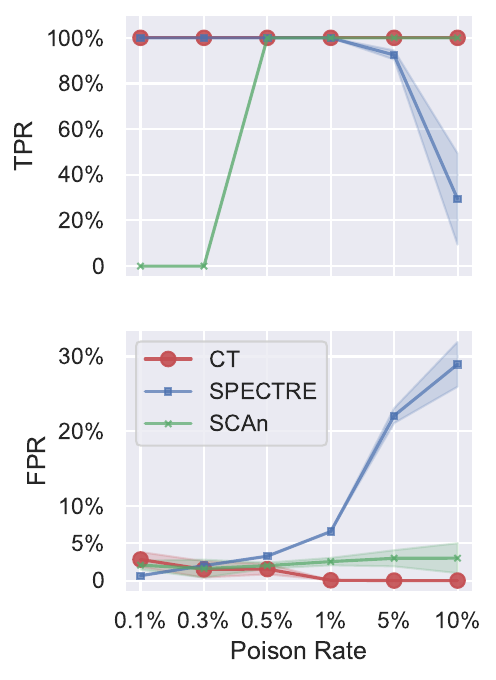}
    \caption{TaCT}
    \label{fig:ablation_poison_rate_TaCT_succint}
\end{subfigure} 
\hfill
\begin{subfigure}{0.155\textwidth}
    \includegraphics[width=\textwidth]{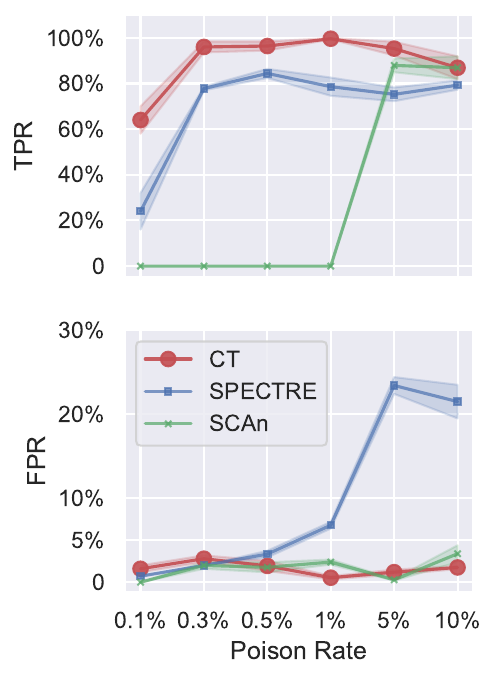}
    \caption{Adap-Patch}
    \label{fig:ablation_poison_rate_adaptive_patch_succint}
\end{subfigure} 
\hfill
\begin{subfigure}{0.155\textwidth}
    \includegraphics[width=\textwidth]{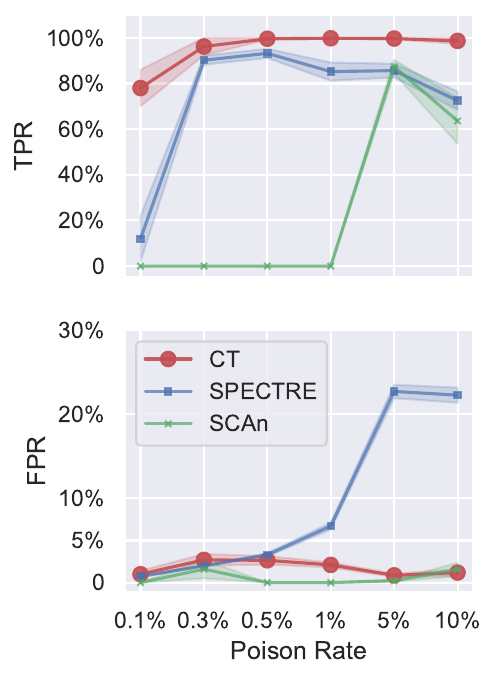}
    \caption{Adap-Blend}
    \label{fig:ablation_poison_rate_adaptive_blend_succint}
\end{subfigure}
\caption{\textbf{Ablation experiments on different poison rates.} Refer to Appendix~\ref{fig:ablation_poison_rate} for ablation results on more attacks.}
\label{fig:ablation_poison_rate_succint}
\end{figure}

%% file: sections/figs/ablation_arch.tex
\begin{figure}
\begin{center}
\includegraphics[width=0.48\textwidth]{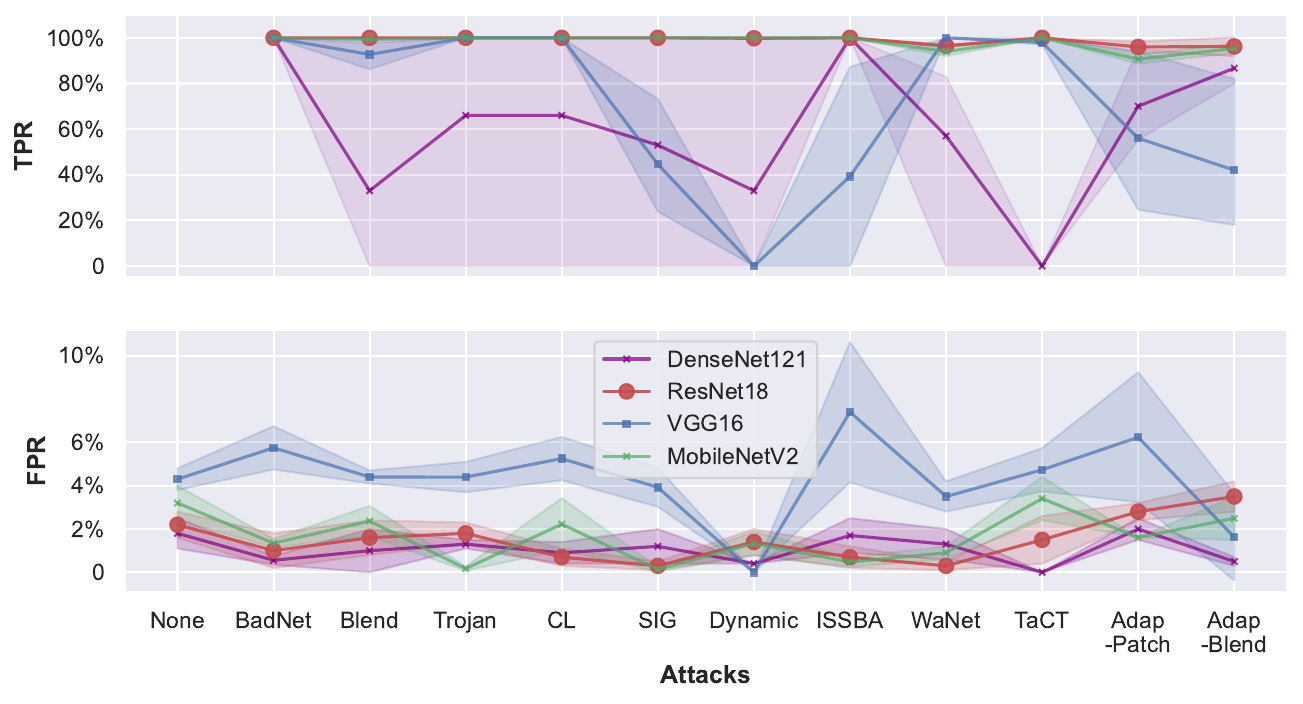}
\end{center}
\caption{
\textbf{Ablation results on different model architectures.}}
\label{fig:ablation_arch}
\end{figure}

%% file: sections/figs/ablation_clean_num.tex
\begin{figure}
\begin{center}
\includegraphics[width=0.48\textwidth]{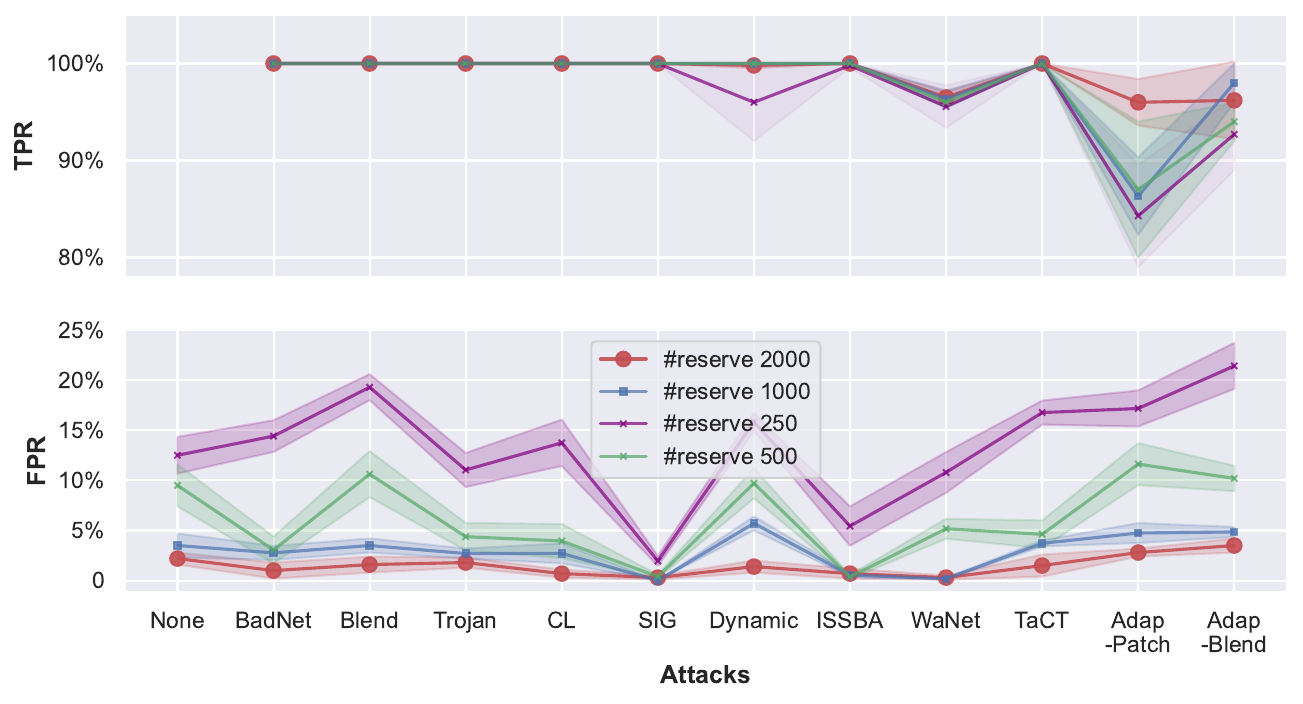}
\end{center}
\caption{
\textbf{Ablation results on size of the reserved clean set.}}
\label{fig:ablation_num_clean}
\end{figure}

%% file: sections/tables/table_few_clean.tex
\begin{table}[H]
\caption{
\textbf{\revision{CT on CIFAR10 with 250 clean samples.} }} 
\centering
\aboverulesep=0.1ex
\belowrulesep=0ex 
\resizebox{1.0\linewidth}{!}{ 
\begin{tabular}{c|cccccc}

\toprule\rule{0pt}{1.0EM}

($\%$)~~  mean~(std) & \multicolumn{2}{c}{No Defense} & \multicolumn{4}{c}{CT (ours)}\\

\cmidrule(lr){2-3} \cmidrule(lr){4-7} \rule{0pt}{1.0EM}

& ACC & ASR & TPR & FPR & ACC & ASR\\
\midrule\rule{0pt}{1.0EM}

No Poison & 94.1 (0.1) & - & - & 12.5 (1.8) & 92.1 (0.2) & - \\
BadNet & 93.8 (0.1) & 100 (0) & 100 (0) & 14.4 (1.6) & 92.1 (0.1) & 1.0 (0.2) \\
Blend & 93.6 (0.1) & 93.5 (0.2) & 100 (0) & 19.3 (1.3) & 91.9 (0.1) & 3.2 (1.3) \\
Trojan & 93.8 (0.1) & 99.9 (0.1) & 100 (0) & 11.0 (1.7) & 92.0 (0.2) & 1.8 (0.3) \\
CL & 93.6 (0) & 99.8 (0.1) & 100 (0) & 13.7 (2.3) & 92.1 (0.3) & 1.4 (0.4) \\
SIG & 93.8 (0) & 80.2 (0.6) & 100 (0) & 2.0 (1.1) & 92.6 (0.2) & 0 (0) \\
Dynamic & 93.8 (0) & 99.3 (0.4) & 96.2 (3.1) & 15.8 (1.0) & 92.0 (0.2) & 5.2 (2.1) \\
ISSBA & 93.7 (0) & 100 (0) & 99.8 (0) & 5.4 (2.0) & 92.7 (0.1) & 0.8 (0) \\
WaNet & 93.0(0.5) & 93.9 (0.8) & 95.6 (2.1) & 10.8 (2.0) & 92.5 (0.3) & 0.9 (0.2) \\
TaCT & 93.6 (0) & 99.6 (0.2) & 100 (0) & 16.8 (1.2) & 92.1 (0.1) & 1.0 (0.3) \\
Adap-Patch & 93.5 (0.2) & 100 (0) & 84.3 (5.3) & 17.2 (1.8) & 91.9 (0.1) & 0.6 (0.1) \\
Adap-Blend & 94.0 (0) & 84.8 (1.1) & 92.7 (4.7) & 21.4 (2.3) & 91.7 (0.2) & 2.1 (0.2) \\
\bottomrule
\end{tabular}
} 
\label{tab:ct-few-clean}
\end{table}

%% file: sections/tables/table_ember.tex
\begin{table}
\caption{
\textbf{Defense results of CT on Ember.} } 
\centering
\aboverulesep=0.1ex
\belowrulesep=0ex 
\resizebox{1.0\linewidth}{!}{ 
\begin{tabular}{l|cccccc}

\toprule\rule{0pt}{1.0EM}

($\%$)~~  mean~(std) & \multicolumn{2}{c}{No Defense} & \multicolumn{4}{c}{CT (ours)}\\

\cmidrule(lr){2-3} \cmidrule(lr){4-7} \rule{0pt}{1.0EM}

& ACC & ASR & TPR & FPR & ACC & ASR\\
\midrule\rule{0pt}{1.0EM}

No Poison & 99.1 (0) & - & - & 4.0~(0) & 98.7~(0.1) & - \\
Unconstrained & 99.1~(0.1) & 84.0~(3.86) & 99.9~(0.0) & 3.0~(0) & 98.8~(0) & 0~(0) \\
Constrained & 99.0~(0.1) & 62.1~(3.7) & 99.8~(0.1) & 3.0~(0) & 98.8~(0) & 0~(0) \\
\bottomrule

\end{tabular}
} 
\label{tab:ember}
\end{table}

%% file: sections/tables/table_imagenet.tex
\begin{table}
\caption{
\textbf{Defense results of CT on ImageNet-1k.} } 
\centering
\aboverulesep=0.1ex
\belowrulesep=0ex 
\resizebox{1.0\linewidth}{!}{ 
\begin{tabular}{l|ccccc}

\toprule\rule{0pt}{1.0EM}

($\%$)  & mean~(std) & No Poison & BadNet & Trojan & Blend \\
\midrule\rule{0pt}{1.0EM}
\multirow{2}*{\shortstack{No Defense}} & \multirow{1}*{\shortstack{ACC}} & 67.5~(0.1) & 67.1~(0.1) & 67.5~(0.2) & 67.6~(0.1)\\
\cline{2-6}\rule{0pt}{1.0EM}
& \multirow{1}*{\shortstack{ASR}} & - & 100~(0) & 98.9~(0.4) & 93.7~(3.8)\\
\midrule\rule{0pt}{1.0EM}
\multirow{4}*{\shortstack{CT (ours)}} & \multicolumn{1}{c}{TPR} & - & 99.5~(0.1) & 100~(0) & 100~(0)\\
& \multicolumn{1}{c}{FPR} & 0.1~(0) & 0.1~(0) & 0.1~(0) & 0.1~(0)\\
\cline{2-6}\rule{0pt}{1.0EM}
& \multirow{1}*{\shortstack{ACC}} & 67.5~(0.1) & 67.6~(0.1) & 67.4~(0.1) & 67.5~(0.1)\\
\cline{2-6}\rule{0pt}{1.0EM}
& \multirow{1}*{\shortstack{ASR}} & - & 0~(0) & 0.1~(0) & 0~(0)\\

\bottomrule

\end{tabular}
} 
\label{tab:imagenet}
\end{table}

%% file: sections/tables/retrain_ViT.tex
\begin{table}
\caption{
\revision{\textbf{Training ViT-B/16 on poisoned ImageNet-1k cleansed by CT:} after CT cleanses the dataset following the same setup in Table~\ref{tab:imagenet}, we train ViT-B/16 (as opposed to ResNet18) with the state-of-the-art GSAM algorithm~\cite{zhuang2022surrogate} on the cleansed dataset and report the new ACC and ASR. (We do one run of each experiment due to the heavy computational overhead; thus, we omit the standard deviation.)}} 
\centering
\aboverulesep=0.1ex
\belowrulesep=0ex 
\resizebox{0.8\linewidth}{!}{ 
\begin{tabular}{c|cccc}

\toprule\rule{0pt}{1.0EM}

($\%$) &  No Poison & BadNet & Trojan & Blend \\
\midrule\rule{0pt}{1.0EM}
\multirow{1}*{\shortstack{ACC}} & 80.7 & 80.8 & 80.6 & 80.7\\
\cline{1-5}\rule{0pt}{1.0EM}
\multirow{1}*{\shortstack{ASR}} & - & 0 & 0 & 0 \\

\bottomrule

\end{tabular}
} 
\label{tab:retrain-vit}
\end{table}

%% file: sections/discussions.tex
\section{Discussion}
\label{sec:discussion}

In Sec~\ref{sec:experiments}, we illustrate nice properties of confusion training~(CT) defense. In particular, we highlight that CT is built on \textbf{an arguably more fundamental premise} that the successful execution of a backdoor poisoning attack must be accompanied by the existence of backdoor correlations and corresponding backdoor poison samples that are fittable. The design of CT makes use of this fundamental premise, by simply attempting to fit those backdoor poison samples~(that we don't have knowledge of, except knowing that they should be fittable), while simultaneously preventing the fitting of samples from the clean distribution~(that we can approximate) by proactively making them less fitable. Even though, \textbf{we keep a conservative attitude so as not to oversell the security of the confusion training technique itself}. After all, it is an empirical defense without a certified guarantee, and the insights we discuss in Sec~\ref{subsec:adaptive-attacks} may also motivate stronger adaptive attacks that can further challenge CT. {Rather}, \textbf{we position CT as evidence to showcase how our proposal of the proactive mindset may inspire future advancement in \revision{backdoor poison samples detection}}. In particular, we hope our formulation and practical insights provided in Sec~\ref{subsec:proactive_mindset} can inspire the design of better proactive backdoor defenses.


%% file: sections/related.tex
\section{Related Work}
\label{sec:related_work}

\subsection{Backdoor Attacks on Neural Networks}


The focus of this study is to address the threat of \textit{backdoor poisoning attacks} on deep neural networks (DNNs), which constitute the most prevalent form of DNN backdoor attacks. These attacks~\cite{gu2017badnets,Chen2017TargetedBA,liu2017trojaning,turner2019label,barni2019new,severi2020exploring,nguyen2020input,li2021invisible,nguyen2021wanet,liu2020reflection,li2021invisible,tang2021demon,qi2023revisiting,wu2022just} involve the manipulation of a few samples in the training dataset by an attacker, resulting in the "poisoning" of the dataset. Victims will then train their own model on this poisoned dataset. Backdoor attacks also include various methods that do not fit within the paradigm of data poisoning. Examples include modifying the training process~\cite{bagdasaryan2021blind,tan2019bypassing} or using backdoored pre-trained models~\cite{yao2019latent,shen2021backdoor} for transfer learning, as well as attacks that occur at the deployment stage~\cite{liu2017fault,qi2021towards,rakin2020tbt,qi2021subnet}. These attacks involve different assumptions and threat models, which are out of the scope of this work.

\subsection{Backdoor Defenses}


\textbf{Backdoor Poison Samples Detection.} This work investigates methods for detecting poison samples in the poisoned dataset as a means of defense. In the existing literature, state-of-the-art approaches~\cite{tran2018spectral,chen2018activationclustering,tang2021demon,hayase21a} of this category consistently build their detectors upon latent separation characteristics, assuming that backdoored models will learn separate latent representations for poison and clean samples. Poison samples can then be identified via cluster analysis in the latent representation space. \citet{tran2018spectral} project the latent representations to the top PCA~\cite{pearson1901liii} direction and observe a bimodal distribution — poison and clean samples form their own modal, respectively. Later, \citet{chen2018activationclustering} independently observe similar separation and propose to use K-means~\cite{lloyd1982least} to separate the clean and poison clusters in the latent space. More recently, \citet{tang2021demon} and \citet{hayase21a} propose to use statistics of the clean distribution to further improve the cluster analysis. Researchers have also proposed using other characteristics such as the entropy of predictions under intentional perturbation~\cite{gao2019strip}, abnormal regions in the backdoored model's saliency map~\cite{chou2020sentinet}, and the speed of model fiting~\cite{li2021anti} to identify poison samples. \citet{zeng2021rethinking} propose to detect artifacts of poison samples in the frequency domain.

\textbf{Other Backdoor Defenses.} There are some other defenses, working on different stages. Neural Cleanse~\cite{wang2019neural} proposes to reverse engineer backdoor triggers. Fine-pruning~\cite{liu2018fine} suggests eliminating the model backdoor via pruning. \citet{du2019robust} and \citet{li2021anti} apply intervention during model training to suppress the influence of poison samples. \citet{xu2019Meta} directly train meta classifiers to predict whether a model is backdoored. \citet{liu2017neural} and \citet{li2021neural} attempt to eliminate the model backdoor via finetuning the model on a small clean dataset. \citet{udeshi2019model} introduce input preprocessing to suppress the effectiveness of backdoor triggers during the inference stage. There are also some certified defenses~\cite{wang2020certifying} that adapt randomized smoothing for the settings of backdoor attacks and achieve nontrivial certified accuracy against backdoor attacks with small $\ell_p$ bounded triggers.

%% file: sections/conclusion.tex
\section{Conclusion}

In this paper, we present a comprehensive investigation into methods for detecting backdoor poison samples in an effort to defend against backdoor poisoning attacks. We uncover a post-hoc workflow underlying many prior works and reveal its limitations. Accordingly, we suggest a paradigm shift by promoting a proactive mindset formulated within a unified framework. We also provide practical insights for grounded implementations. Particularly, we introduce the technique of confusion training~(CT) as a concrete instantiation of the proactive mindset. Our empirical evaluations show that CT is effective across different attack settings, outperforming prior ars. We also show that CT is scalable to large datasets and generalizable to the domain of malware detection other than computer vision. We then provide insights for adaptively attacking CT and find CT is quite robust against a number of adaptive attacks. We position CT as evidence to showcase how our proposal of the proactive mindset may inspire future advancement in \revision{backdoor poison samples detection}.


\section*{Acknowledgements}

This work was supported in part by the National Science Foundation under grants CNS-1553437 and CNS-1704105, the ARL’s Army Artificial Intelligence Innovation Institute (A2I2), the Office of Naval Research Young Investigator Award, the Army Research Office Young Investigator Prize, Schmidt DataX award, Princeton E-ffiliates Award, and Princeton's Gordon Y. S. Wu Fellowship. Any opinions, findings, and conclusions or recommendations expressed in this material are those of the author(s) and do not necessarily reflect the views of the funding agencies.

%% file: sections/appendix.tex
\appendix

\input{sections/appendix_confusion_training_protocol}

\input{sections/appendix_runtime_defenses}





\input{sections/appendix_ablation_poison_rate}

\input{sections/appendix_configurations}

\section{\revision{Theoretical Analysis of Confusion Training in A Simplified Setting}}
\label{appendix:theory}

\input{sections/appendix_theory_long.tex}

%% file: sections/appendix_confusion_training_protocol.tex
\section{Implementations of Confusion Training}
\label{appendix:ct_implementation}

\subsection{Full Algorithms and Implementations}
\label{appendix_subsec:ct_full_algorithms}

In Algorithm~\ref{alg:find_target_classes} and \ref{alg:full_version}, we present the algorithmic details. As formulated in Algorithm~\ref{alg:full_version}, our implementation of CT defense involves multiple rounds of confusion training. In our implementations for the major evaluations in Sec~\ref{sec:experiments}, there are 6 rounds in total. Specifically, for the parameters in Algorithm~\ref{alg:full_version}, we set the number of iteration $K=6$, the distillation ratios $\{r_1=\frac{1}{2}, r_2=\frac{1}{5}, r_3=\frac{1}{25}, r_4=\frac{1}{50}, r_5 = \frac{1}{100}\}$, number of confusion iters $m=6000$, confusion factor $\lambda=20$. The loss function $\loss$ is the standard cross entropy loss. By default, we use a reserved clean set $\Dreserve$ with 2000 samples. We use standard stochastic gradient descent with a weight decay of $10^{-4}$, a momentum of $0.7$, and a learning rate of $0.001$. We set the threshold $u=2$ in Algorithm~\ref{alg:find_target_classes}.


\subsection{\revision{Computational Cost}}
\label{appendix_subsec:ct_computation}

\revision{The primary computational cost of the full CT poison detection pipeline~(Algorithm~\ref{alg:full_version}) comes from the pretraining procedure in Line~\ref{line:full-pretrain}. Given that $K$ rounds of pretraining are conducted over sequentially shrinking datasets $\{D_k\}_{k=1}^K$, the computational overhead is projected to approximate $1 + \sum_{k=1}^{K-1} r_i \approx 2$ times that of a full pretraining (specific to the hyper-parameters discussed in Appendix~\ref{appendix_subsec:ct_full_algorithms}). Furthermore, the $m$ inner iterations (Line~\ref{line:confusion_iters}) resemble several epochs of fine-tuning and consume significantly fewer computational resources than pretraining. In addition, computational costs associated with subsequent steps, spanning from Line~\ref{line:get-target} to Line~\ref{line:the-end}, are relatively negligible compared to the whole procedure. In summary, the complete CT poison detection framework necessitates roughly twice the computational resources required in standard pretraining and in conventional post-hoc approaches~\cite{tran2018spectral,chen2018activationclustering,tang2021demon,hayase21a} (since the latter also primarily involves pre-training). To illustrate this, we executed our confusion training pipeline and the classical poison detection baseline Spectral Signature~\cite{tran2018spectral} on a single GPU in our workstation, which is equipped with 48 Intel Xeon Silver 4214 CPU cores, 384 GB RAM, and 8 GeForce RTX 2080 Ti GPUs. We document the execution time in Table~\ref{tab:execution-time}. Note that obtaining a ready-to-deploy model necessitates further training a downstream model on the cleansed dataset, of which the computational cost depends on the model architecture and training algorithm (as discussed in Sec~\ref{subsec:goal_and_capability_of_our_defense}).}

\begin{table}[h]
\caption{
\revision{\textbf{Execution time on a single RTX 2080 Ti GPU.}}} 
\centering
\aboverulesep=0.1ex
\belowrulesep=0ex 
\resizebox{0.6\linewidth}{!}{ 
\begin{tabular}{c|cc}

\toprule\rule{0pt}{1.0EM}

(minutes) &  Spectral Signature~\cite{tran2018spectral} & CT (ours) \\
\midrule\rule{0pt}{1.0EM}
\multirow{1}*{\shortstack{CIFAR10}} & 77 & 150 \\
\cline{1-3}\rule{0pt}{1.0EM}
\multirow{1}*{\shortstack{GTSRB}} & 41 & 93 \\
\bottomrule
\end{tabular}
}
\label{tab:execution-time}
\end{table}

\input{sections/assets/confusion_training_full.tex}

%% file: sections/assets/confusion_training_full.tex
\begin{algorithm}[h]
\caption{\mbox{Identify Potential Target Classes}}

\textbf{Input:} Dataset $\Dpoiall = \{(\Tilde{x}_i,\Tilde{y}_i)\}_{i=1}^n$, Inference Model $\thetact$, The Number of Classses $C$, Gaussian Density Function $N(\cdot;\mu,\sigma)$, a threshold $u$\\
\textbf{Output:} $T\subseteq \{1,\dots,C\}$, indicating potential target classes
\begin{algorithmic}[1]
\State $T \gets \{\}$
\For{$c=1,...,C$} 
\State $H \gets \{h(\Tilde{x};\thetact) | (\Tilde{x},\Tilde{y}) \in \Dpoiall \land \Tilde{y} = c \}$ \Comment{Latent Vectors}
\State $H_0 \gets \{h(\Tilde{x};\thetact) | (\Tilde{x},\Tilde{y}) \in \Dpoiall \land \Tilde{y} = c \land F(\Tilde{x};\thetact) \ne c\}$
\State $H_1 \gets H \setminus H_0$
\State $U,\Lambda, V \gets SVD_2(H)$ \Comment{Dimension Reduction}
\State $S \gets \{U^T h \ | h \in H\}$ \Comment{Top-2 PCA Directions}
\State $S_0, S_1 \gets \{U^T h \ | h \in H_0\}, \{U^T h \ | h \in H_1\}$ 
\State $\widehat{\mu}, \widehat{\sigma} \gets$ empirical mean and covariance of $S$
\State $\widehat{\mu}_0, \widehat{\sigma}_0 \gets$ empirical mean and covariance of $S_0$ 
\State $\widehat{\mu}_1, \widehat{\sigma}_1 \gets$ empirical mean and covariance of $S_1$
\State $L \gets \sqrt[|S|]{\frac{\prod_{s_0 \in S_0}  N(s_0;\widehat{\mu}_0, \widehat{\sigma}_0)\cdot \prod_{s_1 \in S_1}  N(s_1;\widehat{\mu}_1, \widehat{\sigma}_1)}{\prod_{s \in S}  N(s;\widehat{\mu}, \widehat{\sigma})}}$
\If{$L > u$}
\State $T \gets T \cup \{c\}$
\EndIf
\EndFor
\State \Return $T$
\end{algorithmic}
\label{alg:find_target_classes}
\end{algorithm}

\begin{algorithm}[h]
\caption{\mbox{Full Algorithm of CT Poison Detection}}

\textbf{Input:} Dataset $\Dpoiall = \{(\Tilde{x}_i,\Tilde{y}_i)\}_{i=1}^n$ to be cleansed, reserved clean set $\Dreserve$, loss function $\loss$, confusion factor $\lambda$, number of confusion iterations $m$, number of confusion training rounds $K$, distillation ratios $\{r_i\}_{i=1}^{K-1}$\\
\textbf{Output:} The cleansed dataset $D^*$
\begin{algorithmic}[1]

\State $D_1 \gets \Dpoiall$

\For{$k=1,\dots,K$}

  \State $\thetact \gets \erm(D_k)$
  \Comment{Pretrain}\label{line:full-pretrain}
\For{$i=1,...,m$}\Comment{Confusion Training}\label{line:confusion_iters}
      \State $(\mathbf{\widetilde{X}}_i, \mathbf{\widetilde{Y}}_i) \gets$ a random batch from $D_k$                  
      \State ${\mathbf X}'_i \gets$ a random batch from the unlabeled $\Dreserve$ 
      \State ${\mathbf Y}'_i \gets F({\mathbf X}'_i;\bdmodel)$ \Comment{Pseudo Labels}
      \State ${\mathbf Y}^* \gets$ random mislabels other than ${\mathbf Y}'_i$
      \State $\ell_{ct} \gets  \frac{\loss(f(\mathbf{\widetilde{X}}_i;\theta_{ct}), \mathbf{\widetilde{Y}}_i) + (\lambda-1) \cdot  \loss(f({\mathbf X}'_i;\theta_{ct}), \mathbf{Y}^*)}{\lambda}$ 
      \State $\thetact \gets$ one step gradient descent on $\ell_{\text{ct}}$
\EndFor
\If{$k \textless K$}
\State $D_{k+1} \gets $ sort samples in $\Dpoiall$ according to their loss values $\loss(f(\cdot;\thetact),y)$ and select the first $\lfloor r_k \cdot n \rfloor$ samples with least loss values \Comment{Iterative Poison Distillation}
\EndIf
\EndFor

\State $D^* \gets D$
\State $T \gets$ Apply Algorithm~\ref{alg:find_target_classes} with $\thetact$ \label{line:get-target}

\For{$i=1,...,n$} 
\Comment{Poison Detection}
\If{$\Tilde{y}_i \in T \land \Tilde{y}_i = F(\Tilde{x}_i;\theta_{ct})$} 
\State $D^* \gets$ remove $(\Tilde{x}_i, \Tilde{y}_i)$ from $D^*$
\EndIf
\EndFor \label{line:the-end}

\State \Return $D^*$
\end{algorithmic}
\label{alg:full_version}
\end{algorithm}

%% file: sections/appendix_runtime_defenses.tex
\vspace{-3mm}
\section{\revision{Comparison with Additional Defenses}}
\label{appendix:runtime-defenses}

\revision{A line of work aims to design robust training algorithms that intend to prevent models from learning any backdoor during training~\cite{li2021anti,huang2022backdoor,tao2022model,wang2022training}. 
The conceptual underpinning that propels these methodologies bears considerable relevance to our study, given that they also strive to proactively devise training procedures that are robust enough to counter backdoor attacks. In Table~\ref{tab:main_defense}, we compared CT with ABL~\cite{li2021anti} from this category of defenses. For a more comprehensive comparison with these defenses, we evaluate three additional defenses~(DBD~\cite{huang2022backdoor}, MOTH~\cite{tao2022model} and NONE~\cite{wang2022training}) from this category and evaluate them on CIFAR10 
in Table~\ref{tab:runtime}.  
Our implementations strictly follow the original configurations of their open-source repositories, and our evaluation follows the same setup used for Table~\ref{tab:main_defense}. Notably, CT continues exhibiting stronger resilience than the three additional baselines.}

\input{sections/tables/table_runtime}

%% file: sections/tables/table_runtime.tex
\begin{table}[h]
\caption{
\textbf{\revision{Comparing CT with 3 additional backdoor defenses on CIFAR10.}}} 
\aboverulesep=0.1ex
\belowrulesep=0ex 

\resizebox{\linewidth}{!}{
\begin{tabular}{l|cccccccc}
\toprule\rule{0pt}{1.0EM}
  ($\%$)  &  mean~(std) &
\multicolumn{1}{c}{} &
\multicolumn{1}{c}{No Defense} &
\multicolumn{1}{c}{DBD} & 
\multicolumn{1}{c}{MOTH}  &
\multicolumn{1}{c}{NONE}  &
\multicolumn{1}{c}{CT~(ours)}  \\ 
\midrule\rule{0pt}{1.0EM}

\multirow{22}*{\shortstack{C\\I\\F\\A\\R\\10}} 
& \multirow{2}*{\shortstack{No Poison}} &  ACC &  94.1 (0.1) & 92.2 (0.5)  &  91.8 (0.3) &  93.3 (0.1) &  92.2 (0.3)   \\
& &  ASR  & \gc -  & \gc -  & \gc - & \gc - & \gc -   \\
\cline{2-8}\rule{0pt}{1.0EM}

& \multirow{2}*{\shortstack{BadNet}} &  ACC & 93.8 (0.1)  & 92.2 (0.4)   &  91.7 (0.5) &  93.3 (0.2) &  93.2 (0.1)   \\
& &  ASR  & \pc 100 (0) & \grc 1.8 (0.1)  & \grc 1.2 (0.3) & \grc 1.0 (0) & \grc 0.8 (0.2)   \\
\cline{2-8}\rule{0pt}{1.0EM}

& \multirow{2}*{\shortstack{Blend}} &  ACC & 93.6 (0.1) &  92.0 (0.4)  &  91.2 (0.2) &  93.2 (0.2) &  93.1 (0.2)   \\
& &  ASR  & \pc 93.5 (0.2) & \grc 4.4 (0.3)  & \pc 89.9 (1.5) & \pc 89.0 (0.8) & \grc 1.6 (0.3)   \\
\cline{2-8}\rule{0pt}{1.0EM}

& \multirow{2}*{\shortstack{Trojan}} &  ACC & 93.8 (0.1) &  92.3 (0.3)  &  90.9 (0.2) &  93.1 (0.1) &  92.9 (0.1)   \\
& &  ASR  & \pc 99.9 (0.1) & \grc  2.7 (0.2)  & \grc 7.6 (3.3) & \grc 2.4 (0.2) & \grc 1.7 (0.4)   \\
\cline{2-8}\rule{0pt}{1.0EM}

& \multirow{2}*{\shortstack{CL}} &  ACC & 93.6 (0) &   93.1 (0.3)  &  91.3 (0.3) &  93.3 (0.1) &  93.5 (0.2)  \\
& &  ASR  & \pc 99.8 (0.1) & \pc 89.8 (0.1)  & \grc 1.3 (0.4) & \grc 1.2 (0.1) & \grc 0.9 (0)   \\
\cline{2-8}\rule{0pt}{1.0EM}

& \multirow{2}*{\shortstack{SIG}} &  ACC & 93.8 (0) &  91.5 (0.4)  &  90.9 (0.4) &  93.2~(0.1) &  93.2 (0)   \\
& &  ASR  & \pc 80.2 (0.6) & \pc 49.8 (1.1)  & \pc 98.2 (0.5) & \pc 86.4 (1.0) & \grc 0.1 (0)   \\
\cline{2-8}\rule{0pt}{1.0EM}

& \multirow{2}*{\shortstack{Dynamic}} &  ACC & 93.8 (0) &  92.0 (0.3)  &  91.5 (0.5) &  93.5 (0.2) &  93.2 (0)   \\
& &  ASR  & \pc 99.3 (0.4) & \pc 89.3 (0.1)  & \pc 84.3 (7.8) & \grc 4.4 (0.4) & \grc 3.7 (1.6)   \\
\cline{2-8}\rule{0pt}{1.0EM}

& \multirow{2}*{\shortstack{ISSBA}} &  ACC & 93.7 (0) &  90.6 (0.8)  &  91.0 (0.2) &  93.4 (0.2) &  93.2 (0)   \\
& &  ASR  & \pc 100 (0) & \grc 2.4 (0.1)  & \pc 42.7 (27.5) & \pc 62.2 (24.7) & \grc 0.6 (0)   \\
\cline{2-8}\rule{0pt}{1.0EM}

& \multirow{2}*{\shortstack{WaNet}} &  ACC & 93.0 (0.5) &  90.9 (0.3)  &  90.7 (0.2) &  92.8 (0.1), &  93.0 (0.2)   \\
& &  ASR  & \pc 93.9 (0.8) & \grc 2.1 (0.2)  & \grc 5.4 (3.2) & \pc 91.5 (0.6) & \grc 1.0 (0.1)   \\
\cline{2-8}\rule{0pt}{1.0EM}

& \multirow{2}*{\shortstack{TaCT}} &  ACC & 93.6 (0) &  91.9 (0.4)  &  91.3 (0.1) &  93.3 (0.1) &  92.5 (0.5)   \\
& &  ASR  & \pc 99.6 (0.2) & \grc 2.0 (0.2)
& \pc 35.6 (23.8) & \grc 5.2 (2.7) & \grc 1.2 (0.1)   \\
\cline{2-8}\rule{0pt}{1.0EM}

& \multirow{2}*{\shortstack{Adap-Patch}} &  ACC  & 93.5 (0.2) &  91.8 (0.3)  &  91.4 (0.2) &  93.0 (0.3) &  92.4 (0.1)   \\
& &  ASR  & \pc 100 (0) & \grc 2.6 (0.3)  & \pc 29.9 (15.4) & \grc 6.2 (0.7)  & \grc 1.4 (0.1)   \\
\cline{2-8}\rule{0pt}{1.0EM}

& \multirow{2}*{\shortstack{Adap-Blend}} &  ACC & 94.0 (0) &  92.7 (0.2)  &  91.5 (0.2) &  93.1 (0.2) &  92.8 (0.2)   \\
& &  ASR  & \pc 84.8 (1.1) & \grc 2.4 (0.1)  & \grc 9.9 (7.1) & \pc 47.6 (0.7) & \grc 2.2 (0.1)   \\
\bottomrule
\end{tabular}
}
\label{tab:runtime}
\end{table}

%% file: sections/appendix_ablation_poison_rate.tex
\section{Full Ablation Results on Poison Rates}
\label{appendix:ablation_poison_rate}

\input{sections/figs/ablation_poison_rate}

We hereby present our full ablation experimental configuration and results.

For most attacks, we consider the set of poison rates $\{0.1\%, 0.3\%, 0.5\%, 1\%, 5\%, 10\%\}$ which covers both the high and low poison rate regions. For clean label attacks~(CL and SIG), since only target class samples are poisoned, the maximum poison rate is set to $5\%$ to avoid the trivial case that the entire target class is poisoned. For ISSBA and WaNet, the minimum poison rates are set to $1\%$ and $2\%$ respectively, as these attacks with lower poison rates can no longer be effective. For each setting, we also compare CT with the two strongest baseline detectors SPECTRE and SCAn. 

As shown in Fig~\ref{fig:ablation_poison_rate}, \textit{CT is consistently effective (TPR$\approx$100\%) and outperforms the baselines in both low and high poison rates regions against most attacks}. Though the TPR of CT slightly drops in the ultra-low poison rate of $0.1\%$ against SIG, Dynamic, Adap-Patch, and Adap-Blend, it is still better than or at least comparable with the baselines. CT also suffers some drops of TPR in high poison rates against Adap-Patch, but still outperforms the baselines in these cases.

%% file: sections/figs/ablation_poison_rate.tex
\begin{figure*}[h]
\begin{subfigure}{0.16\textwidth}
\includegraphics[width=\textwidth]{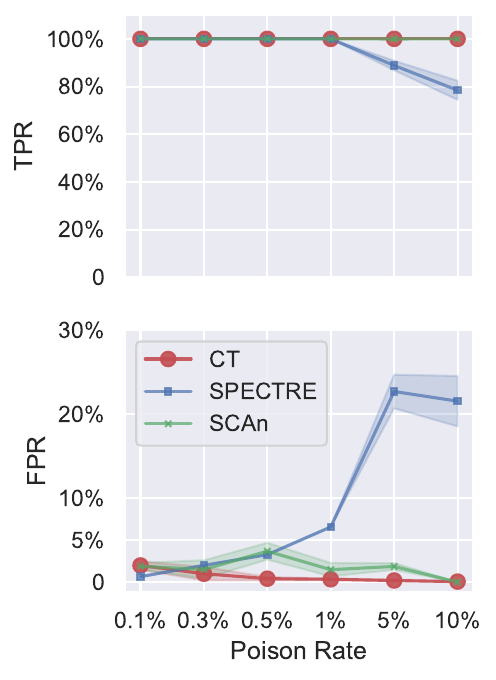}
    \caption{Badnet}
    \label{fig:ablation_poison_rate_badnet}
\end{subfigure}
\hfill
\begin{subfigure}{0.16\textwidth}
\includegraphics[width=\textwidth]{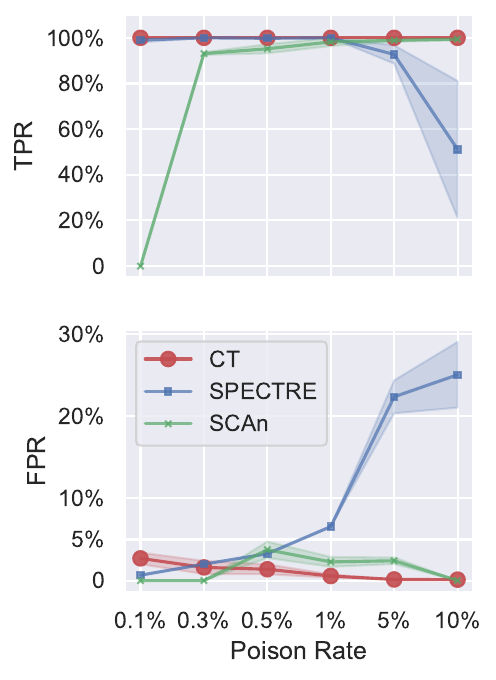}
    \caption{Blend}
    \label{fig:ablation_poison_rate_blend}
\end{subfigure}
\hfill
\begin{subfigure}{0.16\textwidth}
\includegraphics[width=\textwidth]{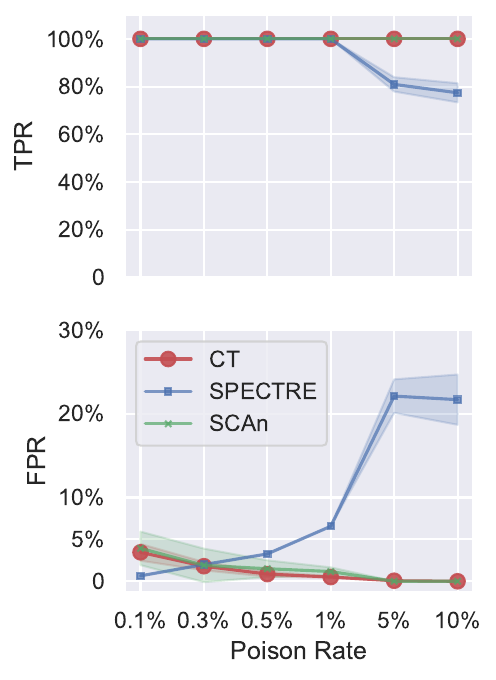}
    \caption{Trojan}
    \label{fig:ablation_poison_rate_trojan}
\end{subfigure}
\hfill
\begin{subfigure}{0.16\textwidth}
    \includegraphics[width=\textwidth]{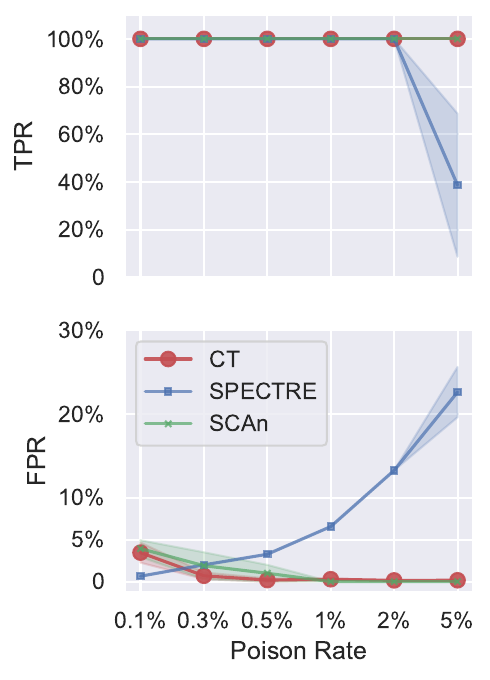}
    \caption{CL}
    \label{fig:ablation_poison_rate_clean_label}
\end{subfigure}   
\hfill
\begin{subfigure}{0.16\textwidth}
    \includegraphics[width=\textwidth]{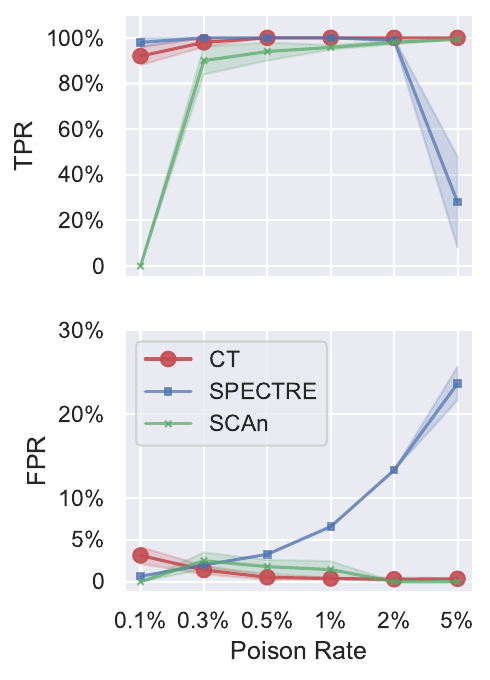}
    \caption{SIG}
    \label{fig:ablation_poison_rate_SIG}
\end{subfigure}   
\hfill
\begin{subfigure}{0.16\textwidth}
    \includegraphics[width=\textwidth]{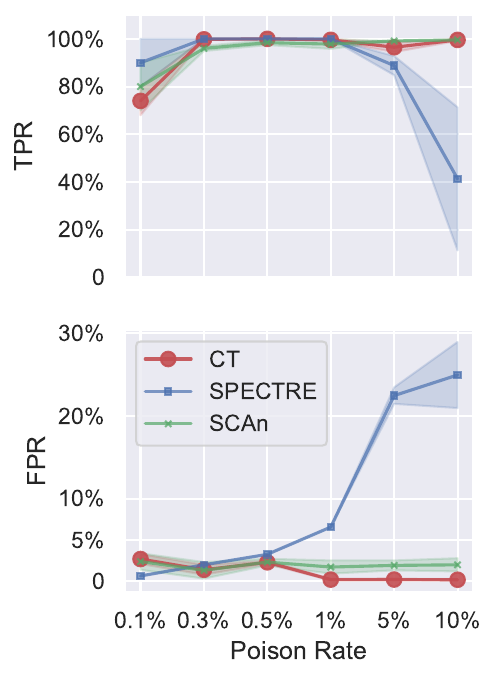}
    \caption{Dynamic}
    \label{fig:ablation_poison_rate_dynamic}
\end{subfigure} 
\hfill
\begin{subfigure}{0.16\textwidth}
    \includegraphics[width=\textwidth]{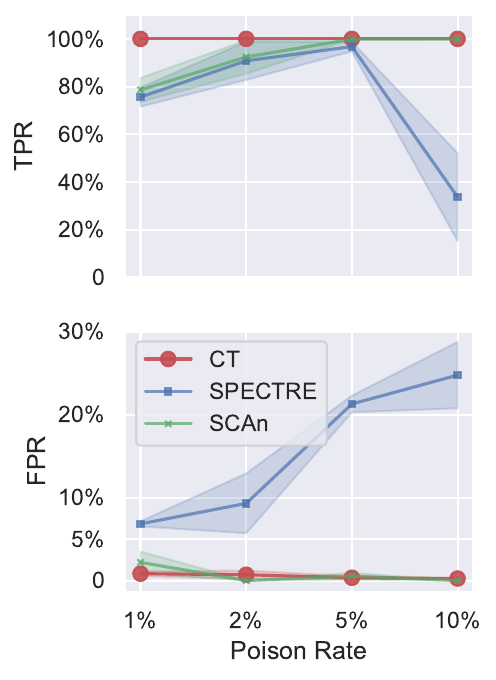}
    \caption{ISSBA}
    \label{fig:ablation_poison_rate_ISSBA}
\end{subfigure} 
\hfill
\begin{subfigure}{0.16\textwidth}
    \includegraphics[width=\textwidth]{sections/imgs/ablation/poison_rate/ablation_p_TaCT.pdf}
    \caption{TaCT}
    \label{fig:ablation_poison_rate_TaCT}
\end{subfigure} 
\hfill
\begin{subfigure}{0.16\textwidth}
    \includegraphics[width=\textwidth]{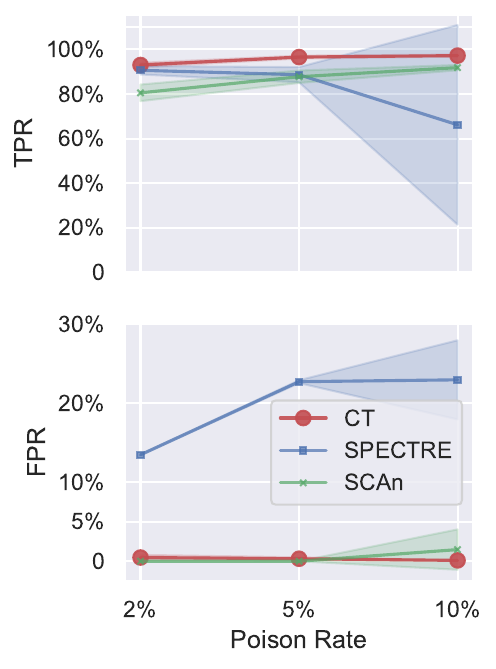}
    \caption{WaNet}
    \label{fig:ablation_poison_rate_WaNet}
\end{subfigure} 
\hfill
\begin{subfigure}{0.16\textwidth}
    \includegraphics[width=\textwidth]{sections/imgs/ablation/poison_rate/ablation_p_adaptive_patch.pdf}
    \caption{Adaptive Patch}
    \label{fig:ablation_poison_rate_adaptive_patch}
\end{subfigure} 
\hfill
\begin{subfigure}{0.16\textwidth}
    \includegraphics[width=\textwidth]{sections/imgs/ablation/poison_rate/ablation_p_adaptive_blend.pdf}
    \caption{Adaptive Blend}
    \label{fig:ablation_poison_rate_adaptive_blend}
\end{subfigure} 
\caption{\textbf{Ablation experiments on different poison rates.} For most attacks, the poison rates are from 0.1\% to 10\%. For CL and SIG, since only target class samples are poisoned, their maximum poison rates are 5\% instead of 10\%, to avoid the entire target class being poisoned which may lead to severe clean performance loss. For ISSBA and WaNet, the minimum poison rates are $1\%$ and $2\%$ respectively, in order to guarantee the backdoor being successfully injected.}
\label{fig:ablation_poison_rate}
\end{figure*}

%% file: sections/appendix_configurations.tex
\section{Configurations of Experiments}
\label{appendix:detailed_configurations_of_baselines}





\subsection{Datasets}

We use four public datasets in our evaluations, including image classification (CIFAR10~\cite{krizhevsky2009learning}, GTSRB~\cite{stallkamp2012man}, and ImageNet~\cite{deng2009imagenet}) and malware classification (EMBER~\cite{anderson2018ember}). 

\textit{CIFAR10.} CIFAR10~\cite{krizhevsky2009learning} is a benchmark dataset for image classification, encompassing common objects such as dogs, cats, and airplanes among its 10 classes. It comprises 50,000 training samples and 10,000 test samples, each with a resolution of 32$\times$32 pixels.

\textit{GTSRB.} The German Traffic Sign Recognition Benchmark (GTSRB)~\cite{stallkamp2012man} is a widely used image classification dataset for traffic sign recognition, aiming at facilitating autonomous driving. The dataset consists of 43 different types of traffic signs, with 39,211 samples in the training set and 12,630 samples in the test set. Prior to conducting experiments, the images were resized to a resolution of 32$\times$32 pixels.

\textit{ImageNet.} The ImageNet dataset~\cite{deng2009imagenet} is a common benchmark for image classification, consisting of 1.3 million training images and 50,000 validation images across 1000 classes. The images in ImageNet are resized and cropped to a resolution of 224$\times$224 pixels before being utilized as inputs for our models.

\textit{Ember.} Ember~\cite{anderson2018ember} is a well-known public resource for malware classification, which includes both malware and goodware samples. The dataset
includes 2,351-dimension features extracted from 1.1 million Windows portable executable files. In our experiment, we use labeled training and test sets that comprise 600,000 and 200,000 samples, respectively, with an equal distribution of benign and malicious samples.

For each dataset, we first follow the default training/test set split by Torchvision~\cite{marcel2010torchvision}. Recall that our defender is assumed to have a small reserved clean set at hand~(as defined in Sec~\ref{subsec:goal_and_capability_of_our_defense}). To implement this setting, for CIFAR10 and GTSRB, we further randomly pick {2000} samples from the test split to simulate the reserved clean set, and leave the rest part of the test split for evaluation. For, ImageNet and Ember, we randomly pick {5000} samples from the test split to simulate the reserved clean set.

\subsection{Training Backdoored Models}
\label{appendix_subsec:details_training_backdoored_models}

For all models on image datasets (CIFAR10, GTSRB and ImageNet), we use ResNet18~\cite{he2016deep} as the architecture; for the Ember dataset, we use the default EmberNN~\cite{severi2021explanation} as the malware detection architecture. SGD with a momentum of 0.9, a weight decay of $10^{-4}$ ($10^{-6}$ for Ember), and a batch size of 128 (256 for ImageNet and 512 for Ember), is used for optimization. Initially, we set the learning rate to $0.1$. On CIFAR10, we follow the standard 200 epochs stochastic gradient descent procedure, with an initial learning rate $0.1$, which is then multiplied by a factor of $0.1$ at the epochs of $100$ and $150$. On GTSRB we use 100 epochs of training, with an initial learning rate $0.01$, which is then multiplied by $0.1$ at the epochs of $30$ and $60$. On ImageNet, we train the model for 90 epochs, with an initial learning rate $0.1$, which is then multiplied by $0.1$ at the epochs of $30$ and $60$. On Ember, we train EmberNN for 10 epochs with learning rate $0.1$.

\subsection{Configurations of Baseline Attacks}
\label{appendix_subsec:details_baseline_attacks}


During implementing these attacks, we follow the protocols and suggested default configurations of their original papers and open-source implementations. In Table~\ref{tab:attack_config}, we summarize the hyperparameters that are used for each poison strategy. Specifically, \textit{Target Class} is the class that the backdoor trigger is correlated to, \textit{Poison Rate} denotes the portion of training samples that are stamped with the trigger and labeled to the target class. In addition, TaCT~\cite{tang2021demon} only chooses samples from a \textit{Source Class} to construct poison samples, and they will also randomly pick a portion~(\textit{Cover Rate}) of samples from some \textit{Cover Classes} and also plant triggers to them while keeping them still correctly labeled as their semantic labels. WaNet~\cite{nguyen2021wanet}, Adap-Blend and Adap-Patch~\cite{qi2023revisiting} also keep a portion~(\textit{Cover Rate}) of samples planted with triggers but still correctly labeled. Besides, for all these attacks we implement, we use the same trigger patterns to those suggested by their original papers.

For constrained and unconstrained attacks on Ember, we follow the strategies proposed in the original paper~\cite{severi2021explanation}. The constrained attack confines backdoor triggers to those features that are really editable
in PE files, while the unconstrained attack allows modifying all features during poisoning. The poison rate is 1\% for both attacks. For the constrainted attack, the trigger watermark size is 17, with attack strategy "LargeAbsSHAP x MinPopulation"; for the unconstrained attack, the trigger watermark size is 32, with attack strategy "Combined Feature Value Selector".

\input{sections/tables/attack_config}

\subsection{Configurations of Baseline Defenses}
\label{appendix_subsec:details_baseline_defenses}


As mentioned in Sec~\ref{subsubsec:exp_setup}, we compare confusion training with 14 backdoor defenses, including 7 backdoor poison samples detectors~\cite{tran2018spectral,chen2018activationclustering,gao2019strip,tang2021demon,hayase21a,chou2020sentinet,zeng2021rethinking}), and 7 defenses of other types~(\cite{liu2018fine,wang2019neural,li2021anti,li2021neural,huang2022backdoor,tao2022model,wang2022training}.

Some important configurations of these defenses are specified as follows. Refer to our code for more implementation details.
\begin{itemize}
    \item SentiNet~\cite{chou2020sentinet} is adapted to a poison detector for training sets, set with a 5\% FPR. For inputs with patch triggers, we provide SentiNet with the oracle knowledge of patch locations; for other inputs, SentiNet locates the top 15\% pixels with the highest GradCAM~\cite{selvaraju2017grad} activation scores.
    \item STRIP~\cite{gao2019strip} is also adapted to a poison detector for training set, set with a 10\% FPR.
    \item SS~\cite{tran2018spectral} removes $\min(1.5 * |\Dpoi/\Dpoiall|, 0.5 * \text{class size}$ suspected samples from every class.
    \item AC~\cite{chen2018activationclustering} cleanses any clusters with size $<$35\% of the class size.
    \item Frequency~\cite{zeng2021rethinking} predicts samples to be poison or clean with a binary classifier. We directly use their provided pretrained model for that purpose.
    \item SCAn~\cite{tang2021demon} cleanses classes with scores larger than $e$.
    \item SPECTRE~\cite{hayase21a} removes $\min(1.5 * |\Dpoi/\Dpoiall|, 0.5 * \text{class size})$ suspected samples only from the top $C/2$ classes with the highest QUE score.
    \item FP~\cite{liu2018fine} prunes the neurons of the last layer before the classifier layers of the model. The maximum prune ratio is set to 99\% for CIFAR10 and 75\% for GTSRB, both with 100 finetuning epochs and maximum allowed accuracy drop threshold of 10\%.
    \item NC~\cite{wang2019neural} reverse-engineers the potential trigger for each class. The class where its trigger norm has the maximum anomaly indice >2 (if exists) is the suspicious class, where the corresponding reversed trigger is used to unlearn the backdoor.
    \item ABL~\cite{li2021anti} isolates suspicious training samples and use them to unlearn the backdoor of the model. Since our poison rates are significantly smaller than those the original paper has estimated, we isolate 0.1\% samples (0.5\% for GTSRB) by Flooding method (\texttt{flooding}=0.3) in 15 epochs (5 epochs for GTSRB).
    \item NAD~\cite{li2021neural} first trains a teacher model based on the poison model for 10 epochs, then distills the teacher model to obtain a student model for 20 epochs.
    \item  {DBD~\cite{huang2022backdoor} first runs 1000 epochs of contrastive learning to learn a purified feature extractor and then 200 epochs of semi-supervised fine-tuning to get the final model.}
    \item {MOTH~\cite{tao2022model} takes 0.1 of batch samples to samp trigger during orthogonalization, a warm ratio of 0.5 during warmup. For consistency, 2000 clean samples are used for hardening.} 
    \item {NONE~\cite{wang2022training} takes 200 epochs of pertaining and 20 epochs of defense loops. The maximal reset fraction is set to 0.03.}
\end{itemize}

%% file: sections/tables/attack_config.tex
\begin{table}
\caption{
\textbf{Hyperparameters of Baseline Attacks} 
} 
\centering
\resizebox{0.99\linewidth}{!}{ 
\begin{tabular}{cccc}
\toprule
 & 
\multicolumn{1}{c}{CIFAR10~\cite{krizhevsky2009learning}} & \multicolumn{1}{c}{GTSRB~\cite{stallkamp2012man}} &
\multicolumn{1}{c}{ImageNet~\cite{deng2009imagenet}} \cr
\midrule

\multirow{1}*{BadNet~\cite{gu2017badnets}} &
\multirow{1}*{\shortstack{Poison Rate = $0.3\%$}} & 
\multirow{1}*{\shortstack{Poison Rate = $1.0\%$}} &  
\multirow{1}*{\shortstack{Poison Rate = $1.0\%$}}\cr
\midrule
\multirow{1}*{Blend~\cite{Chen2017TargetedBA}} &
\multirow{1}*{\shortstack{Poison Rate = $0.3\%$}} & 
\multirow{1}*{\shortstack{Poison Rate = $1.0\%$}} &  
\multirow{1}*{\shortstack{Poison Rate = $1.0\%$}}\cr
\midrule
\multirow{1}*{Trojan~\cite{liu2017trojaning}} &
\multirow{1}*{\shortstack{Poison Rate = $0.3\%$}} & 
\multirow{1}*{\shortstack{Poison Rate = $1.0\%$}} &  
\multirow{1}*{\shortstack{Poison Rate = $1.0\%$}}\cr
\midrule
\multirow{1}*{CL~\cite{turner2019label}} &
\multirow{1}*{\shortstack{Poison Rate = $0.3\%$}} & 
\multirow{1}*{/} &  
\multirow{1}*{/}\cr
\midrule
\multirow{1}*{SIG~\cite{barni2019new}} & 
\multirow{1}*{\shortstack{Poison Rate = $2.0\%$}} & 
\multirow{1}*{\shortstack{Poison Rate = $2.0\%$}} &  
\multirow{1}*{/}\cr
\midrule
\multirow{1}*{Dynamic~\cite{nguyen2020input}} & 
\multirow{1}*{\shortstack{Poison Rate = $0.3\%$}} & 
\multirow{1}*{\shortstack{Poison Rate = $0.3\%$}} &  
\multirow{1}*{/}\cr
\midrule
\multirow{1}*{ISSBA~\cite{li2021invisible}} &
\multirow{1}*{\shortstack{Poison Rate = $2.0\%$}} & 
\multirow{1}*{/} &  
\multirow{1}*{/}\cr
\midrule
\multirow{2}*{WaNet~\cite{nguyen2021wanet}} & 
\multirow{2}*{\shortstack{Poison Rate = $5.0\%$ \\ Cover Rate = $10.0\%$}} &  
\multirow{2}*{\shortstack{Poison Rate = $5.0\%$ \\ Cover Rate = $10.0\%$}} & 
\multirow{2}*{/}\cr\cr
\midrule
\multirow{4}*{TaCT~\cite{tang2021demon}} &
\multirow{4}*{\shortstack{Poison Rate = $0.3\%$\\ Source Class = 1 \\ Cover Classes = 5,7 \\ Cover Rate = $0.3\%$}} &  
\multirow{4}*{\shortstack{Poison Rate = $0.5\%$\\ Source Class = 1 \\ Cover Classes = 5,7 \\ Cover Rate = $0.5\%$}} &  
\multirow{4}*{/}\cr\cr\cr\cr
\midrule
\multirow{2}*{Adap-Patch~\cite{qi2023revisiting}} &
\multirow{2}*{\shortstack{Poison Rate = $0.3\%$\\ Cover Rate = $0.3\%$}} &  
\multirow{2}*{\shortstack{Poison Rate = $0.3\%$\\ Cover Rate = $0.3\%$}} &  
\multirow{2}*{/}\cr\cr
\midrule
\multirow{2}*{Adap-Blend~\cite{qi2023revisiting}} &
\multirow{2}*{\shortstack{Poison Rate = $0.3\%$\\ Cover Rate = $0.6\%$}} &  
\multirow{2}*{\shortstack{Poison Rate = $0.5\%$\\ Cover Rate = $1.0\%$}} &  
\multirow{2}*{/}\cr\cr
\bottomrule
\end{tabular}
}
\label{tab:attack_config}
\end{table}

%% file: sections/appendix_theory_long.tex
\subsection{\revision{Overview of The Analysis}}
\label{appendix:theory-overview}

To formally justify the effectiveness of confusion training from a theoretical perspective, we  study a simplified setting where we train an overparameterized linear model for binary classification tasks. Specifically, we study the classic Gaussian class-conditional model \citep{chatterji2021finite} as the training data distribution.  \revision{We use a simplified setting as a full theoretical treatment of the training dynamics of neural networks is widely recognized as a complex endeavor. Furthermore, our simplifying assumptions have been frequently employed in the corpus of deep learning theory literature, such as overparameterized linear models and binary data ~\cite{li2019stochastic,zancato2020predicting,mori2022power}. The primary intention of our analysis is to offer an understanding and insight into the operational principles underlying the confusion training approach. However, it is crucial to emphasize that our analysis  \textbf{does not} extend to providing a provable security guarantee for confusion training within the context of general deep learning models.}


\begin{theorem}[Dynamics of Confusion Training (informal)]
For a sufficiently small learning rate, given the specified data distribution and backdoor attack, for an overparameterized model initialized with $\thetact \leftarrow \A(\Dpoiall)$, there is an integer $T > 0$ such that when training on confusion dataset for $T$ iterations, it holds with high probability (with respect to data generation process) that the poison detector $(\charf,\decf)$ produced by Confusion Training will have a TPR of at least $1-\epsilon$ and FPR of at most $\epsilon$ on the poisoned dataset $\Dpoiall$, for an exponentially small $\eps$.
\end{theorem}
\begin{proof-sketch}
The detailed setup and proof sketch are deferred to Appendix~\ref{appendix:theory-proof}. 
The proof uses the result from \citet{chatterji2021finite} such that under sufficiently small learning rate and mild assumptions on the number of samples, the training dynamics of overparameterized linear model are tractable. This allows us to track the forgetting dynamics of clean and backdoored data points during the confusion training stage. 
\end{proof-sketch}

\subsection{\revision{Proof}}
\label{appendix:theory-proof}

In this section, we formally characterize the forgetting dynamics of backdoored and clean data in the confusion training process, in a simplified setting of the overparameterized linear model and binary classification. 

\paragraph{Setup and Notations.} 
We study a classic data distribution as follows: denote index set $\I_{-1} := \{1, \ldots, k\}$ and $\I_{+1} := \{k+1, \ldots, 2k\}$ where $k$ is a positive integer $< d/2$. 
Given any constant $\mu > 0$, we write $\bmu_{\I} \in \R^d$ to denote the vector with $\mu$ at indices in $\I$, and $0$ at all the rest of the indices. 
The conditional distribution of the data points given the label $y$ is constructed as follows: 
\begin{enumerate}
    \item If $y = -1$, then $x \sim \cleandist_{-1} := \bmu_{\I_{-1}} + \N(0, \Ind_d)$. 
    \item If $y = +1$, then $x \sim \cleandist_{+1} := \bmu_{\I_{+1}} + \N(0, \Ind_d)$. 
\end{enumerate}
Moreover, let $\I_{\poi} := \{2k+1, \ldots, 3k\}$. 
We consider a simple tigger function as follows: $\trigger(x) = x + \bmu_{\I_{\poi}}$. 
In this analysis, we assume the target label $t = +1$. 

Recall that for confusion training, the model is first trained on the corrupted dataset that contains poisoned data 
\begin{align}
    \Da = \Dclean \cup \Dpoi
\end{align}
, and then is fine-tuned on the mix of the corrupted datasets and reserved datasets with random labels
\begin{align}
    \Dconf := \Da \cup \Dreserve
\end{align}
Since we are considering the case of binary classification, we assume all labels in $\Dreserve$ are flipped for simplicity. 

We denote all data points in $\Dclean$ with label $-1$ as $\Dzero$, and all data points in $\Dclean$ with label $+1$ as $\Done$. That is, $\Dclean = \Dzero \cup \Done \cup \Dpoi$. 
We denote all data points in $\Dreserve$ that is drawn from $\cleandist_{-1}$ as $\Dzerobar$, and all data points in $\Dreserve$ that is drawn from $\cleandist_{+1}$ as $\Donebar$. That is, $\Dreserve = \Dzerobar \cup \Donebar$. 
Overall, we have 
\begin{align}
    \Dconf = \Dzero \cup \Done \cup \Dpoi \cup \Dzerobar \cup \Donebar
\end{align}

We analyze the learning and forgetting dynamics of finetuning on $\thetact \leftarrow \erm(\Dpoiall)$ when using exponential loss $\loss(\theta, (x, y)) = \exp( - y (\theta x))$. 
Following \citet{chatterji2021finite}, we make the following assumptions: for a constant $C > 0$,
\begin{enumerate}
    \item The failure probability $0 \le \delta \le 1/C$. 
    \item The number of samples $\tilden := |\Dclean| \ge C \log(1/\delta)$. 
    \item The data dimension $d \ge C \max( \tilden^2 \log(\tilden/\delta), \tilden(k \mu^2) )$, and $k \mu^2 \ge C \log(\tilden/\delta)$. 
\end{enumerate}

We can obtain the following lemma about the separability of $\Dpoiall$. 
\begin{lemma}
With probability at least $1-\delta$, $\Dpoiall$ is linearly separable. 
\end{lemma}

We perform gradient descent with a fixed learning rate $\eta$, 
\begin{align}
    \thetact^{(t+1)} = \thetact^{(t)} - \eta \sum_{(x, y) \in \Dconf} \loss'(y \thetact^{(t)} x) \cdot (y x)
\end{align}
Note that $\thetact^{(0)} = \A(\Dpoiall)$. 

\paragraph{Training Dynamics of Overparameterized Linear Models.}
For sufficiently small $\eta$, if we train on linearly separable $\Dpoiall$ for $T_0$ iterations, \citet{soudry2018implicit} shows that 
\begin{align}
    \thetact^{(0)} \approx \widehat \thetact \log(T_0)
\end{align}
where $\widehat \thetact$ is the max-margin solution 
\begin{align}
    \widehat \thetact = \argmin_{\theta \in \R^d} \norm{\theta} ~~s.t.~~y\theta x \ge 1~~\forall (x, y) \in \Dpoiall
\end{align}

By representer's theorem, we can decompose the model parameters $\thetact^{(t)}$ to 
\begin{align}
    \thetact^{(t)} = \thetact^{(0)} + \sum_{(x, y) \in \Dconf} \beta_{(x, y)}^{(t)} y x 
    \label{eq:representer}
\end{align}
for $\beta_{(x, y)}^{(t)} \ge 0$. 
We sometimes omit the time stamp and write $\beta_{(x, y)}$ for readability.

For a clean $(x^*, y^*)$, we denote the correctness of $\thetact^{(t)}$ on it as $\atclean$. 
For a backdoored $(x^*, y^*)$, we denote the correctness of $\thetact^{(t)}$ on it as $\atbackdoor$. 

By expression (\ref{eq:representer}), we have
\begin{align*}
a_{(x^*, y^*)}
&= (y^*) \thetact^{(t)} (x^*) \\
&= (y^*) (x^*) \left[ \thetact^{(0)} + \sum_{(x, y) \in \Dconf} \beta_{(x, y)} y x \right] \\
&= (y^*) (x^*) \thetact^{(0)} + \sum_{(x, y) \in \Dconf} \beta_{(x, y)} y (y^*) x (x^*)
\end{align*}

Denote the following quantities:
\begin{align*}
    \epsone &:= \sum_{i=1}^d \N(0, 1) \cdot \N(0, 1) \\
    \epstwo &:= \N(0, k\mu^2) \\
    \epsb &:= \N(0, \kb \nu^2)
\end{align*}

Moreover, we use $\azeroclean$ and  $\azerobackdoor$ to denote the correctness of the max-margin solution on clean and backdoored data, respectively. 

For a clean data $(\xstar, -1)$, we have 

\begin{small}
\begin{align*}
    &\atclean \\
    &= \azeroclean \log(T_0) + \epsone \sum_{(x, y) \in \Dconf} \beta_{(x, y)} \\
    & + \left( \sum_{(x, y) \in \Dzero} \beta_{(x, y)}
    - \sum_{(x, y) \in \Dzerobar} \beta_{(x, y)}\right) [ k\mu^2 + \epsone + 2 \epstwo ] \\
    & - \left( \sum_{(x, y) \in \Done} \beta_{(x, y)}
    - \sum_{(x, y) \in \Donebar} \beta_{(x, y)}\right) [ \epsone + 2 \epstwo ] \\ 
    & - \sum_{(x, y) \in \Dpoi} \beta_{(x, y)} [ k\mu^2 + \epsone + 2 \epstwo + \epsb ] \\
    &= \azeroclean \log(T_0) + \\
    &+ k\mu^2 \left( \sum_{(x, y) \in \Dzero} \beta_{(x, y)}
    - \sum_{(x, y) \in \Dzerobar} \beta_{(x, y)} 
    - \sum_{(x, y) \in \Dpoi} \beta_{(x, y)}
    \right) \\
    &
    + \N \left(0, 2k\mu^2  \sum_{(x, y) \in \Dconf} \beta_{(x, y)} + \kb \nu^2 \sum_{(x, y) \in \Dpoi} \beta_{(x, y)} \right)
\end{align*}
\end{small}

Similarly, for a backdoored data $(\xstar, +1)$, we have 

\begin{small}
\begin{align*}
    &\atbackdoor \\
    &= \azerobackdoor \log(T_0) + \epsone \sum_{(x, y) \in \Dconf} \beta_{(x, y)}\\
    &+ k\mu^2 \left( - \sum_{(x, y) \in \Dzero} \beta_{(x, y)} + \sum_{(x, y) \in \Dzerobar} \beta_{(x, y)} + \sum_{(x, y) \in \Dpoi} \beta_{(x, y)}
    \right) \\
    &+ \N \left(0, 2k\mu^2  \sum_{(x, y) \in \Dconf} \beta_{(x, y)} + \kb \nu^2 \sum_{(x, y) \in \Dpoi} \beta_{(x, y)} \right) \\
    &+ \N\left(0, \kb \nu^2 \sum_{(x, y) \in \Dconf} \beta_{(x, y)} \right)
    + \kb \nu^2 \sum_{(x, y) \in \Dpoi} \beta_{(x, y)}
\end{align*}
\end{small}

Denote 
\begin{align*}
    C_1 &:= k\mu^2 \left( - \sum_{(x, y) \in \Dzero} \beta_{(x, y)} + \sum_{(x, y) \in \Dzerobar} \beta_{(x, y)} + \sum_{(x, y) \in \Dpoi} \beta_{(x, y)}
    \right) \\
    C_2 &:= \kb \nu^2 \sum_{(x, y) \in \Dpoi} \beta_{(x, y)} \\
    Z_1 &:= \zone \\
    Z_2 &:= \ztwo \\
    Z_3 &:= \zthree
\end{align*}

Our goal is to find a $t$ s.t. $\Pr[\atbackdoor > 0]$ is large and $\Pr[\atclean > 0]$ is small. Note that 
\begin{align}
    \Pr[\atbackdoor < 0] 
    &= \Pr[ Z_1 + Z_2 + Z_3 \le -C_1 -C_2 ] \\
    &\le  \Pr[ Z_1 + Z_2 + Z_3 \le -C_1 ] \\
    &= \Pr[ Z_1 + Z_2 + Z_3 \ge C_1 ] \\
    &= \Pr[\atclean > 0] 
\end{align}
Therefore, we can pick the time $t^* = \argmax_t C_1(t)$. 
Due to the identical distribution of $\Dzero$ and $\Dzerobar$, we would expect $- \sum_{(x, y) \in \Dzero} \beta_{(x, y)}^{(t)} + \sum_{(x, y) \in \Dzerobar} \beta_{(x, y)}^{(t)} \approx 0$ for all $t$. 
Hence, $C_1 \approx \sum_{(x, y) \in \Dpoi} \beta_{(x, y)}^{(t)} = O(k^2 \mu^2)$. 
Let $\Delta := \sum_{(x, y) \in \Dpoi} \beta_{(x, y)}^{(t^*)} / k$
Therefore, at time $t^*$, we have 
\begin{align}
    \Pr[\atbackdoor < 0] 
    \le O(\exp(-k^2\mu^2 \Delta / d)) 
\end{align}
and 
\begin{align}
    \Pr[\atclean > 0] 
    \le O(\exp(-k^2\mu^2 \Delta / d)) 
\end{align}

%% file: usenix.bbl
\begin{thebibliography}{69}
\providecommand{\natexlab}[1]{#1}
\providecommand{\url}[1]{\texttt{#1}}
\expandafter\ifx\csname urlstyle\endcsname\relax
  \providecommand{\doi}[1]{doi: #1}\else
  \providecommand{\doi}{doi: \begingroup \urlstyle{rm}\Url}\fi

\bibitem[Anderson and Roth(2018)]{anderson2018ember}
Hyrum~S Anderson and Phil Roth.
\newblock Ember: an open dataset for training static pe malware machine
  learning models.
\newblock \emph{arXiv preprint arXiv:1804.04637}, 2018.

\bibitem[Bagdasaryan and Shmatikov(2021)]{bagdasaryan2021blind}
Eugene Bagdasaryan and Vitaly Shmatikov.
\newblock Blind backdoors in deep learning models.
\newblock In \emph{30th USENIX Security Symposium (USENIX Security 21)}, pages
  1505--1521, 2021.

\bibitem[Barni et~al.(2019)Barni, Kallas, and Tondi]{barni2019new}
Mauro Barni, Kassem Kallas, and Benedetta Tondi.
\newblock A new backdoor attack in cnns by training set corruption without
  label poisoning.
\newblock In \emph{2019 IEEE International Conference on Image Processing
  (ICIP)}, pages 101--105. IEEE, 2019.

\bibitem[Chatterji and Long(2021)]{chatterji2021finite}
Niladri~S Chatterji and Philip~M Long.
\newblock Finite-sample analysis of interpolating linear classifiers in the
  overparameterized regime.
\newblock \emph{The Journal of Machine Learning Research}, 22\penalty0
  (1):\penalty0 5721--5750, 2021.

\bibitem[Chen et~al.(2018)Chen, Carvalho, Baracaldo, Ludwig, Edwards, Lee,
  Molloy, and Srivastava]{chen2018activationclustering}
Bryant Chen, Wilka Carvalho, Nathalie Baracaldo, Heiko Ludwig, Benjamin
  Edwards, Taesung Lee, Ian Molloy, and Biplav Srivastava.
\newblock Detecting backdoor attacks on deep neural networks by activation
  clustering.
\newblock \emph{arXiv preprint arXiv:1811.03728}, 2018.

\bibitem[Chen et~al.(2017)Chen, Liu, Li, Lu, and Song]{Chen2017TargetedBA}
Xinyun Chen, Chang Liu, Bo~Li, Kimberly Lu, and Dawn~Xiaodong Song.
\newblock Targeted backdoor attacks on deep learning systems using data
  poisoning.
\newblock \emph{ArXiv}, abs/1712.05526, 2017.

\bibitem[Choquette-Choo et~al.(2021)Choquette-Choo, Tramer, Carlini, and
  Papernot]{choquette2021label}
Christopher~A Choquette-Choo, Florian Tramer, Nicholas Carlini, and Nicolas
  Papernot.
\newblock Label-only membership inference attacks.
\newblock In \emph{International Conference on Machine Learning}, pages
  1964--1974. PMLR, 2021.

\bibitem[Chou et~al.(2020)Chou, Tramer, and Pellegrino]{chou2020sentinet}
Edward Chou, Florian Tramer, and Giancarlo Pellegrino.
\newblock Sentinet: Detecting localized universal attack against deep learning
  systems.
\newblock \emph{IEEE SPW 2020}, 2020.

\bibitem[Deng et~al.(2009)Deng, Dong, Socher, Li, Li, and
  Fei-Fei]{deng2009imagenet}
Jia Deng, Wei Dong, Richard Socher, Li-Jia Li, Kai Li, and Li~Fei-Fei.
\newblock Imagenet: A large-scale hierarchical image database.
\newblock In \emph{2009 IEEE conference on computer vision and pattern
  recognition}, pages 248--255. Ieee, 2009.

\bibitem[Dosovitskiy et~al.(2020)Dosovitskiy, Beyer, Kolesnikov, Weissenborn,
  Zhai, Unterthiner, Dehghani, Minderer, Heigold, Gelly,
  et~al.]{dosovitskiy2020image}
Alexey Dosovitskiy, Lucas Beyer, Alexander Kolesnikov, Dirk Weissenborn,
  Xiaohua Zhai, Thomas Unterthiner, Mostafa Dehghani, Matthias Minderer, Georg
  Heigold, Sylvain Gelly, et~al.
\newblock An image is worth 16x16 words: Transformers for image recognition at
  scale.
\newblock \emph{arXiv preprint arXiv:2010.11929}, 2020.

\bibitem[Du et~al.(2019)Du, Jia, and Song]{du2019robust}
Min Du, Ruoxi Jia, and Dawn Song.
\newblock Robust anomaly detection and backdoor attack detection via
  differential privacy.
\newblock \emph{arXiv preprint arXiv:1911.07116}, 2019.

\bibitem[Gao et~al.(2019)Gao, Xu, Wang, Chen, Ranasinghe, and
  Nepal]{gao2019strip}
Yansong Gao, Change Xu, Derui Wang, Shiping Chen, Damith~C Ranasinghe, and
  Surya Nepal.
\newblock Strip: A defence against trojan attacks on deep neural networks.
\newblock In \emph{Proceedings of the 35th Annual Computer Security
  Applications Conference}, pages 113--125, 2019.

\bibitem[Gu et~al.(2017)Gu, Dolan-Gavitt, and Garg]{gu2017badnets}
Tianyu Gu, Brendan Dolan-Gavitt, and Siddharth Garg.
\newblock Badnets: Identifying vulnerabilities in the machine learning model
  supply chain.
\newblock \emph{arXiv preprint arXiv:1708.06733}, 2017.

\bibitem[Hayase et~al.(2021)Hayase, Kong, Somani, and Oh]{hayase21a}
Jonathan Hayase, Weihao Kong, Raghav Somani, and Sewoong Oh.
\newblock Spectre: defending against backdoor attacks using robust statistics.
\newblock In Marina Meila and Tong Zhang, editors, \emph{Proceedings of the
  38th International Conference on Machine Learning}, volume 139 of
  \emph{Proceedings of Machine Learning Research}, pages 4129--4139. PMLR,
  18--24 Jul 2021.
\newblock URL \url{https://proceedings.mlr.press/v139/hayase21a.html}.

\bibitem[He et~al.(2016)He, Zhang, Ren, and Sun]{he2016deep}
Kaiming He, Xiangyu Zhang, Shaoqing Ren, and Jian Sun.
\newblock Deep residual learning for image recognition.
\newblock In \emph{Proceedings of the IEEE conference on computer vision and
  pattern recognition}, pages 770--778, 2016.

\bibitem[Huang et~al.(2017)Huang, Liu, Van Der~Maaten, and
  Weinberger]{huang2017densely}
Gao Huang, Zhuang Liu, Laurens Van Der~Maaten, and Kilian~Q Weinberger.
\newblock Densely connected convolutional networks.
\newblock In \emph{Proceedings of the IEEE conference on computer vision and
  pattern recognition}, pages 4700--4708, 2017.

\bibitem[Huang et~al.(2022)Huang, Li, Wu, Qin, and Ren]{huang2022backdoor}
Kunzhe Huang, Yiming Li, Baoyuan Wu, Zhan Qin, and Kui Ren.
\newblock Backdoor defense via decoupling the training process.
\newblock \emph{arXiv preprint arXiv:2202.03423}, 2022.

\bibitem[Krizhevsky et~al.(2009)Krizhevsky, Hinton,
  et~al.]{krizhevsky2009learning}
Alex Krizhevsky, Geoffrey Hinton, et~al.
\newblock Learning multiple layers of features from tiny images.
\newblock 2009.

\bibitem[Li et~al.(2019)Li, Tai, and Weinan]{li2019stochastic}
Qianxiao Li, Cheng Tai, and E~Weinan.
\newblock Stochastic modified equations and dynamics of stochastic gradient
  algorithms i: Mathematical foundations.
\newblock \emph{The Journal of Machine Learning Research}, 20\penalty0
  (1):\penalty0 1474--1520, 2019.

\bibitem[Li et~al.(2021{\natexlab{a}})Li, Lyu, Koren, Lyu, Li, and
  Ma]{li2021anti}
Yige Li, Xixiang Lyu, Nodens Koren, Lingjuan Lyu, Bo~Li, and Xingjun Ma.
\newblock Anti-backdoor learning: Training clean models on poisoned data.
\newblock \emph{Advances in Neural Information Processing Systems}, 34,
  2021{\natexlab{a}}.

\bibitem[Li et~al.(2021{\natexlab{b}})Li, Lyu, Koren, Lyu, Li, and
  Ma]{li2021neural}
Yige Li, Xixiang Lyu, Nodens Koren, Lingjuan Lyu, Bo~Li, and Xingjun Ma.
\newblock Neural attention distillation: Erasing backdoor triggers from deep
  neural networks.
\newblock \emph{arXiv preprint arXiv:2101.05930}, 2021{\natexlab{b}}.

\bibitem[Li et~al.(2022)Li, Jiang, Li, and Xia]{li2022backdoor}
Yiming Li, Yong Jiang, Zhifeng Li, and Shu-Tao Xia.
\newblock Backdoor learning: A survey.
\newblock \emph{IEEE Transactions on Neural Networks and Learning Systems},
  2022.

\bibitem[Li et~al.(2021{\natexlab{c}})Li, Li, Wu, Li, He, and
  Lyu]{li2021invisible}
Yuezun Li, Yiming Li, Baoyuan Wu, Longkang Li, Ran He, and Siwei Lyu.
\newblock Invisible backdoor attack with sample-specific triggers.
\newblock In \emph{Proceedings of the IEEE/CVF International Conference on
  Computer Vision}, pages 16463--16472, 2021{\natexlab{c}}.

\bibitem[Lin et~al.(2014)Lin, Maire, Belongie, Hays, Perona, Ramanan,
  Doll{\'a}r, and Zitnick]{lin2014microsoft}
Tsung-Yi Lin, Michael Maire, Serge Belongie, James Hays, Pietro Perona, Deva
  Ramanan, Piotr Doll{\'a}r, and C~Lawrence Zitnick.
\newblock Microsoft coco: Common objects in context.
\newblock In \emph{European conference on computer vision}, pages 740--755.
  Springer, 2014.

\bibitem[Liu et~al.(2018{\natexlab{a}})Liu, Dolan-Gavitt, and
  Garg]{liu2018fine}
Kang Liu, Brendan Dolan-Gavitt, and Siddharth Garg.
\newblock Fine-pruning: Defending against backdooring attacks on deep neural
  networks.
\newblock In \emph{International Symposium on Research in Attacks, Intrusions,
  and Defenses}, pages 273--294. Springer, 2018{\natexlab{a}}.

\bibitem[Liu et~al.(2021)Liu, Li, Wen, and Li]{liu2021removing}
Xuankai Liu, Fengting Li, Bihan Wen, and Qi~Li.
\newblock Removing backdoor-based watermarks in neural networks with limited
  data.
\newblock In \emph{2020 25th International Conference on Pattern Recognition
  (ICPR)}, pages 10149--10156. IEEE, 2021.

\bibitem[Liu et~al.(2017{\natexlab{a}})Liu, Wei, Luo, and Xu]{liu2017fault}
Yannan Liu, Lingxiao Wei, Bo~Luo, and Qiang Xu.
\newblock Fault injection attack on deep neural network.
\newblock In \emph{2017 IEEE/ACM International Conference on Computer-Aided
  Design (ICCAD)}, pages 131--138. IEEE, 2017{\natexlab{a}}.

\bibitem[Liu et~al.(2018{\natexlab{b}})Liu, Ma, Aafer, Lee, Zhai, Wang, and
  Zhang]{liu2017trojaning}
Yingqi Liu, Shiqing Ma, Yousra Aafer, W.~Lee, Juan Zhai, Weihang Wang, and
  X.~Zhang.
\newblock Trojaning attack on neural networks.
\newblock In \emph{NDSS}, 2018{\natexlab{b}}.

\bibitem[Liu et~al.(2020)Liu, Ma, Bailey, and Lu]{liu2020reflection}
Yunfei Liu, Xingjun Ma, James Bailey, and Feng Lu.
\newblock Reflection backdoor: A natural backdoor attack on deep neural
  networks.
\newblock In \emph{European Conference on Computer Vision}, pages 182--199.
  Springer, 2020.

\bibitem[Liu et~al.(2017{\natexlab{b}})Liu, Xie, and Srivastava]{liu2017neural}
Yuntao Liu, Yang Xie, and Ankur Srivastava.
\newblock Neural trojans.
\newblock In \emph{2017 IEEE International Conference on Computer Design
  (ICCD)}, pages 45--48. IEEE, 2017{\natexlab{b}}.

\bibitem[Lloyd(1982)]{lloyd1982least}
Stuart Lloyd.
\newblock Least squares quantization in pcm.
\newblock \emph{IEEE transactions on information theory}, 28\penalty0
  (2):\penalty0 129--137, 1982.

\bibitem[Marcel and Rodriguez(2010)]{marcel2010torchvision}
S{\'e}bastien Marcel and Yann Rodriguez.
\newblock Torchvision the machine-vision package of torch.
\newblock In \emph{Proceedings of the 18th ACM international conference on
  Multimedia}, pages 1485--1488, 2010.

\bibitem[Mori et~al.(2022)Mori, Ziyin, Liu, and Ueda]{mori2022power}
Takashi Mori, Liu Ziyin, Kangqiao Liu, and Masahito Ueda.
\newblock Power-law escape rate of sgd.
\newblock In \emph{International Conference on Machine Learning}, pages
  15959--15975. PMLR, 2022.

\bibitem[Nguyen and Tran(2021)]{nguyen2021wanet}
Anh Nguyen and Anh Tran.
\newblock Wanet--imperceptible warping-based backdoor attack.
\newblock \emph{arXiv preprint arXiv:2102.10369}, 2021.

\bibitem[Nguyen and Tran(2020)]{nguyen2020input}
Tuan~Anh Nguyen and Anh Tran.
\newblock Input-aware dynamic backdoor attack.
\newblock \emph{Advances in Neural Information Processing Systems},
  33:\penalty0 3454--3464, 2020.

\bibitem[Paul et~al.(2021)Paul, Ganguli, and Dziugaite]{paul2021deep}
Mansheej Paul, Surya Ganguli, and Gintare~Karolina Dziugaite.
\newblock Deep learning on a data diet: Finding important examples early in
  training.
\newblock \emph{Advances in Neural Information Processing Systems},
  34:\penalty0 20596--20607, 2021.

\bibitem[Pearson(1901)]{pearson1901liii}
Karl Pearson.
\newblock Liii. on lines and planes of closest fit to systems of points in
  space.
\newblock \emph{The London, Edinburgh, and Dublin philosophical magazine and
  journal of science}, 2\penalty0 (11):\penalty0 559--572, 1901.

\bibitem[Qi et~al.(2021)Qi, Zhu, Xie, and Yang]{qi2021subnet}
Xiangyu Qi, Jifeng Zhu, Chulin Xie, and Yong Yang.
\newblock Subnet replacement: Deployment-stage backdoor attack against deep
  neural networks in gray-box setting.
\newblock \emph{arXiv preprint arXiv:2107.07240}, 2021.

\bibitem[Qi et~al.(2022)Qi, Xie, Pan, Zhu, Yang, and Bu]{qi2021towards}
Xiangyu Qi, Tinghao Xie, Ruizhe Pan, Jifeng Zhu, Yong Yang, and Kai Bu.
\newblock Towards practical deployment-stage backdoor attack on deep neural
  networks.
\newblock In \emph{Proceedings of the IEEE/CVF Conference on Computer Vision
  and Pattern Recognition}, pages 13347--13357, 2022.

\bibitem[Qi et~al.(2023)Qi, Xie, Li, Mahloujifar, and Mittal]{qi2023revisiting}
Xiangyu Qi, Tinghao Xie, Yiming Li, Saeed Mahloujifar, and Prateek Mittal.
\newblock Revisiting the assumption of latent separability for backdoor
  defenses.
\newblock In \emph{International Conference on Learning Representations}, 2023.

\bibitem[Rakin et~al.(2020)Rakin, He, and Fan]{rakin2020tbt}
Adnan~Siraj Rakin, Zhezhi He, and Deliang Fan.
\newblock Tbt: Targeted neural network attack with bit trojan.
\newblock In \emph{Proceedings of the IEEE/CVF Conference on Computer Vision
  and Pattern Recognition}, pages 13198--13207, 2020.

\bibitem[Russakovsky et~al.(2015)Russakovsky, Deng, Su, Krause, Satheesh, Ma,
  Huang, Karpathy, Khosla, Bernstein, et~al.]{russakovsky2015imagenet}
Olga Russakovsky, Jia Deng, Hao Su, Jonathan Krause, Sanjeev Satheesh, Sean Ma,
  Zhiheng Huang, Andrej Karpathy, Aditya Khosla, Michael Bernstein, et~al.
\newblock Imagenet large scale visual recognition challenge.
\newblock \emph{International journal of computer vision}, 115\penalty0
  (3):\penalty0 211--252, 2015.

\bibitem[Sandler et~al.(2018)Sandler, Howard, Zhu, Zhmoginov, and
  Chen]{sandler2018mobilenetv2}
Mark Sandler, Andrew Howard, Menglong Zhu, Andrey Zhmoginov, and Liang-Chieh
  Chen.
\newblock Mobilenetv2: Inverted residuals and linear bottlenecks.
\newblock In \emph{Proceedings of the IEEE conference on computer vision and
  pattern recognition}, pages 4510--4520, 2018.

\bibitem[Selvaraju et~al.(2017)Selvaraju, Cogswell, Das, Vedantam, Parikh, and
  Batra]{selvaraju2017grad}
Ramprasaath~R Selvaraju, Michael Cogswell, Abhishek Das, Ramakrishna Vedantam,
  Devi Parikh, and Dhruv Batra.
\newblock Grad-cam: Visual explanations from deep networks via gradient-based
  localization.
\newblock In \emph{Proceedings of the IEEE international conference on computer
  vision}, pages 618--626, 2017.

\bibitem[Severi et~al.(2020)Severi, Meyer, Coull, and
  Oprea]{severi2020exploring}
Giorgio Severi, Jim Meyer, Scott Coull, and Alina Oprea.
\newblock Exploring backdoor poisoning attacks against malware classifiers.
\newblock \emph{arXiv preprint arXiv:2003.01031}, 2020.

\bibitem[Severi et~al.(2021)Severi, Meyer, Coull, and
  Oprea]{severi2021explanation}
Giorgio Severi, Jim Meyer, Scott Coull, and Alina Oprea.
\newblock $\{$Explanation-Guided$\}$ backdoor poisoning attacks against malware
  classifiers.
\newblock In \emph{30th USENIX Security Symposium (USENIX Security 21)}, pages
  1487--1504, 2021.

\bibitem[Shen et~al.(2021)Shen, Ji, Zhang, Li, Chen, Shi, Fang, Yin, and
  Wang]{shen2021backdoor}
Lujia Shen, Shouling Ji, Xuhong Zhang, Jinfeng Li, Jing Chen, Jie Shi,
  Chengfang Fang, Jianwei Yin, and Ting Wang.
\newblock Backdoor pre-trained models can transfer to all.
\newblock \emph{arXiv preprint arXiv:2111.00197}, 2021.

\bibitem[Simonyan and Zisserman(2014)]{vgg}
Karen Simonyan and Andrew Zisserman.
\newblock Very deep convolutional networks for large-scale image recognition.
\newblock \emph{arXiv preprint arXiv:1409.1556}, 2014.

\bibitem[Soudry et~al.(2018)Soudry, Hoffer, Nacson, Gunasekar, and
  Srebro]{soudry2018implicit}
Daniel Soudry, Elad Hoffer, Mor~Shpigel Nacson, Suriya Gunasekar, and Nathan
  Srebro.
\newblock The implicit bias of gradient descent on separable data.
\newblock \emph{The Journal of Machine Learning Research}, 19\penalty0
  (1):\penalty0 2822--2878, 2018.

\bibitem[Stallkamp et~al.(2012)Stallkamp, Schlipsing, Salmen, and
  Igel]{stallkamp2012man}
Johannes Stallkamp, Marc Schlipsing, Jan Salmen, and Christian Igel.
\newblock Man vs. computer: Benchmarking machine learning algorithms for
  traffic sign recognition.
\newblock \emph{Neural networks}, 32:\penalty0 323--332, 2012.

\bibitem[Tan and Shokri(2019)]{tan2019bypassing}
Te~Juin~Lester Tan and Reza Shokri.
\newblock Bypassing backdoor detection algorithms in deep learning.
\newblock \emph{arXiv preprint arXiv:1905.13409}, 2019.

\bibitem[Tang et~al.(2021)Tang, Wang, Tang, and Zhang]{tang2021demon}
Di~Tang, XiaoFeng Wang, Haixu Tang, and Kehuan Zhang.
\newblock Demon in the variant: Statistical analysis of dnns for robust
  backdoor contamination detection.
\newblock In \emph{30th $\{$USENIX$\}$ Security Symposium ($\{$USENIX$\}$
  Security 21)}, 2021.

\bibitem[Tao et~al.(2022)Tao, Liu, Shen, Xu, An, Zhang, and
  Zhang]{tao2022model}
Guanhong Tao, Yingqi Liu, Guangyu Shen, Qiuling Xu, Shengwei An, Zhuo Zhang,
  and Xiangyu Zhang.
\newblock Model orthogonalization: Class distance hardening in neural networks
  for better security.
\newblock In \emph{2022 IEEE Symposium on Security and Privacy (SP)}, pages
  1372--1389. IEEE, 2022.

\bibitem[Thomee et~al.(2016)Thomee, Shamma, Friedland, Elizalde, Ni, Poland,
  Borth, and Li]{thomee2016yfcc100m}
Bart Thomee, David~A Shamma, Gerald Friedland, Benjamin Elizalde, Karl Ni,
  Douglas Poland, Damian Borth, and Li-Jia Li.
\newblock Yfcc100m: The new data in multimedia research.
\newblock \emph{Communications of the ACM}, 59\penalty0 (2):\penalty0 64--73,
  2016.

\bibitem[Tran et~al.(2018)Tran, Li, and Madry]{tran2018spectral}
Brandon Tran, Jerry Li, and Aleksander Madry.
\newblock Spectral signatures in backdoor attacks.
\newblock In \emph{Advances in Neural Information Processing Systems}, pages
  8000--8010, 2018.

\bibitem[Truong et~al.(2020)Truong, Jones, Hutchinson, August, Praggastis,
  Jasper, Nichols, and Tuor]{truong2020systematic}
Loc Truong, Chace Jones, Brian Hutchinson, Andrew August, Brenda Praggastis,
  Robert Jasper, Nicole Nichols, and Aaron Tuor.
\newblock Systematic evaluation of backdoor data poisoning attacks on image
  classifiers.
\newblock In \emph{Proceedings of the IEEE/CVF Conference on Computer Vision
  and Pattern Recognition Workshops}, pages 788--789, 2020.

\bibitem[Turner et~al.(2019)Turner, Tsipras, and Madry]{turner2019label}
Alexander Turner, Dimitris Tsipras, and Aleksander Madry.
\newblock Label-consistent backdoor attacks.
\newblock \emph{arXiv preprint arXiv:1912.02771}, 2019.

\bibitem[Udeshi et~al.(2019)Udeshi, Peng, Woo, Loh, Rawshan, and
  Chattopadhyay]{udeshi2019model}
Sakshi Udeshi, Shanshan Peng, Gerald Woo, Lionell Loh, Louth Rawshan, and
  Sudipta Chattopadhyay.
\newblock Model agnostic defence against backdoor attacks in machine learning.
\newblock \emph{arXiv preprint arXiv:1908.02203}, 2019.

\bibitem[Wang et~al.(2020)Wang, Cao, Gong, et~al.]{wang2020certifying}
Binghui Wang, Xiaoyu Cao, Neil~Zhenqiang Gong, et~al.
\newblock On certifying robustness against backdoor attacks via randomized
  smoothing.
\newblock \emph{arXiv preprint arXiv:2002.11750}, 2020.

\bibitem[Wang et~al.(2019)Wang, Yao, Shan, Li, Viswanath, Zheng, and
  Zhao]{wang2019neural}
Bolun Wang, Yuanshun Yao, Shawn Shan, Huiying Li, Bimal Viswanath, Haitao
  Zheng, and Ben~Y Zhao.
\newblock Neural cleanse: Identifying and mitigating backdoor attacks in neural
  networks.
\newblock In \emph{2019 IEEE Symposium on Security and Privacy (SP)}, pages
  707--723. IEEE, 2019.

\bibitem[Wang et~al.(2022)Wang, Ding, Zhai, and Ma]{wang2022training}
Zhenting Wang, Hailun Ding, Juan Zhai, and Shiqing Ma.
\newblock Training with more confidence: Mitigating injected and natural
  backdoors during training.
\newblock \emph{Advances in Neural Information Processing Systems},
  35:\penalty0 36396--36410, 2022.

\bibitem[Wu et~al.(2022)Wu, Wang, Sehwag, Mahloujifar, and Mittal]{wu2022just}
Tong Wu, Tianhao Wang, Vikash Sehwag, Saeed Mahloujifar, and Prateek Mittal.
\newblock Just rotate it: Deploying backdoor attacks via rotation
  transformation.
\newblock \emph{arXiv preprint arXiv:2207.10825}, 2022.

\bibitem[Xu et~al.(2021)Xu, Wang, Li, Borisov, Gunter, and Li]{xu2019Meta}
Xiaojun Xu, Qi~Wang, Huichen Li, Nikita Borisov, Carl~A Gunter, and Bo~Li.
\newblock Detecting ai trojans using meta neural analysis.
\newblock In \emph{Proceedings of the IEEE Symposium on Security and Privacy
  (May 2021)}, 2021.

\bibitem[Yao et~al.(2019)Yao, Li, Zheng, and Zhao]{yao2019latent}
Yuanshun Yao, Huiying Li, Haitao Zheng, and Ben~Y Zhao.
\newblock Latent backdoor attacks on deep neural networks.
\newblock In \emph{Proceedings of the 2019 ACM SIGSAC Conference on Computer
  and Communications Security}, pages 2041--2055, 2019.

\bibitem[Zancato et~al.(2020)Zancato, Achille, Ravichandran, Bhotika, and
  Soatto]{zancato2020predicting}
Luca Zancato, Alessandro Achille, Avinash Ravichandran, Rahul Bhotika, and
  Stefano Soatto.
\newblock Predicting training time without training.
\newblock \emph{Advances in Neural Information Processing Systems},
  33:\penalty0 6136--6146, 2020.

\bibitem[Zeng et~al.(2021)Zeng, Park, Mao, and Jia]{zeng2021rethinking}
Yi~Zeng, Won Park, Z~Morley Mao, and Ruoxi Jia.
\newblock Rethinking the backdoor attacks' triggers: A frequency perspective.
\newblock In \emph{Proceedings of the IEEE/CVF International Conference on
  Computer Vision}, pages 16473--16481, 2021.

\bibitem[Zeng et~al.(2022)Zeng, Pan, Jahagirdar, Jin, Lyu, and
  Jia]{zeng2022sift}
Yi~Zeng, Minzhou Pan, Himanshu Jahagirdar, Ming Jin, Lingjuan Lyu, and Ruoxi
  Jia.
\newblock How to sift out a clean data subset in the presence of data
  poisoning?
\newblock \emph{arXiv preprint arXiv:2210.06516}, 2022.

\bibitem[Zhao et~al.(2020)Zhao, Chen, Das, Ramamurthy, and
  Lin]{zhao2020bridging}
Pu~Zhao, Pin-Yu Chen, Payel Das, Karthikeyan~Natesan Ramamurthy, and Xue Lin.
\newblock Bridging mode connectivity in loss landscapes and adversarial
  robustness.
\newblock \emph{arXiv preprint arXiv:2005.00060}, 2020.

\bibitem[Zhuang et~al.(2022)Zhuang, Gong, Yuan, Cui, Adam, Dvornek, Tatikonda,
  Duncan, and Liu]{zhuang2022surrogate}
Juntang Zhuang, Boqing Gong, Liangzhe Yuan, Yin Cui, Hartwig Adam, Nicha
  Dvornek, Sekhar Tatikonda, James Duncan, and Ting Liu.
\newblock Surrogate gap minimization improves sharpness-aware training.
\newblock \emph{arXiv preprint arXiv:2203.08065}, 2022.

\end{thebibliography}
